\documentclass[english,british,10pt,journal,compsoc]{IEEEtran}
\usepackage[T1]{fontenc}
\usepackage[latin9]{inputenc}
\usepackage{array}
\usepackage{url}
\usepackage{multirow}
\usepackage{amsmath}
\usepackage{graphicx}
\usepackage{esint}

\makeatletter

\providecommand{\tabularnewline}{\\}
\newcommand{\lyxdot}{.}

\usepackage{colortbl,amssymb,cite,authblk}

\usepackage[switch, displaymath]{lineno}

\newcommand{\chk}{{\centering\checkmark}}
\newcommand{\greychk}{\textcolor[rgb]{0.568,0.639,0.690}{\chk}}

\newcommand\celln{ }

\newcommand\cellp{\cellcolor[rgb]{0.8510,    0.9176,    0.8275}}
\newcommand\cellz{\cellcolor[rgb]{0.7882,    0.8549,    0.9725}}

\newcommand*\diff{\mathop{}\!\mathrm{d}}

\def\paragraph{\@startsection{paragraph}{4}{\z@}{1.5ex plus 1.5ex minus 0.5ex}{0ex}{\normalfont\normalsize\bfseries}}

  \author[1]{Jan Hosang}
  \author[1]{Rodrigo Benenson}
  \author[2]{Piotr Dollár\,}
  \author[1]{Bernt Schiele}
  \affil[1]{Max Planck Institute for Informatics}
  \affil[2]{\,Facebook AI Research (FAIR)}

\@ifundefined{showcaptionsetup}{}{%
 \PassOptionsToPackage{caption=false}{subfig}}
\usepackage{subfig}
\makeatother

\usepackage{babel}
\begin{document}
\title{What makes for effective detection proposals?}

\IEEEtitleabstractindextext{%
\begin{abstract}
Current top performing object detectors employ detection proposals
to guide the search for objects, thereby avoiding exhaustive sliding
window search across images. Despite the popularity and widespread
use of detection proposals, it is unclear which trade-offs are made
when using them during object detection. We provide an in-depth analysis
of twelve proposal methods along with four baselines regarding proposal
repeatability, ground truth annotation recall on PASCAL, ImageNet,
and MS COCO, and their impact on DPM, R-CNN, and Fast R-CNN detection
performance. Our analysis shows that for object detection improving
proposal localisation accuracy is as important as improving recall.
We introduce a novel metric, the average recall (AR), which rewards
both high recall and good localisation and correlates surprisingly
well with detection performance. Our findings show common strengths
and weaknesses of existing methods, and provide insights and metrics
for selecting and tuning proposal methods. \end{abstract}


}
\maketitle
\IEEEdisplaynontitleabstractindextext

\section{\label{sec:Introduction}Introduction}

\IEEEPARstart{U}{ntil} recently, the most successful approaches to
object detection utilised the well known ``sliding window'' paradigm
\cite{Papageorgiou2000,Viola2004Ijvc,Felzenszwalb2010Pami}, in which
a computationally efficient classifier tests for object presence in
every candidate image window. Sliding window classifiers scale linearly
with the number of windows tested, and while single-scale detection
requires classifying around $10^{4}$ -- $10^{5}$ windows per image,
the number of windows grows by an order of magnitude for multi-scale
detection. Modern detection datasets \cite{Everingham2014Ijcv,Deng2009Cvpr,mscoco2015}
also require the prediction of object aspect ratio, further increasing
the search space to $10^{6}$ -- $10^{7}$ windows per image.

The steady increase in complexity of the core classifiers has led
to improved detection quality, but at the cost of significantly increased
computation time per window \cite{Wang2013Iccv,Girshick2014Cvpr,Szegedy2014arXiv,He2014Eccv,Cinbis2013Iccv}.
One approach for overcoming the tension between computational tractability
and high detection quality is through the use of ``detection proposals''
\cite{Alexe2010Cvpr,Carreira2010Cvpr,Endres2010Eccv,Sande2011Iccv}.
Under the assumption that all objects of interest share common visual
properties that distinguish them from the background, one can design
or train a method that, given an image, outputs a set of proposal
regions that are likely to contain objects. If high object recall
can be reached with considerably fewer windows than used by sliding
window detectors, significant speed-ups can be achieved, enabling
the use of more sophisticated classifiers. 

Current top performing object detectors for PASCAL \cite{Everingham2014Ijcv}
and ImageNet \cite{Deng2009Cvpr} all use detection proposals \cite{Wang2013Iccv,Girshick2014Cvpr,Szegedy2014arXiv,He2014Eccv,Cinbis2013Iccv,Girshick2015arXiv}.
In addition to allowing for use of more sophisticated classifiers,
the use of detection proposals alters the data distribution that the
classifiers handle. This may also improve detection quality by reducing
spurious false positives. 

Most papers on generating detection proposals perform fairly limited
evaluations, comparing results using only a subset of metrics, datasets,
and competing methods. In this work, we aim to revisit existing work
on proposals and compare most publicly available methods in a unified
framework. While this requires us to carefully re-examine the metrics
and settings for evaluating proposals, it allows us to better understand
the benefits and limitations of current methods.
\begin{figure}
\begin{centering}
\includegraphics[bb=0bp 0bp 365bp 190bp,clip,width=1\columnwidth]{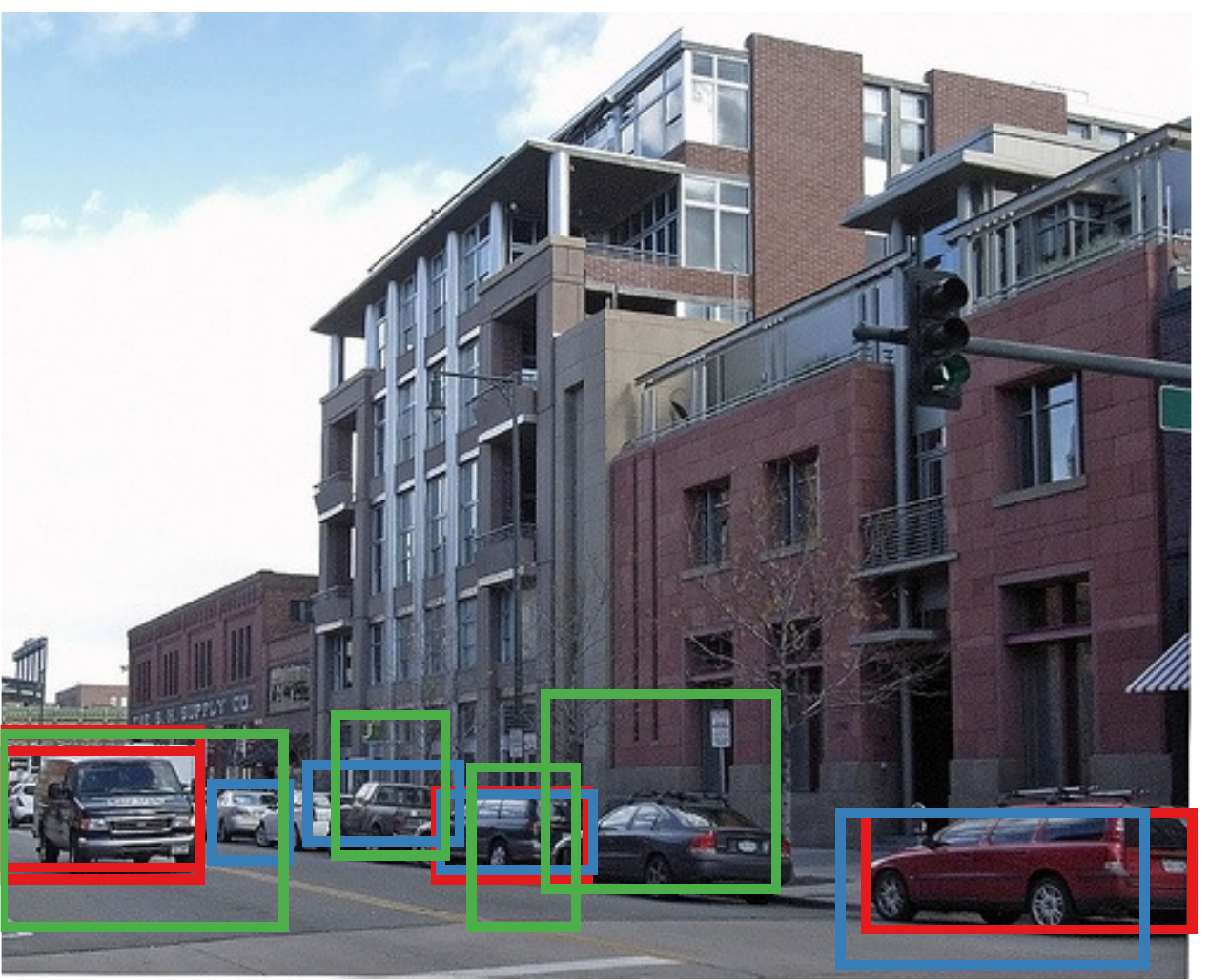}
\par\end{centering}

\protect\caption{\label{fig:teaser}What makes object detection proposals effective?}
\end{figure}

The contributions of this work are as follows:
\begin{itemize}
\item In \S\ref{sec:Methods} we provide a systematic overview of detection
proposal methods and define simple baselines that serve as reference
points. We discuss the taxonomy of proposal methods, and describe
commonalities and differences of the various approaches.
\item In \S\ref{sec:Repeatibility} we introduce the notion of proposal
repeatability, discuss its relevance when considering proposals for
detection, and measure the repeatability of existing methods. The
results are somewhat unexpected.
\item In \S\ref{sec:Proposal-recall} we study object recall on the PASCAL
VOC 2007 test set\cite{Everingham2014Ijcv}, and for the first time,
over the larger and more diverse ImageNet 2013 \cite{Deng2009Cvpr}
and MS COCO 2014 \cite{mscoco2015} validation sets. The latter allows
us to examine possible biases towards PASCAL objects categories. Overall,
these experiments are substantially broader in scope than previous
work, both in the number of methods evaluated and datasets used.
\item In \S\ref{sec:Using-detection-proposals} we evaluate the influence
of different proposal methods on DPM \cite{Felzenszwalb2010Pami},
R-CNN \cite{Girshick2014Cvpr}, and Fast R-CNN \cite{Girshick2015arXiv}
detection performance. Based on our results, we introduce a novel
evaluation metric, the average recall (AR). We show that AR is highly
correlated with detector performance, more so than previous metrics,
and we advocate AR to become the standard metric for evaluating proposals.
Our experiments provide the first clear guidelines for selecting and
tuning proposal methods for object detection.
\end{itemize}
All evaluation scripts and method bounding boxes used in this work
are publicly available to facilitate the reproduction of our evaluation%
\footnote{Project page: \url{http://goo.gl/uMhkAs}%
}. The results presented in this paper summarise results of over 500
experiments on multiple data sets and required multiple months of
CPU time.

An earlier version of this work appeared in \cite{Hosang2014Bmvc}.

\section{\label{sec:Methods}Detection proposal methods}

\begin{table*}
\begin{centering}
\begin{tabular}{ccccccccccc}
\multirow{2}{*}{{\footnotesize{}Method}} &  & \multirow{2}{*}{{\footnotesize{}Approach}} & {\footnotesize{}Outputs} & {\footnotesize{}Outputs} & {\footnotesize{}Control} & {\footnotesize{}Time} &  & {\footnotesize{}Repea-} & {\footnotesize{}Recall} & {\footnotesize{}Detection}\tabularnewline
 &  &  & {\footnotesize{}Segments} & {\footnotesize{}Score} & {\footnotesize{}\#proposals} & {\footnotesize{}(sec.)} &  & {\footnotesize{}tability} & {\footnotesize{}Results} & {\footnotesize{}Results}\tabularnewline
\hline 
\texttt{\footnotesize{}Bing }{\footnotesize{}\cite{Cheng2014Cvpr}} &  & {\footnotesize{}Window scoring} &  & \chk & \chk & 0.2 &  & $\star\star\star$ & $\star$ & $\cdot$\tabularnewline
\texttt{\footnotesize{}CPMC }{\footnotesize{}\cite{Carreira2012Pami}} &  & {\footnotesize{}Grouping} & \chk & \chk & \greychk & 250 &  & - & $\star\star$ & $\star$\tabularnewline
\texttt{\footnotesize{}EdgeBoxes }{\footnotesize{}\cite{Zitnick2014Eccv}} &  & {\footnotesize{}Window scoring} &  & \chk & \chk & 0.3 &  & $\star\star$ & $\star\star\star$ & $\star\star\star$\tabularnewline
\texttt{\footnotesize{}Endres }{\footnotesize{}\cite{Endres2014Pami}} &  & {\footnotesize{}Grouping} & \chk & \chk & \chk & 100 &  & - & $\star\star\star$ & $\star\star$\tabularnewline
\texttt{\footnotesize{}Geodesic }{\footnotesize{}\cite{Kraehenbuehl2014Eccv}} &  & {\footnotesize{}Grouping} & \chk &  & \greychk & 1 &  & $\star$ & $\star\star\star$ & $\star\star$\tabularnewline
\texttt{\footnotesize{}MCG }{\footnotesize{}\cite{Arbelaez2014Cvpr}} &  & {\footnotesize{}Grouping} & \chk & \chk & \greychk & 30 &  & $\star$ & $\star\star\star$ & $\star\star\star$\tabularnewline
\texttt{\footnotesize{}Objectness }{\footnotesize{}\cite{Alexe2012Pami}} &  & {\footnotesize{}Window scoring} &  & \chk & \chk & 3 &  & $\cdot$ & $\star$ & $\cdot$\tabularnewline
\texttt{\footnotesize{}Rahtu }{\footnotesize{}\cite{Rahtu2011Iccv}} &  & {\footnotesize{}Window scoring} &  & \chk & \chk & 3 &  & $\cdot$ & $\cdot$ & $\star$\tabularnewline
\texttt{\footnotesize{}RandomizedPrim's }{\footnotesize{}\cite{Manen2013Iccv}} &  & {\footnotesize{}Grouping} & \chk &  & \chk & 1 &  & $\star$ & $\star$ & $\star\star$\tabularnewline
\texttt{\footnotesize{}Rantalankila }{\footnotesize{}\cite{Rantalankila2014Cvpr}} &  & {\footnotesize{}Grouping} & \chk &  & \greychk & 10 &  & $\star\star$ & $\cdot$ & $\star\star$\tabularnewline
\texttt{\footnotesize{}Rigor }{\footnotesize{}\cite{Humayun2014Cvpr}} &  & {\footnotesize{}Grouping} & \chk &  & \greychk & 10 &  & $\star$ & $\star\star$ & $\star\star$\tabularnewline
\texttt{\footnotesize{}SelectiveSearch }{\footnotesize{}\cite{Uijlings2013Ijcv}} &  & {\footnotesize{}Grouping} & \chk & \chk & \greychk & 10 &  & $\star\star$ & $\star\star\star$ & $\star\star\star$\tabularnewline
\hline 
\texttt{\footnotesize{}Gaussian} &  &  &  &  & \chk & 0 &  & $\cdot$ & $\cdot$ & $\star$\tabularnewline
\texttt{\footnotesize{}SlidingWindow} &  &  &  &  & \chk & 0 &  & $\star\star\star$ & $\cdot$ & $\cdot$\tabularnewline
\texttt{\footnotesize{}Superpixels} &  &  & \chk &  &  & 1 &  & $\star$ & $\cdot$ & $\cdot$\tabularnewline
\texttt{\footnotesize{}Uniform} &  &  &  &  & \chk & 0 &  & $\cdot$ & $\cdot$ & $\cdot$\tabularnewline
\end{tabular}
\par\end{centering}

\protect\caption{\label{tab:methods-comparison}Comparison of different detection proposal
methods. Grey check-marks indicate that the number of proposals is
controlled by indirectly adjusting parameters. Repeatability, quality,
and detection rankings are provided as rough summary of the experimental
results: ``-'' indicates no data, ``$\cdot$'', ``$\star$'',
``$\star\star$'', ``$\star\star\star$'' indicate progressively
better results. These guidelines were obtained based on experiments
presented in sections \S\ref{sec:Repeatibility}, \S\ref{sec:Proposal-recall},
and \S\ref{sec:Using-detection-proposals}, respectively.}
\end{table*}
Detection proposals are similar in spirit to interest point detectors
\cite{Tuytelaars2008FoundationsandTrends,Mikolajczyk2005Ijcv}. Interest
points allow for focusing attention to the most salient and distinctive
locations in an image, greatly reducing computation for subsequent
tasks such as classification, retrieval, matching, and detection.
Likewise, object proposals considerably reduce computation compared
to the dense (sliding window) detection framework by generating candidate
proposals that may contain objects. This in turn enables use of expensive
classifiers per window \cite{Wang2013Iccv,Girshick2014Cvpr,Szegedy2014arXiv,He2014Eccv,Cinbis2013Iccv}. 

It is worthwhile noting that interest points were dominant when computing
feature descriptors densely was prohibitive. However, with improved
algorithmic efficiency and increased computational power, it is now
standard practice to use dense feature extraction \cite{Tuytelaars2010Cvpr}.
The opposite trend has occurred in object detection, where the dense
sliding window framework has been overtaken by use of proposals. We
aim to understand if detection proposals improve detection accuracy
or if their use is strictly necessary for computational reasons. While
in this work we focus on the impact of proposals on detection, proposals
have applications beyond object detection, as we discuss in \S\ref{sec:Discussion-Conclusion}.

Two general approaches for generating object proposals have emerged:
\emph{grouping methods} and \emph{window scoring methods}. These are
perhaps best exemplified by the early and well known \texttt{Se\-lec\-ti\-ve\-Search
}\cite{Sande2011Iccv} and \texttt{Ob\-ject\-ness} \cite{Alexe2010Cvpr}
proposal methods. We survey these approaches in \S\ref{sub:Grouping-proposal-methods}
and \S\ref{sub:Window-scoring-proposal-methods}, followed by an
overview of alternate approaches in \S\ref{sub:Alternate-proposal-methods}
and baselines in \S\ref{sub:Baseline-proposal-methods}. Finally,
we consider the connection between proposals and cascades in \S\ref{sub:Proposals-versus-cascades}
and provide additional method details in \S\ref{sub:Controlling-the-number-of-proposals}.

The survey that follows is meant to be exhaustive. However, for the
purpose of our evaluations, we only consider methods for which source
code is available. We cover a diverse set of methods (in terms of
quality, speed, and underlying approach). Table \ref{tab:methods-comparison}
gives an overview of the 12 selected methods (plus 4 baselines).%
\footnote{We mark the evaluated methods with a `$\dagger$' in the following
listing.%
} Table \ref{tab:methods-comparison} also indicates high level information
regarding the output of each method and a qualitative overview of
the results of the evaluations performed in the remainder of this
paper.

In this paper we concentrate on class-agnostic proposals for single-frame,
bounding box detection. For proposal methods that output segmentations
instead of bounding boxes, we convert the output to bounding boxes
for the purpose of our evaluation. Methods that operate on videos
and require temporal information (e.g.~\cite{Fragkiadaki2015Cvpr})
are considered outside the scope of this work.

\subsection{\label{sub:Grouping-proposal-methods}Grouping proposal methods}

Grouping proposal methods attempt to generate multiple (possibly overlapping)
segments that are likely to correspond to objects. The simplest such
approach would be to directly use the output of any hierarchical image
segmentation algorithm, e.g.~Gu et al.~\cite{Gu2009Cvpr} use the
segmentation produced by gPb \cite{Arbelaez2011Pami}. To increase
the number of candidate segments, most methods attempt to diversify
such hierarchies, e.g.~by using multiple low level segmentations
\cite{Carreira2012Pami,Uijlings2013Ijcv,Manen2013Iccv} or starting
with an over-segmentation and randomising the merge process \cite{Manen2013Iccv}.
The decision to merge segments is typically based on a diverse set
of cues including superpixel shape, appearance cues, and boundary
estimates (typically obtained from \cite{Arbelaez2011Pami,Dollar2015Pami}).

We classify grouping methods into three types according to how they
generate proposals. Broadly speaking, methods generate region proposals
by grouping superpixels (SP), often using \cite{Felzenszwalb2004IJCV},
solving multiple graph cut (GC) problems with diverse seeds, or directly
from edge contours (EC), e.g. from~\cite{Arbelaez2011Pami,Dollar2015Pami}.
In the method descriptions below the type of each method is marked
by SP, GC, or EC accordingly.

We note that while all the grouping approaches have the strength of
producing a segmentation mask of the object, we evaluate only the
enclosing bounding box proposals. 
\begin{itemize}
\item \texttt{\textbf{SelectiveSearch}}$^{\dagger SP}$\texttt{\,}\cite{Sande2011Iccv,Uijlings2013Ijcv}
greedily merges superpixels to generate proposals. The method has
no learned parameters, instead features and similarity functions for
merging superpixels are manually designed. \texttt{Se\-lec\-tive\-Search}
has been broadly used as the proposal method of choice by many state-of-the-art
object detectors, including the R-CNN and Fast R-CNN detectors\cite{Girshick2014Cvpr,Girshick2015arXiv}. 
\item \noindent \texttt{\textbf{RandomizedPrim's}}$^{\dagger SP}$\texttt{\,}\cite{Manen2013Iccv}
uses similar features as \texttt{Se\-lec\-ti\-ve\-Search}, but
introduces a randomised superpixel merging process in which all probabilities
have been learned. Speed is substantially improved.
\item \noindent \texttt{\textbf{Rantalankila}}$^{\dagger SP}$\texttt{\,}\cite{Rantalankila2014Cvpr}
proposes a superpixel merging strategy similar to \texttt{Se\-lec\-ti\-ve\-Search},
but using different features. In a subsequent stage, the generated
segments are used as seeds for solving graph cuts in the spirit of
\texttt{CPMC} (see below) to generate more proposals.
\item \texttt{\textbf{Chang}}$\,{}^{SP}$\,\cite{Chang2011Iccv} combines
saliency and \texttt{Ob\-ject\-ness} with a graphical model to merge
superpixels into figure/background segmentations.
\item \noindent \texttt{\textbf{CPMC}}$^{\dagger GC}$\texttt{\,}\cite{Carreira2010Cvpr,Carreira2012Pami}
avoids initial segmentations and computes graph cuts with several
different seeds and unaries directly on pixels. The resulting segments
are ranked using a large pool of features.
\item \noindent \texttt{\textbf{Endres}}$^{\dagger GC}$\texttt{\,}\cite{Endres2010Eccv,Endres2014Pami}
builds a hierarchical segmentation from occlusion boundaries and solves
graph cuts with different seeds and parameters to generate segments.
The proposals are ranked based on a wide range of cues and in a way
that encourages diversity.
\item \noindent \texttt{\textbf{Rigor}}$^{\dagger GC}$\texttt{\,}\cite{Humayun2014Cvpr}
is a somewhat improved variant of \texttt{CPMC} that speeds computation
considerably by re-using computation across multiple graph-cut problems
and using the fast edge detectors from \cite{Lim2013Cvpr,Dollar2015Pami}.
\item \noindent \texttt{\textbf{Geodesic}}$^{\dagger EC}$\texttt{\textbf{\,}}\cite{Kraehenbuehl2014Eccv}
starts from an over-segmentation of the image based on \cite{Dollar2015Pami}.
Classifiers are used to place seeds for a geodesic distance transform.
Level sets of each of the distance transforms define the figure/ground
segmentations that are the proposals.
\item \noindent \texttt{\textbf{MCG}}$^{\dagger EC}$\texttt{\,}\cite{Arbelaez2014Cvpr}
introduces a fast algorithm for computing multi-scale hierarchical
segmentations building on \cite{Dollar2015Pami}. Segments are merged
based on edge strength and the resulting object hypotheses are ranked
using cues such as size, location, shape, and edge strength.
\end{itemize}

\subsection{\label{sub:Window-scoring-proposal-methods}Window scoring proposal
methods}

An alternate approach for generating detection proposals is to score
each candidate window according to how likely it is to contain an
object. Compared to grouping approaches these methods usually only
return bounding boxes and tend to be faster. Unless window sampling
is performed very densely, this approach typically generates proposals
with low localisation accuracy. Some methods counteract this by refining
the location of the generated windows.
\begin{itemize}
\item \noindent \texttt{\textbf{Objectness}}$^{\dagger}$\texttt{\,}\cite{Alexe2010Cvpr,Alexe2012Pami}
is one of the earliest and well known proposal methods. An initial
set of proposals is selected from salient locations in an image, these
proposals are then scored according to multiple cues including colour,
edges, location, size, and the strong ``superpixel straddling''
cue.
\item \noindent \texttt{\textbf{Rahtu}}$^{\dagger}$\texttt{\,}\cite{Rahtu2011Iccv}
begins with a large pool of proposal regions generated from individual
superpixels, pairs and triplets of superpixels, and multiple randomly
sampled boxes. The scoring strategy used by \texttt{Ob\-ject\-ness}
is revisited, and improvements are proposed. \cite{Blaschko2013Scia}
adds additional low-level features and highlights the importance of
properly tuned non-maximum suppression.
\item \noindent \texttt{\textbf{Bing}}$^{\dagger}$\texttt{\,}\cite{Cheng2014Cvpr}
uses a simple linear classifier trained over edge features and applied
in a sliding window manner. Using adequate approximations a very fast
class agnostic detector is obtained (1 ms/image on CPU). However,
it was shown that the classifier has minimal influence and similar
performance can be obtained \emph{without} looking at the image \cite{Zhao2014Bmvc}.
This image independent method is named \texttt{Cracking\-Bing}\texttt{\textbf{.}}
\item \noindent \texttt{\textbf{EdgeBoxes}}$^{\dagger EC}$\texttt{\textbf{\,}}\cite{Zitnick2014Eccv}
also starts from a coarse sliding window pattern, but builds on object
boundary estimates (obtained via structured decision forests \cite{DollarICCV13edges,Dollar2015Pami})
and adds a subsequent refinement step to improve localisation. No
parameters are learned. The authors propose tuning the density of
the sliding window pattern and the threshold of the non-maximum suppression
to tune the method for different overlap thresholds (see \S\ref{sec:Using-detection-proposals}).
\item \texttt{\textbf{Feng}}\,\cite{Feng2011Iccv} poses proposal generation
as the search for salient image content and introduces new saliency
measures, including the ease with which a potential object can be
composed from the rest of the image. The sliding window paradigm is
used and every location scored according to the saliency cues.
\item \texttt{\textbf{Zhang}}\,\cite{Zhang2011Cvpr} proposes to train
a cascade of ranking SVMs on simple gradient features. The first stage
has separate classifiers for each scale and aspect ratio; the second
stage ranks all proposals from the previous stage. All SVMs are trained
using structured output learning to score windows higher that overlap
more with objects. Because the cascade is trained and tested over
the same set of categories, it is unclear how well this approach generalises
across categories.
\item \texttt{\textbf{RandomizedSeeds}}\,\cite{Bergh2013Iccv} uses multiple
randomised SEED superpixel maps \cite{Bergh2014Ijcv} to score each
candidate window. The scoring is done using a simple metric similar
to ``superpixel straddling'' from \texttt{Ob\-ject\-ness}, no
additional cues are used. The authors show that using multiple superpixel
maps significantly improves recall.
\end{itemize}

\subsection{\noindent \label{sub:Alternate-proposal-methods}Alternative proposal
methods}
\begin{itemize}
\item \texttt{\textbf{ShapeSharing}}\,\cite{Kim2012Eccv} is a non-parametric,
data-driven method that transfers object shapes from exemplars into
test images by matching edges. The resulting regions are subsequently
merged and refined by solving graph cuts.
\item \noindent \texttt{\textbf{Multibox}}\,\cite{Szegedy2014arXiv,Erhan2014Cvpr}
trains a neural network to directly regress a fixed number of proposals
in the image without sliding the network over the image. Each of the
proposals has its own location bias to diversify the location of the
proposals. The authors report top results on ImageNet.
\end{itemize}

\subsection{\label{sub:Baseline-proposal-methods}Baseline proposal methods}

We additionally consider a set of baselines that serve as reference
points. Like all evaluated methods described earlier, the following
baselines are class independent:
\begin{itemize}
\item \texttt{\textbf{Uniform}}$^{\dagger}$: To generate proposals, we
uniformly sample the bounding box centre position, square root area,
and log aspect ratio. We estimate the range of these parameters on
the PASCAL VOC 2007 training set after discarding 0.5\% of the smallest
and largest values, so that our estimated distribution covers 99\%
of the data.
\item \texttt{\textbf{Gaussian}}$^{\dagger}$: Likewise, we estimate a multivariate
Gaussian distribution for the bounding box centre position, square
root area, and log aspect ratio. After calculating mean and covariance
on the training set we sample proposals from this distribution.
\item \texttt{\textbf{SlidingWindow}}$^{\dagger}$: We place windows on
a regular grid as is common for sliding window object detectors. The
requested number of proposals is distributed across windows sizes
(width and height), and for each window size, we place the windows
uniformly. This procedure is inspired by the implementation of \texttt{Bing}~\cite{Cheng2014Cvpr,Zhao2014Bmvc}.
\item \texttt{\textbf{Superpixels}}$^{\dagger}$: As we will show, superpixels
have an important influence on the behaviour of proposal methods.
Since five of the evaluated methods build on \cite{Felzenszwalb2004IJCV},
we use it as a baseline: each low-level segment is used as a detection
proposal. This method serves as a lower-bound on recall for methods
using superpixels.
\end{itemize}
It should be noted that with the exception of \texttt{Super\-pixels},
all the baselines generate proposal windows independent of the image
content. \texttt{Sliding\-Window} is deterministic given the image
size (similar to \texttt{Cracking\-Bing}), while the \texttt{Uni\-form}
and \texttt{Gaussian} baselines are stochastic.

\subsection{\label{sub:Proposals-versus-cascades}Proposals versus cascades}

Many proposal methods utilise image features to generate candidate
windows. One can interpret this process as a discriminative one; given
such features a method quickly determines whether a window should
be considered for detection. Indeed, many of the surveyed methods
include some form of discriminative learning (\texttt{Selective\-Search}
and \texttt{Edge\-Boxes} are notable exceptions). As such, proposal
methods are related to cascades \cite{Viola2004Ijvc,Bourdev2005Cvpr,Harzallah2009Iccv,Dollar2012Eccv},
which use a fast but inaccurate classifier to discard a vast majority
of unpromising proposals. Although traditionally used for class specific
detection, cascades can also apply to sets of categories \cite{Torralba2007Pami,Zehnder2008Bmvc}.

The key distinction between traditional cascades and proposal methods
is that the latter is required to generalise beyond object classes
observed during training. So what allows discriminatively trained
proposal methods to generalise to unseen categories? A key assumption
is that training a classifier for a large enough number of categories
is sufficient to generalise to unseen categories (for example, after
training on cats and dogs proposals may generalise to other animals).
Additionally, the discriminative power of the classifier is often
limited (e.g. \texttt{Bing} and \texttt{Zhang}), thus preventing overfitting
to the training classes and forcing the classifier to learn coarse
properties shared by all object (e.g. ``objects are roundish'').
This key distinction is also noted in \cite{Chavali2015arXiv}. We
test the generalisation of proposal methods by evaluating on datasets
with many additional classes in \S\ref{sec:Proposal-recall}.

\subsection{\label{sub:Controlling-the-number-of-proposals}Controlling the number
of proposals}

In this work we will perform an extensive apples-to-apples comparison
of the 12 methods (plus 4 baselines) listed in table \ref{tab:methods-comparison}.
In order to be able to compare amongst methods, for each method we
need to control the number of proposals produced per image. By default,
the evaluated methods provide variable numbers of detection proposals,
ranging from just a few ($\sim\negmedspace10^{2}$) to a large number
($\sim\negmedspace10^{5}$). Additionally, some methods output sorted
or scored proposals, while others do not. Having more proposals increases
the chance for high recall, thus for each method in all experiments
we attempt to carefully control the number of generated proposals.
Details are provided next.

Albeit not all having explicit control over the number of proposals,
\texttt{Objectness}, \texttt{CPMC}, \texttt{Endres}, \texttt{Selective
Search}, \texttt{Rahtu}, \texttt{Bing}, \texttt{MCG}, and \texttt{EdgeBoxes}
do provide scored or sorted proposals so we can use the top $k$.
\texttt{Rantalankila}, \texttt{Rigor}, and \texttt{Geodesic} provide
neither direct control over the number of proposals nor sorted proposals,
but indirect control over $k$ can be obtained by altering other parameters.
Thus, we record the number of produced proposals on a subset of the
images for different parameters and linearly interpolate between the
parameter settings to control $k$. For \texttt{RandomizedPrim's},
which lacks any control over the number of proposals, we randomly
sample $k$ proposals.

Finally, we observed a number of methods produce duplicate proposals.
All such duplicates were removed.

\section{\label{sec:Repeatibility}Proposal repeatability}

Training a detector on detection proposals rather than on all sliding
windows modifies the appearance distribution of both positive and
negative windows. In section \ref{sec:Proposal-recall}, we look into
how well the different object proposals overlap with ground truth
annotations of objects, which is an analysis of the positive window
distribution. In this section we analyse the distribution of negative
windows: if the proposal method does not consistently propose windows
on similar image content without objects or with partial objects,
the classifier may have difficulty generating scores on negative windows
on the test set. As an extreme, motivational example, consider a proposal
method that generates proposals containing only objects on the training
set but containing both objects and negative windows on the test set.
A classifier trained on such proposals would be unable to differentiate
objects from background, thus at test time would give useless scores
for the negative windows. Thus we expect that a consistent appearance
distribution for proposals \textit{on the background} is likewise
relevant for a detector.

\begin{figure}
\begin{centering}
\subfloat[\label{fig:rotation-20-degrees}Example rotation of $20{}^{\circ}$.]{\begin{centering}
\includegraphics[width=0.48\columnwidth]{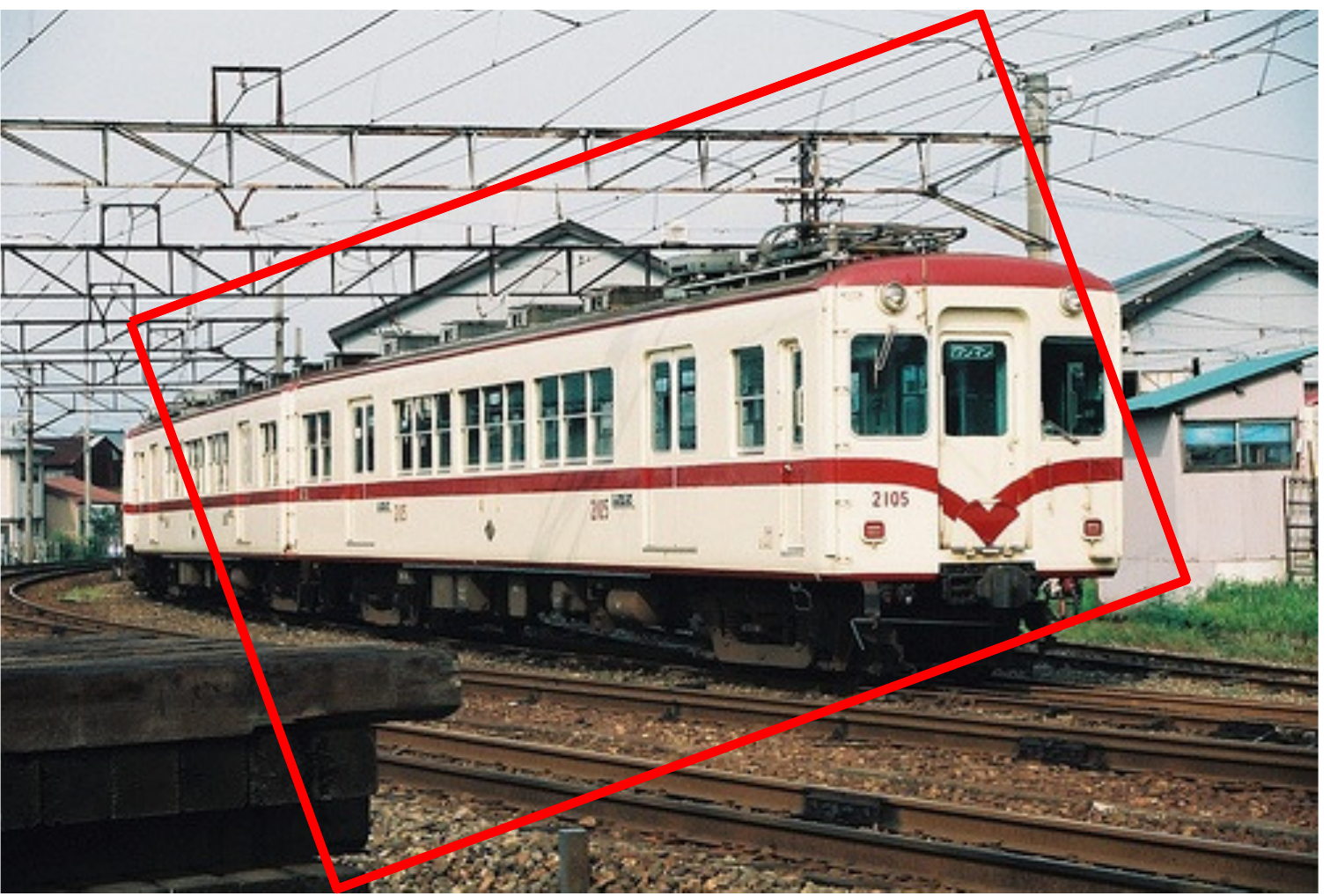}
\par\end{centering}

}\subfloat[\label{fig:rotation-20-degrees-crop}Resulting crop from (a).]{\begin{centering}
\includegraphics[width=0.48\columnwidth]{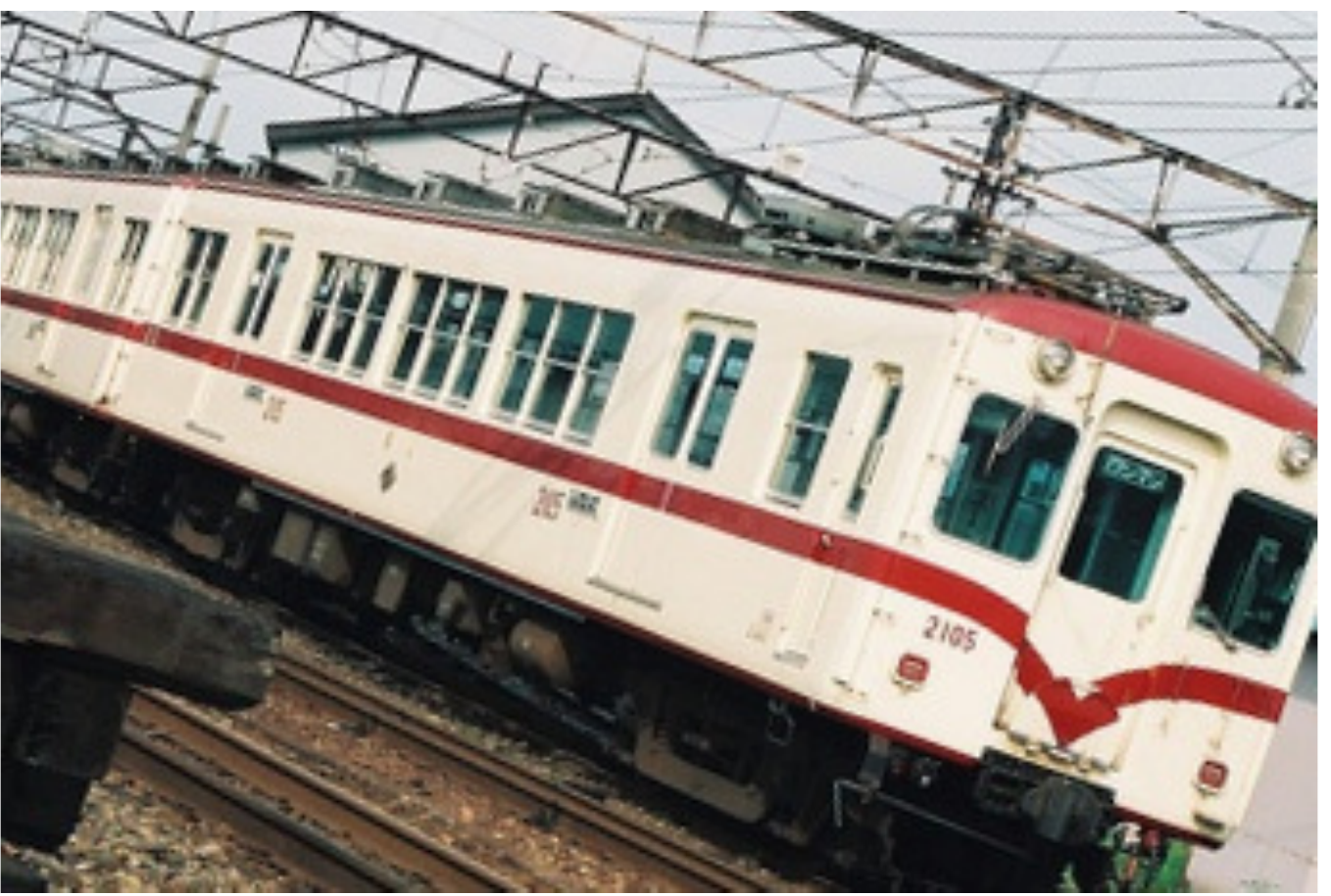}
\par\end{centering}

}
\par\end{centering}

\begin{centering}
\subfloat[\label{fig:rotation-5-degrees}Example rotation of $-5{}^{\circ}$.]{\begin{centering}
\includegraphics[width=0.48\columnwidth]{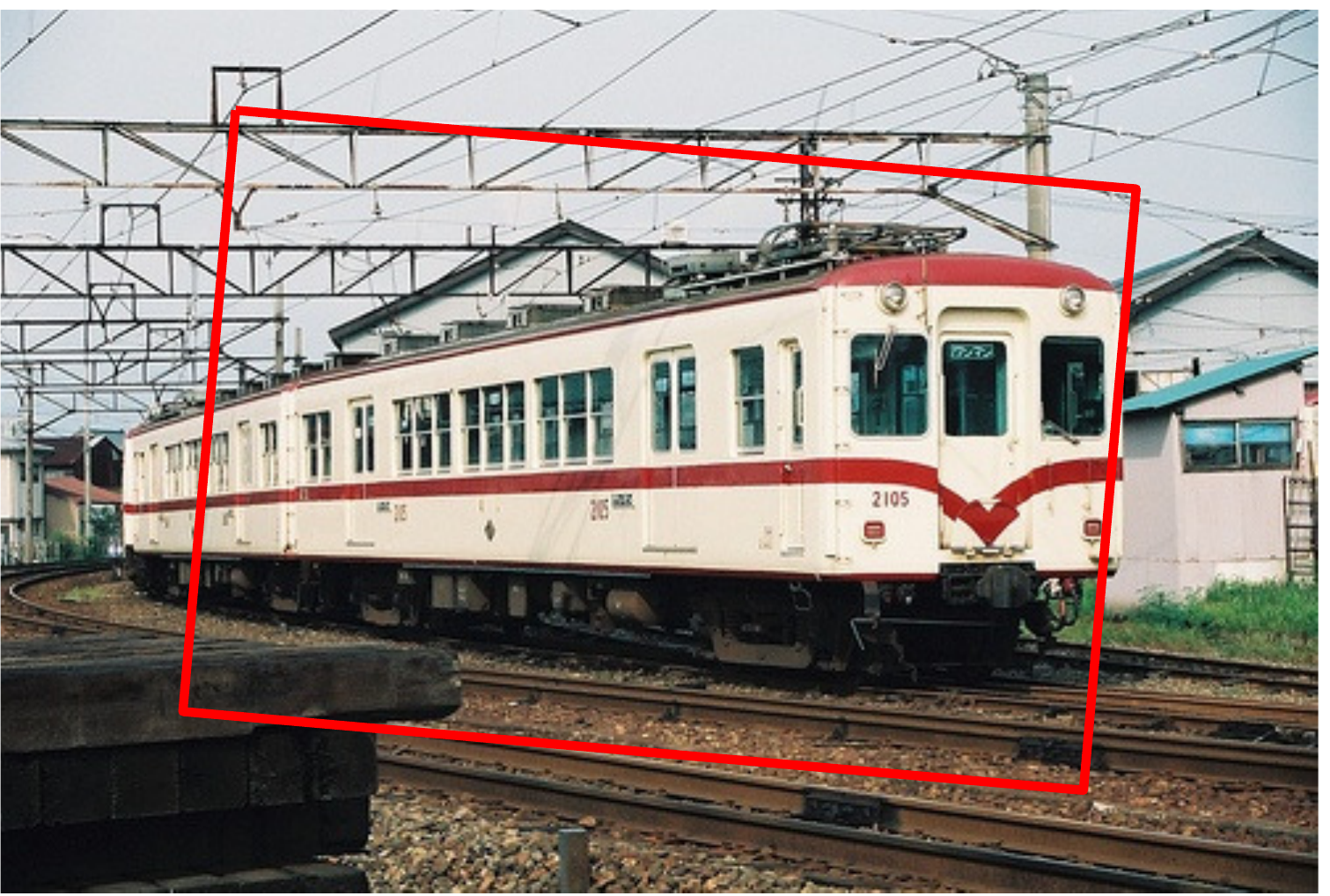}
\par\end{centering}

}\subfloat[\label{fig:rotation-5-degrees-crop}Resulting crop from (c).]{\begin{centering}
\includegraphics[width=0.48\columnwidth]{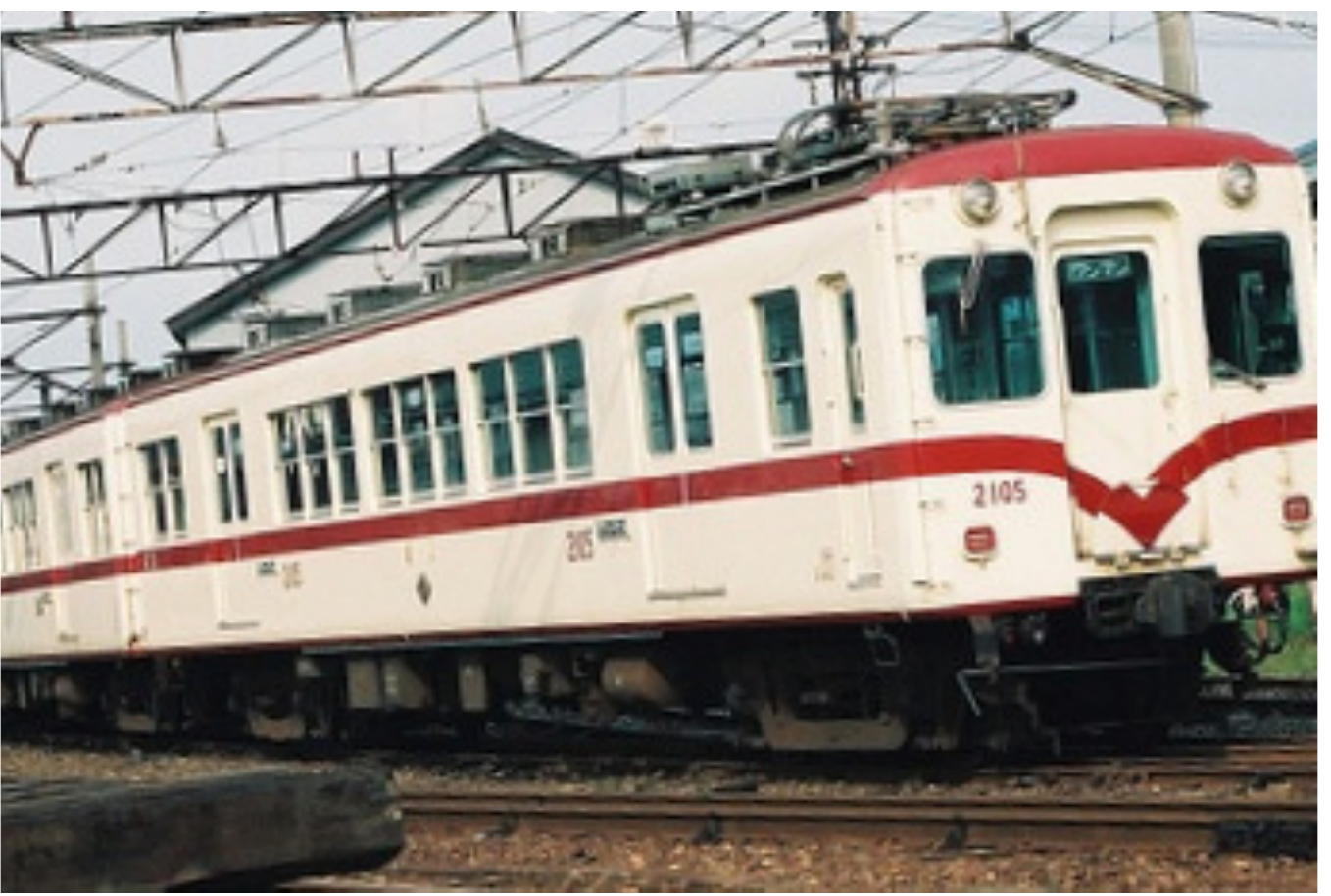}
\par\end{centering}

}
\par\end{centering}

\vspace{0.5em}

\protect\caption{\label{fig:rotation-examples}Examples of rotation perturbation. (a)
shows the largest rectangle with the same aspect as the original image
that can fit into the image under a $20{}^{\circ}$ rotation, and
(b) the resulting crop. All other rotations are cropped to the same
dimensions, e.g. the $-5{}^{\circ}$ rotation in (c) to the crop in
(d).}
\end{figure}
\begin{figure}[t]
\centering{}\includegraphics[width=0.82\columnwidth]{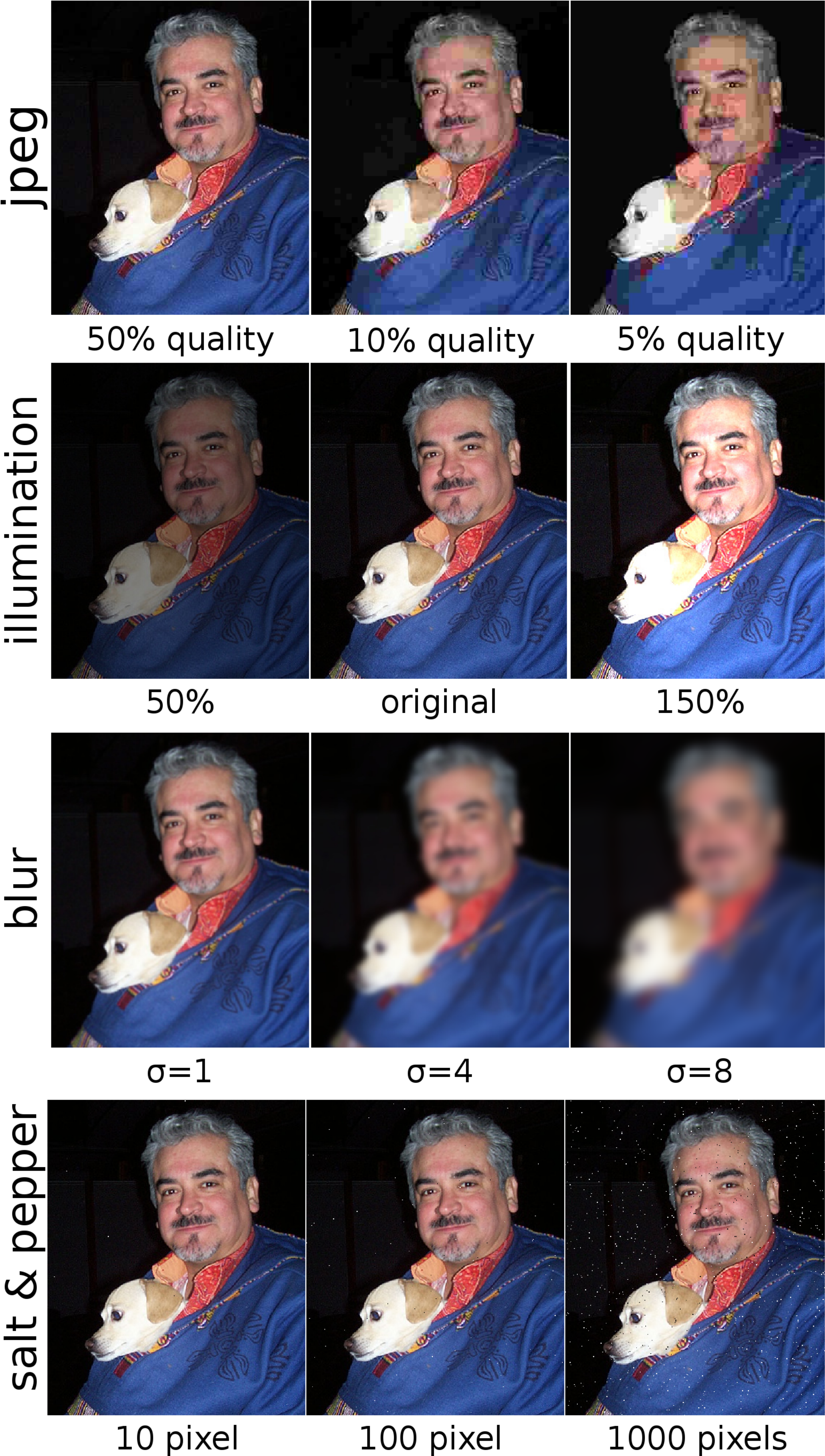}\protect\caption{\label{fig:pertubations-range}Illustration of the perturbation ranges
used for the repeatability experiments.}
\end{figure}
We call the property of proposals being placed on similar image content
the \emph{repeatability} of a proposal method. Intuitively proposals
should be repeatable on slightly different images with the same content.
To evaluate repeatability we compare proposals that are generated
for one image with proposals generated for a slightly modified version
of the same image. PASCAL VOC \cite{Everingham2014Ijcv} does not
contain suitable images. An alternative is the dataset of \cite{Mikolajczyk2005Ijcv},
but it only consists of 54 images and even fewer objects. Instead,
we opt to apply synthetic transformations to PASCAL images.

\subsection{Evaluation protocol for repeatability}

\begin{figure*}
\hspace*{\fill}\includegraphics[width=0.12\textwidth]{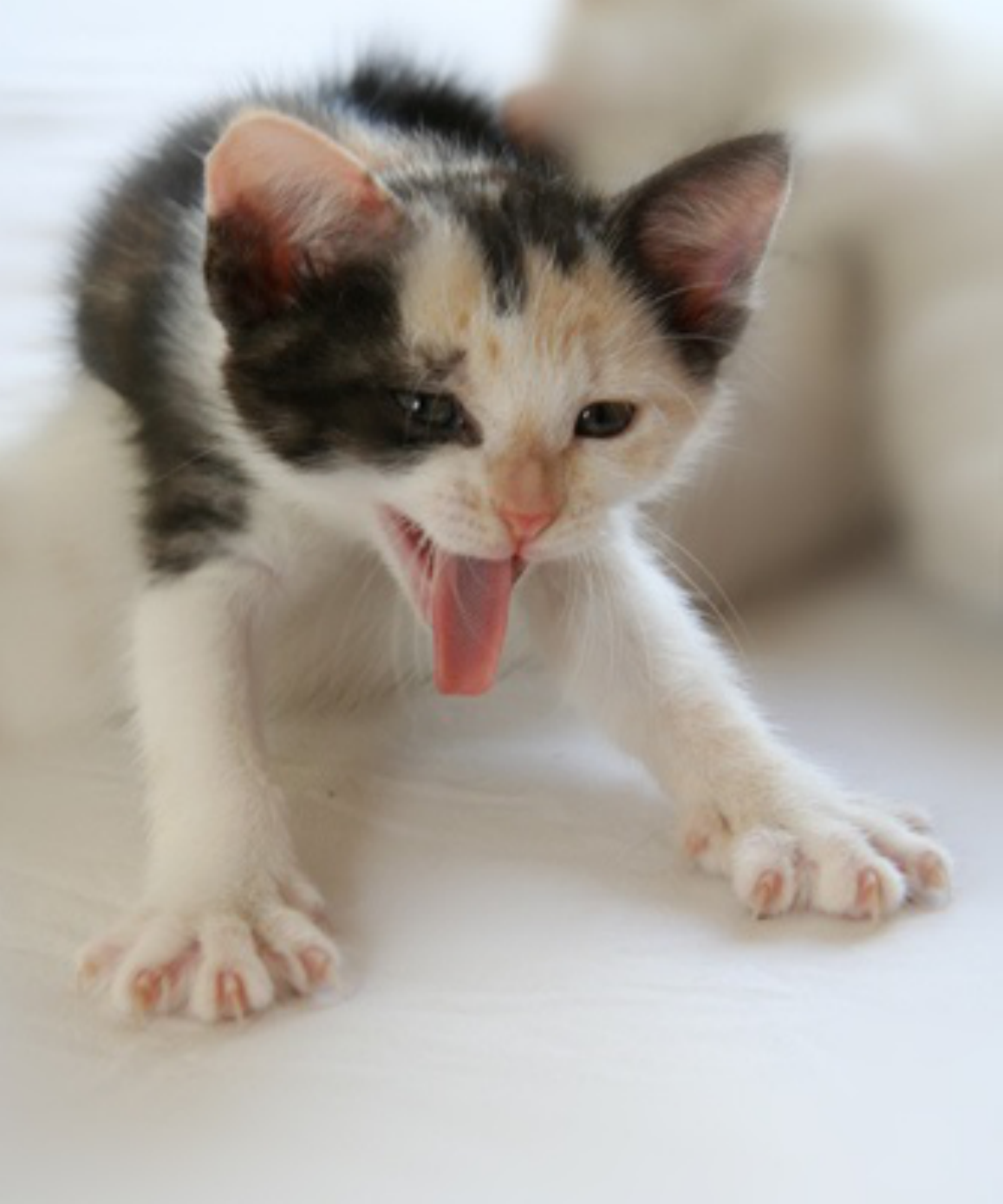}\hspace*{\fill}\includegraphics[width=0.12\textwidth]{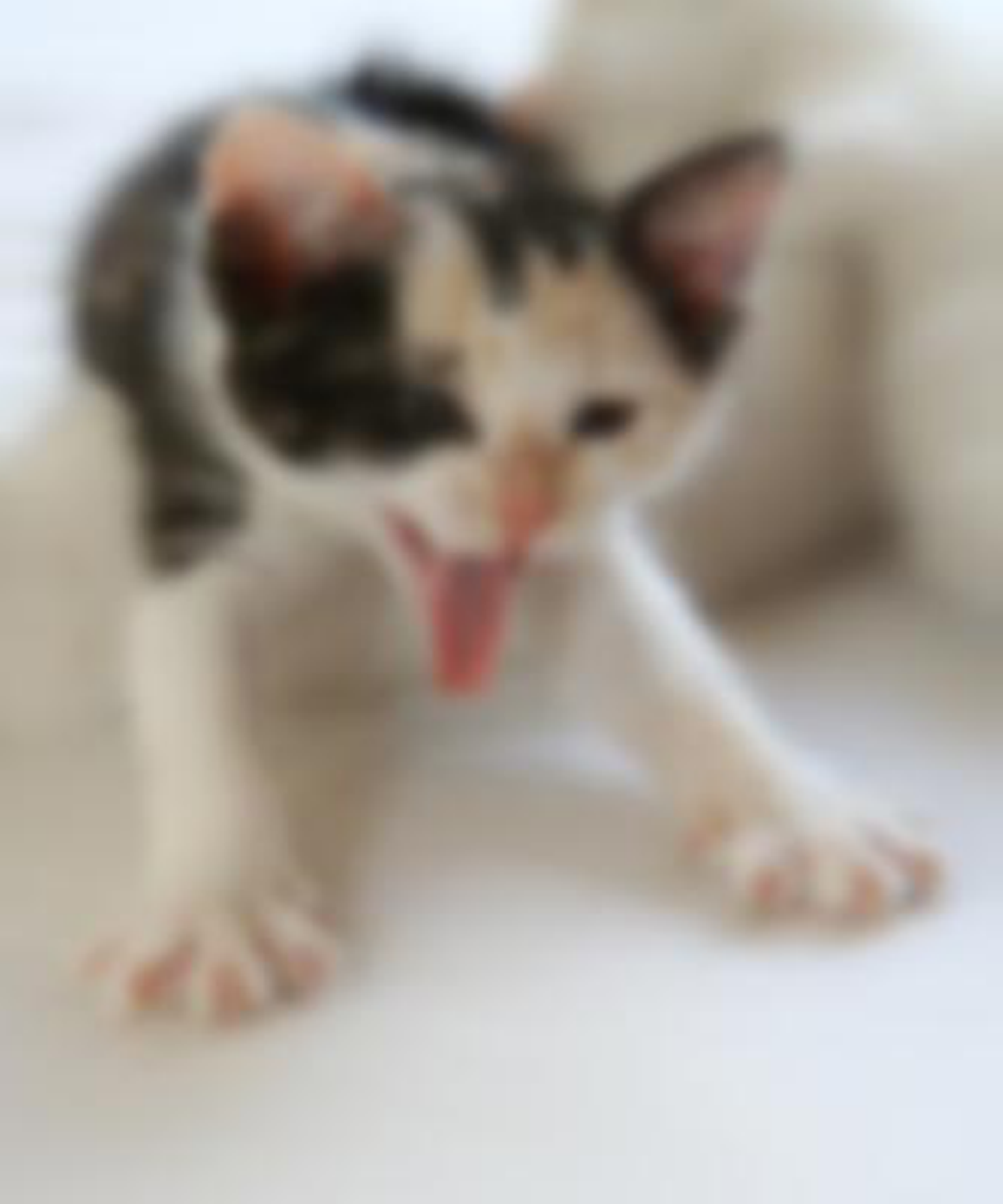}\hspace*{\fill}\includegraphics[width=0.12\textwidth]{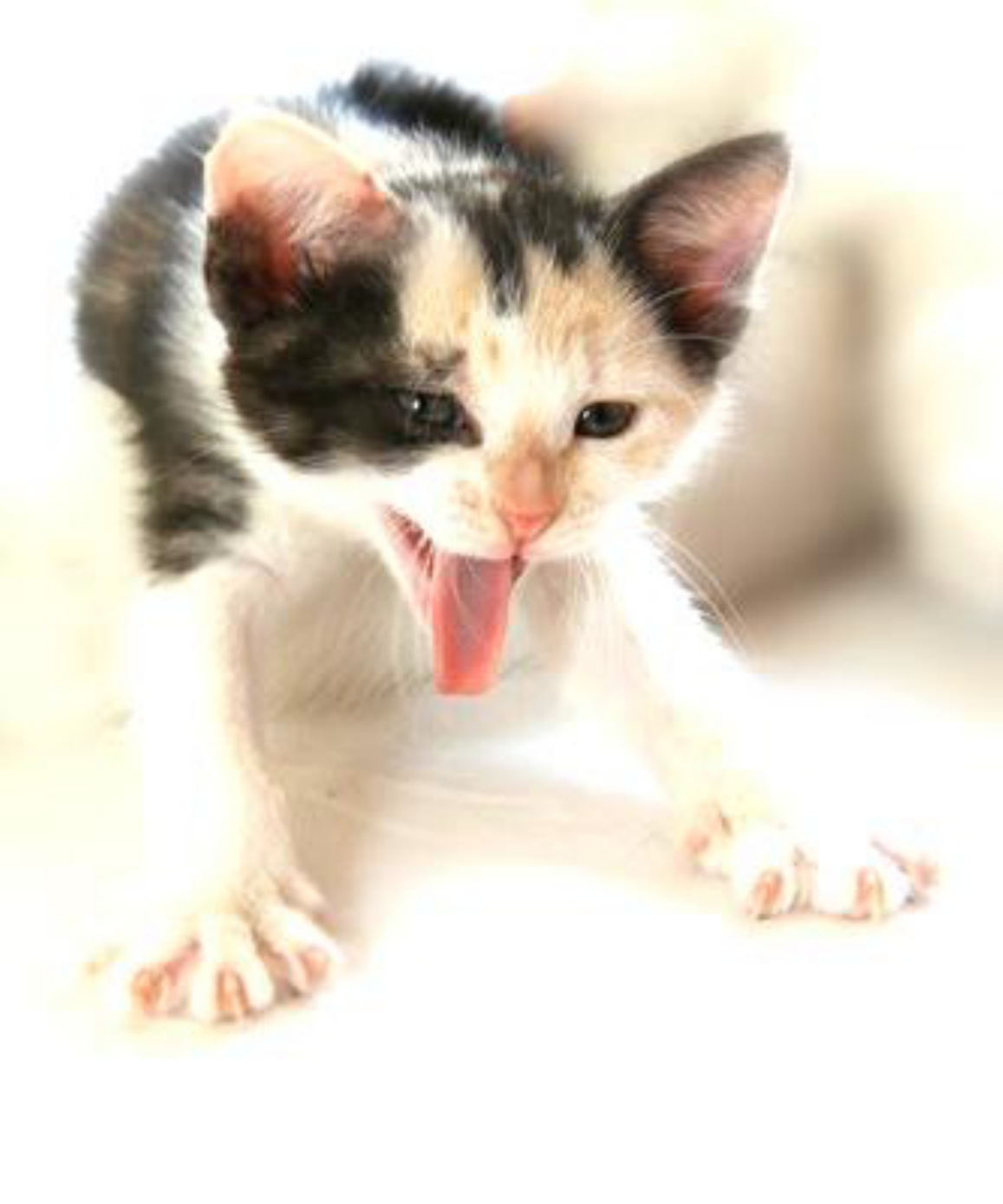}\hspace*{\fill}\includegraphics[width=0.12\textwidth]{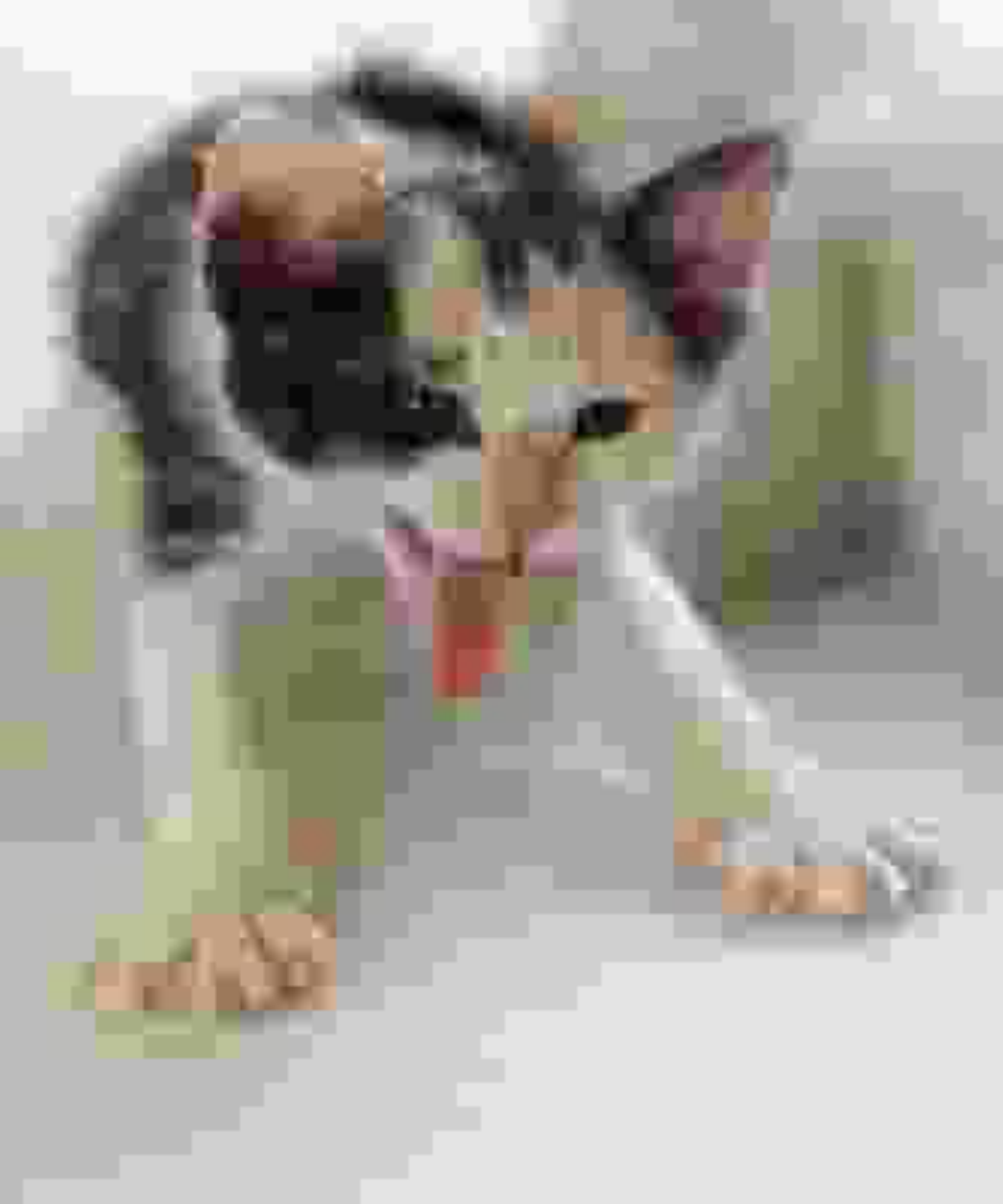}\hspace*{\fill}\includegraphics[width=0.12\textwidth]{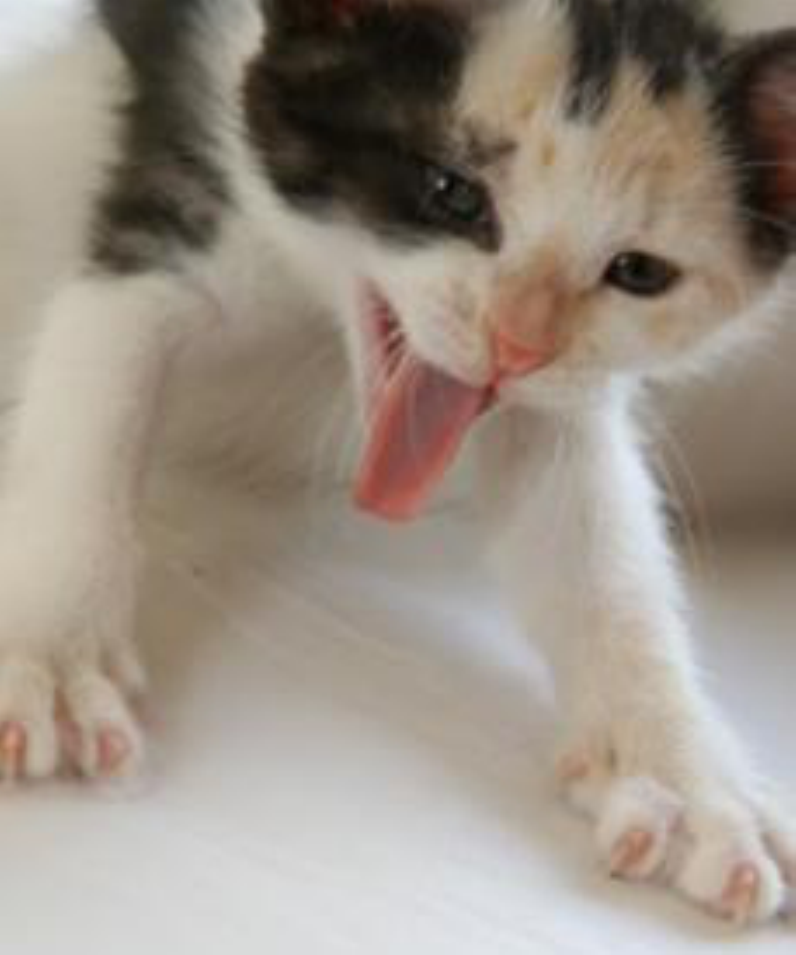}\hspace*{\fill}\includegraphics[width=0.12\textwidth]{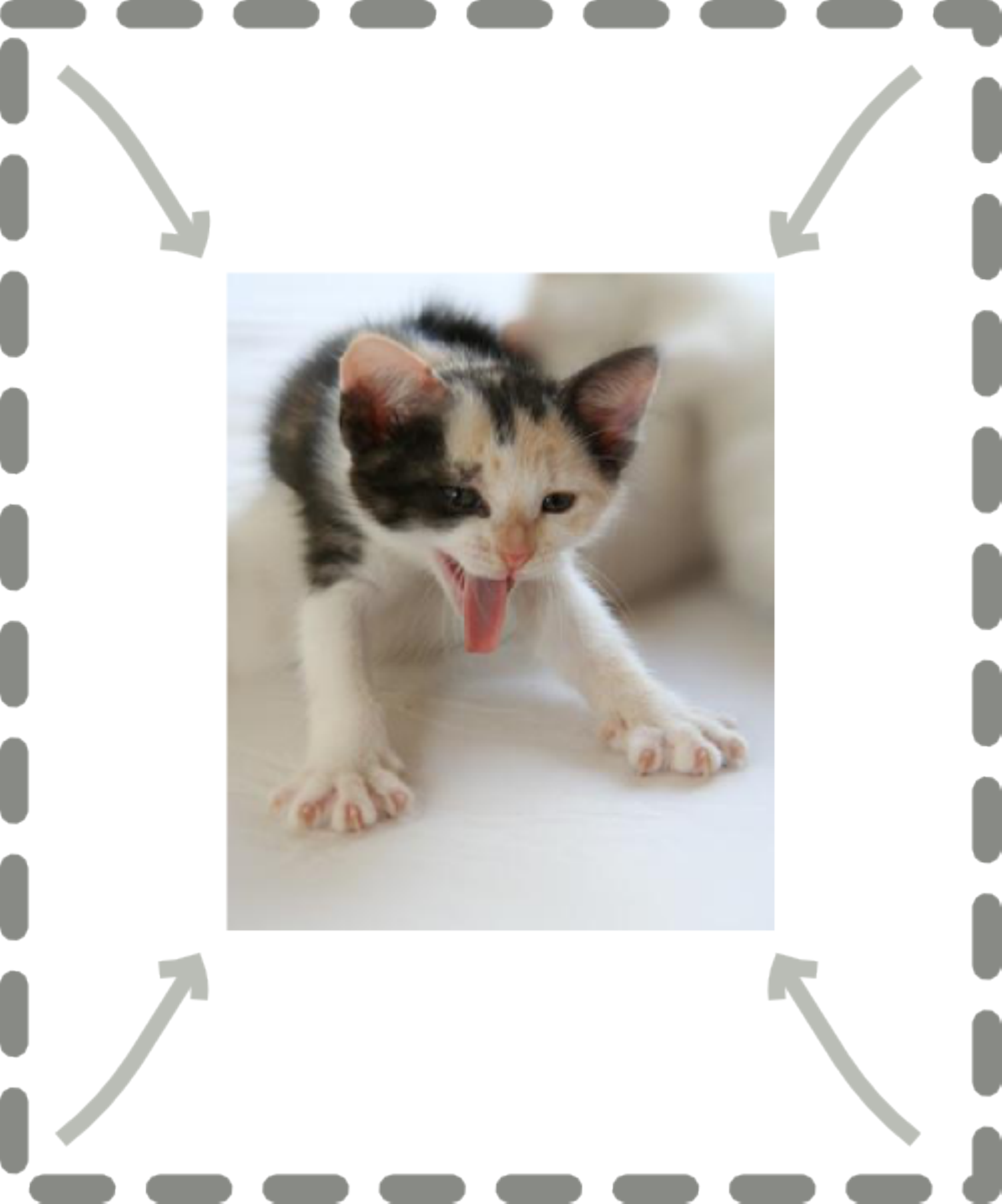}\hspace*{\fill}\includegraphics[width=0.12\textwidth]{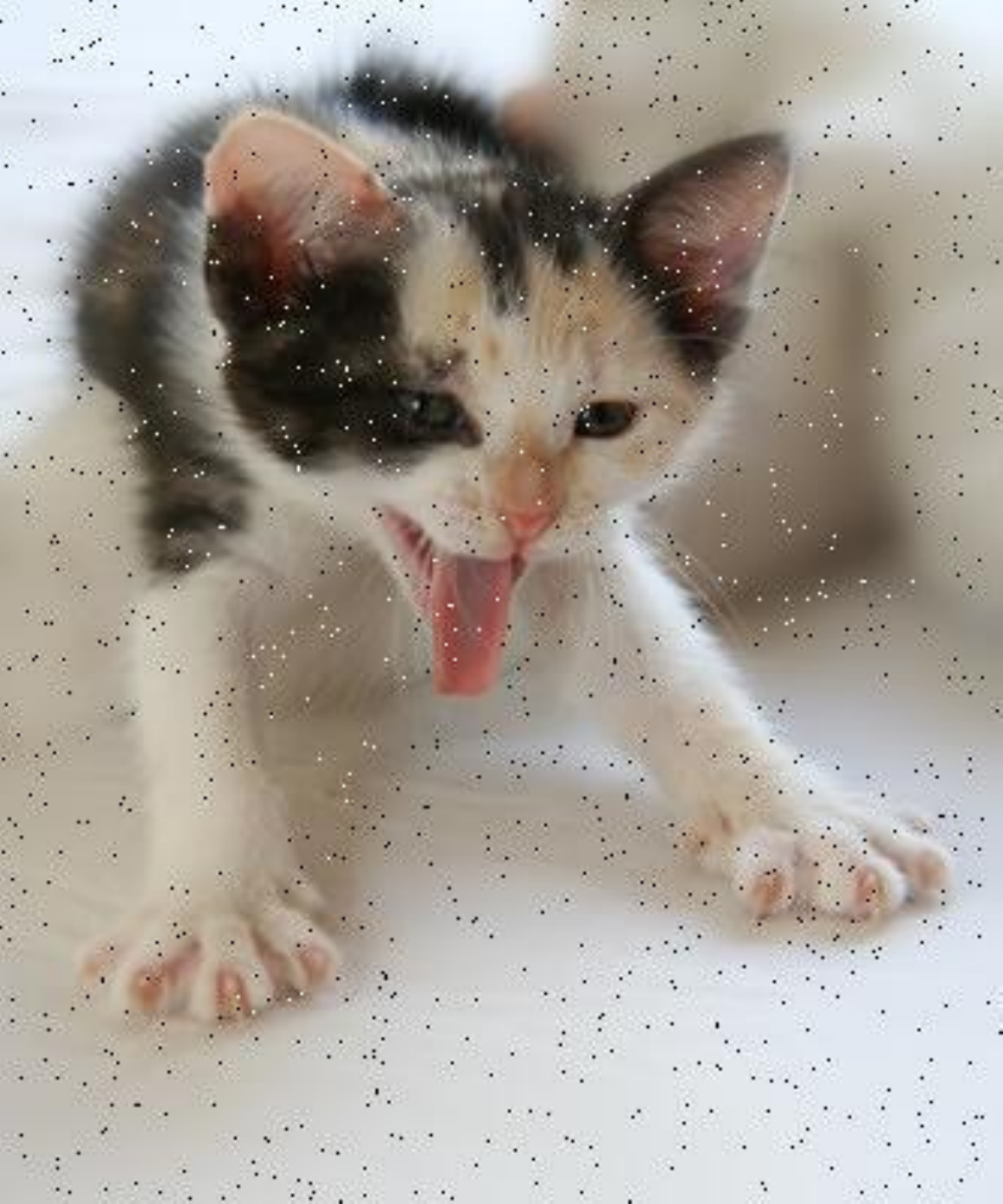}\hspace*{\fill}

\protect\caption{\label{fig:perturbations-example}Example of the image perturbations
considered. Top to bottom, left to right: original, blur, illumination,
JPEG artefact, rotation, scale perturbations, and ``salt and pepper''
noise.}
\end{figure*}

Our evaluation protocol is inspired by \cite{Mikolajczyk2005Ijcv},
which evaluates interest point repeatability. For each image in the
PASCAL VOC 2007 test set \cite{Everingham2014Ijcv}, we generate several
perturbed versions. We consider blur, rotation, scale, illumination,
JPEG compression, and \textquotedblleft salt and pepper\textquotedblright{}
noise (see figures~\ref{fig:pertubations-range}-\ref{fig:perturbations-example}). 

For each pair of reference and perturbed images we compute detection
proposals with a given method (generating $1000$ windows per image).
The proposals are projected back from the perturbed into the reference
image and then matched to the proposals in the reference image. In
the case of rotation, all proposals whose centre lies outside the
image after projection are removed before matching. For matching we
use the intersection over union (IoU) criterion and we solve the resulting
bipartite matching problem greedily for efficiency reasons. Given
the matching, we plot the recall for every IoU threshold and \emph{define
the repeatability to be the area under this ``recall versus IoU threshold''
curve}\textit{ between IoU $0$ and $1$}%
\footnote{In contrast to the average recall (AR) used in later sections, we
use the area under the entire curve. We are interested in how much
proposals change, which is independent of the PASCAL overlap criterion.%
}. This is similar to computing the average best overlap (ABO, see
\S\ref{sec:additional-metrics}) for the proposals on the reference
image. Methods that propose windows at similar locations at high IoU---and
thus on similar image content---are more repeatable, since the area
under the curve is larger.

One issue regarding such proposal matching is that large windows are
more likely to match than smaller ones since the same perturbation
will have a larger relative effect on smaller windows. This effect
is important to consider since different methods have very different
distributions of proposal window sizes as can be seen in figure~\ref{fig:distribution-of-windows-sizes}.
To reduce the impact of this effect, we bin the original image windows
by area into 10 groups, and evaluate the area under the recall versus
IoU curve per size group. In figure~\ref{fig:fluctuation-per-size-group}
we show the recall versus IoU curve for a small blur perturbation
for each of the 10 groups. As expected, large proposals have higher
repeatability. In order to measure repeatability independently of
the distribution of windows sizes, in all remaining repeatability
experiments in figure \ref{fig:repeatability} we show the (unweighted)
average across the 10 size groups.

We omit the slowest two methods, \texttt{CPMC} and \texttt{Endres},
due to computational constraints (these experiments require running
the detectors \textasciitilde{}50 times on the entire PASCAL test
set, once for every perturbation).

\subsection{Repeatability experiments and results}

\begin{figure*}
\begin{centering}
\vspace{-5em}
\begin{tabular}{ccc}
\subfloat[\label{fig:distribution-of-windows-sizes}Histogram of proposal sizes
on PASCAL.]{\begin{centering}
\includegraphics[width=0.3\textwidth]{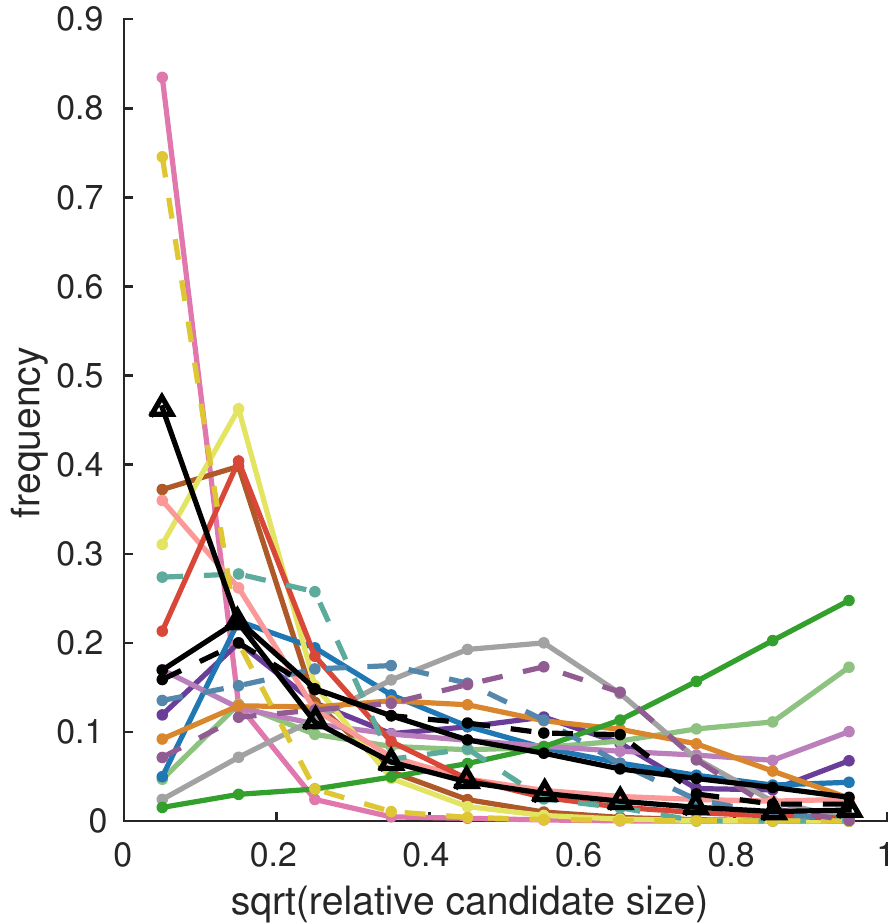}
\par\end{centering}

} & \subfloat[\label{fig:fluctuation-per-size-group}Example of recall at different
scales.]{\begin{centering}
\includegraphics[width=0.3\textwidth]{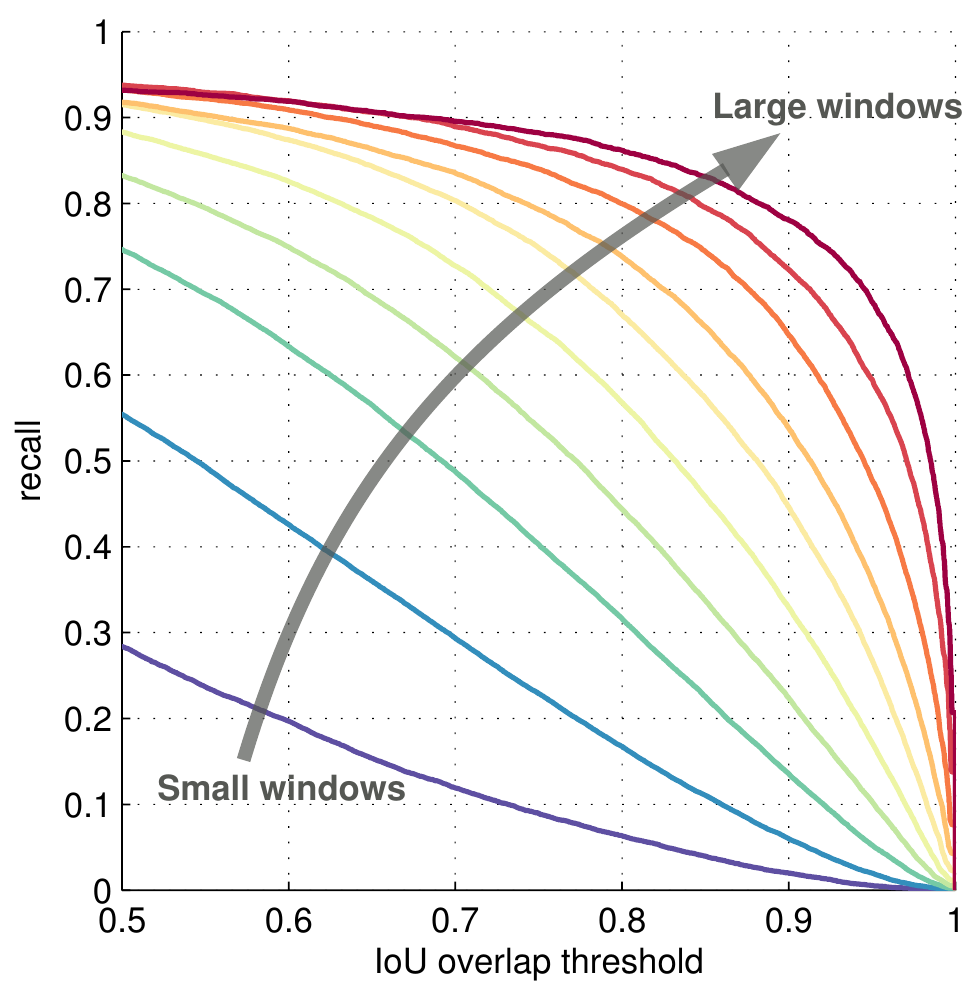}
\par\end{centering}

} & \subfloat[\label{fig:repeatiblity-scale-change}Scale.]{\begin{centering}
\includegraphics[width=0.3\textwidth]{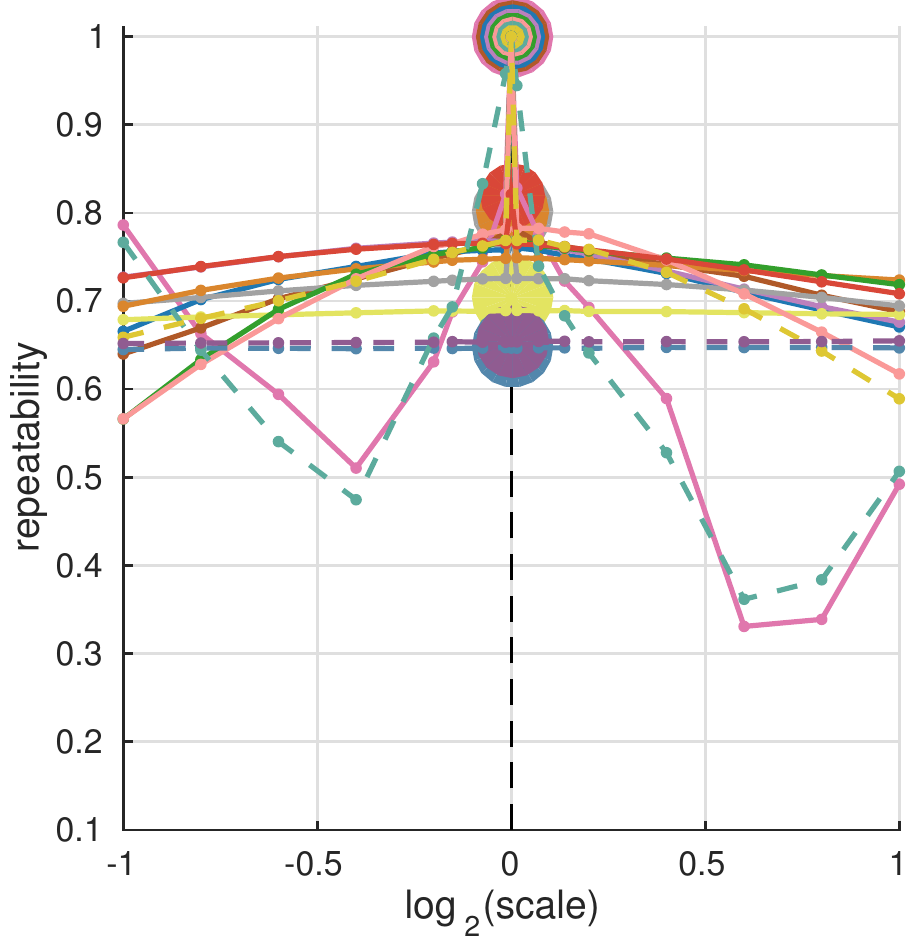}
\par\end{centering}

}\tabularnewline
\subfloat[\label{fig:repeatiblity-jpeg-change}JPEG artefacts.]{\begin{centering}
\includegraphics[width=0.3\textwidth]{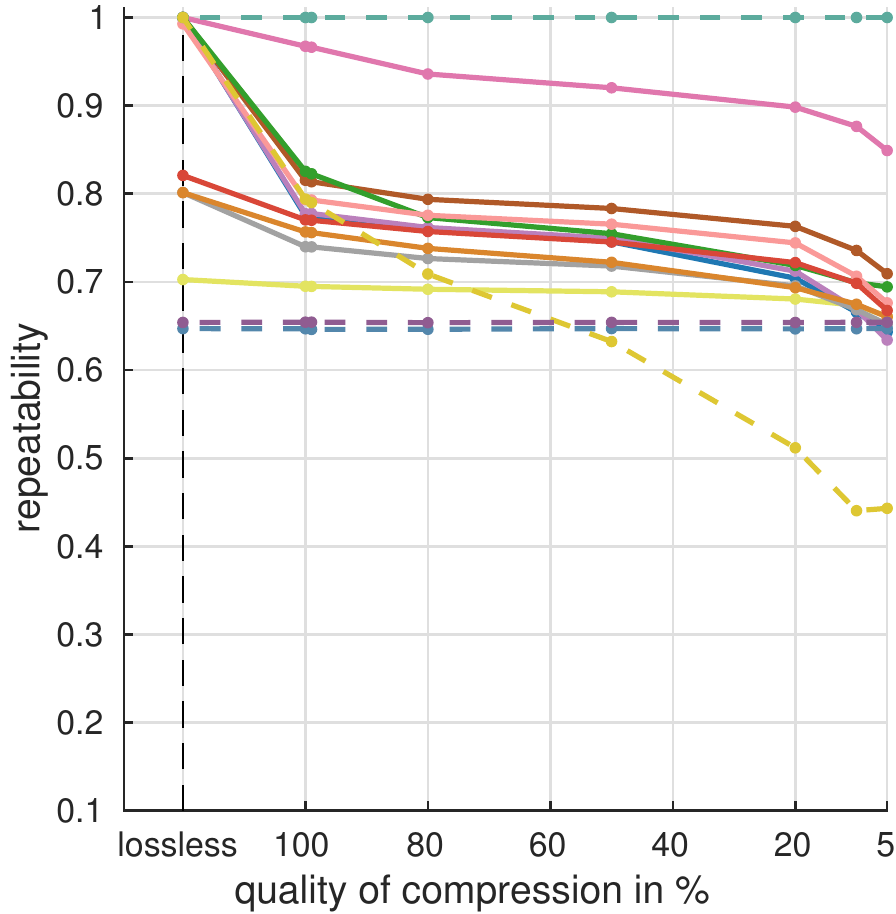}
\par\end{centering}

} & \subfloat[\label{fig:repeatiblity-rotation-change}Rotation.]{\begin{centering}
\includegraphics[width=0.3\textwidth]{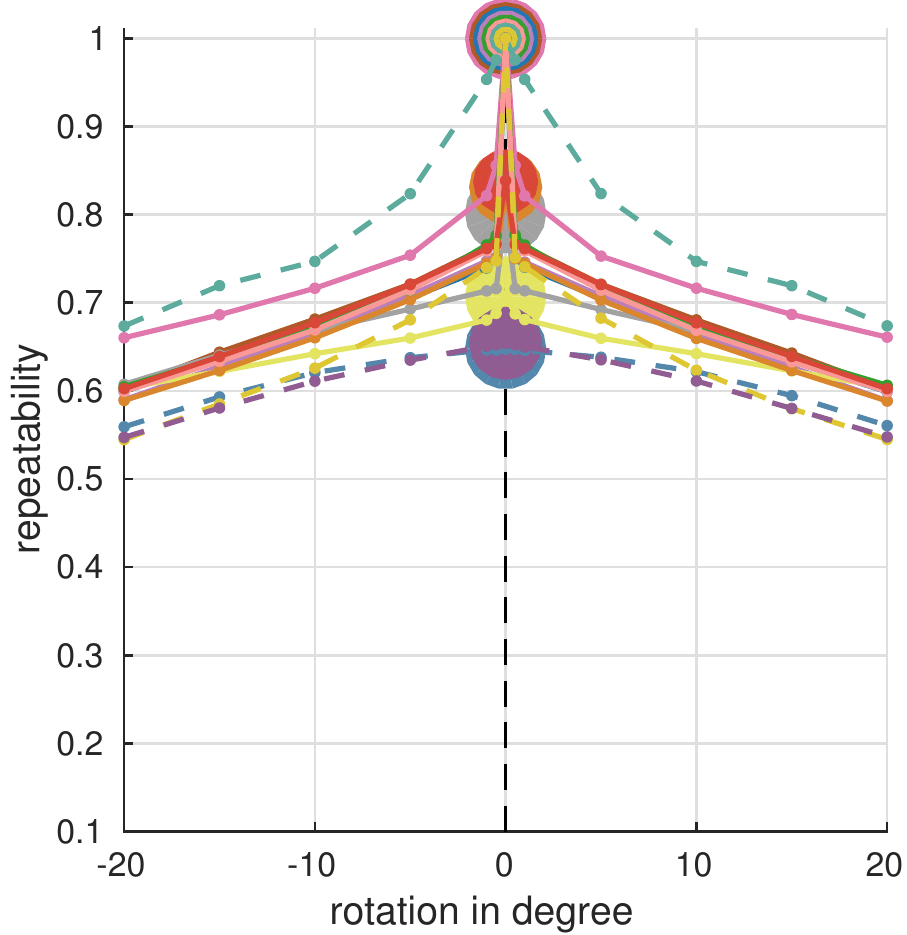}
\par\end{centering}

} & \subfloat[\label{fig:repeatiblity-illumination-change}Illumination.]{\begin{centering}
\includegraphics[width=0.3\textwidth]{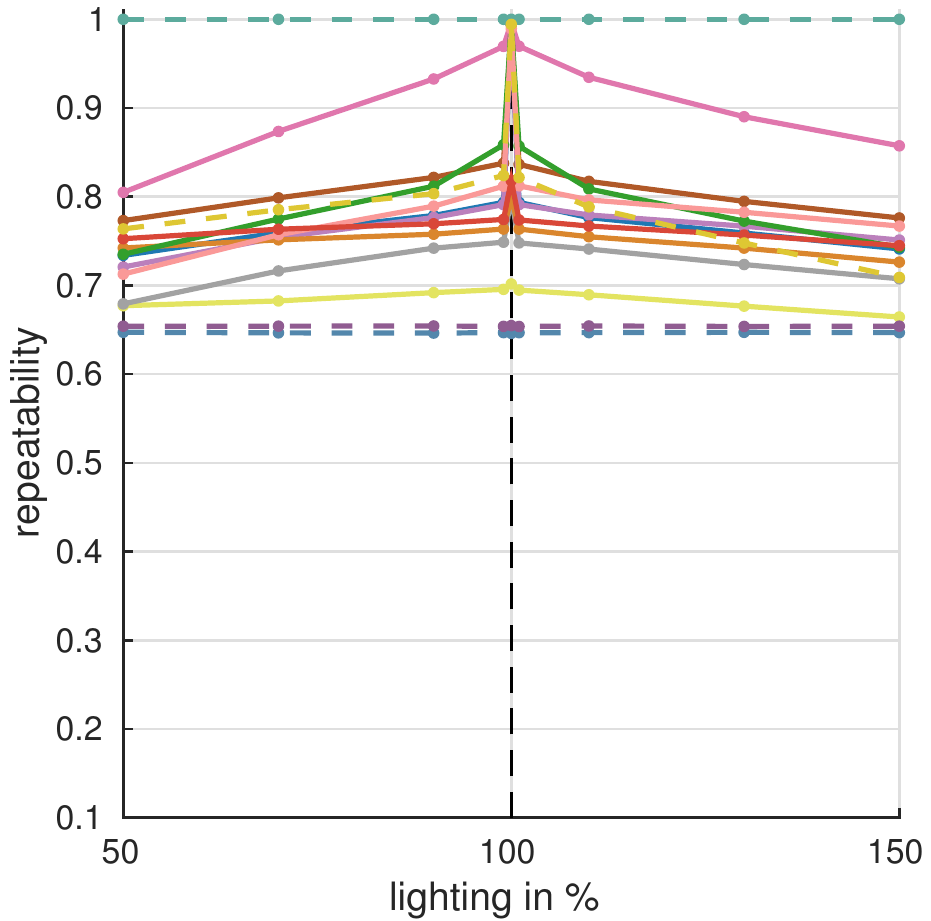}
\par\end{centering}

}\tabularnewline
\subfloat[\label{fig:repeatiblity-blurring}Blur.]{\begin{centering}
\includegraphics[width=0.3\textwidth]{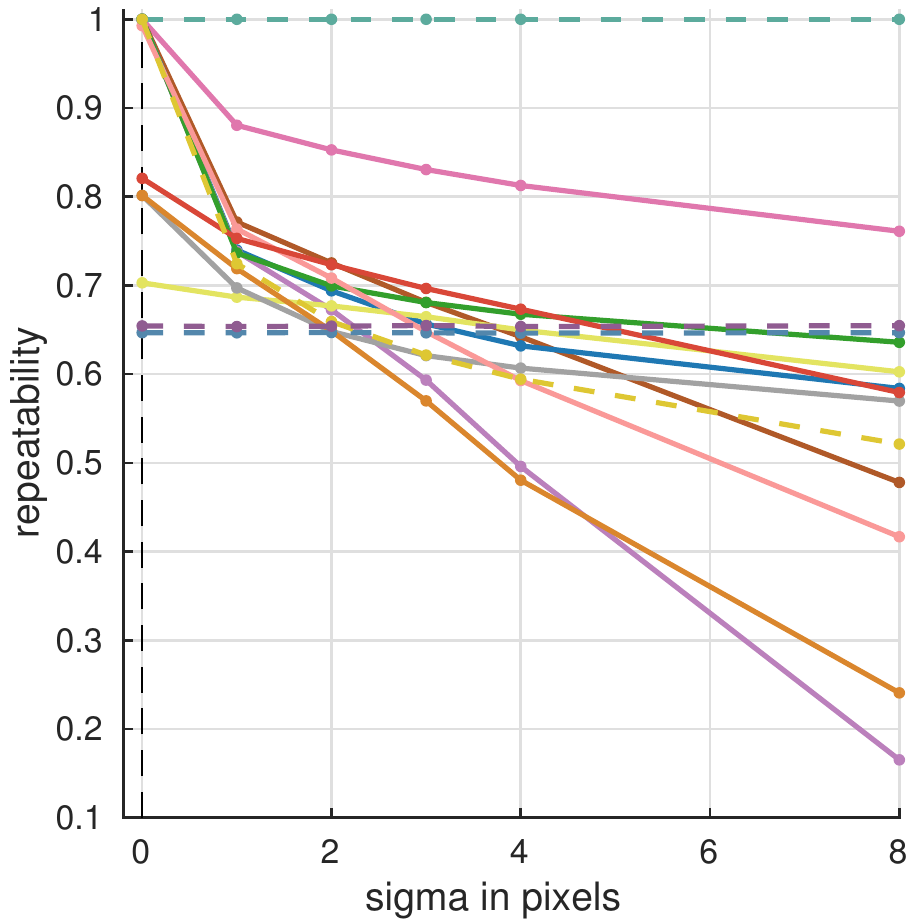}
\par\end{centering}

} & \subfloat[\label{fig:repeatiblity-saltnpepper}Salt and pepper noise.]{\begin{centering}
\includegraphics[width=0.3\textwidth]{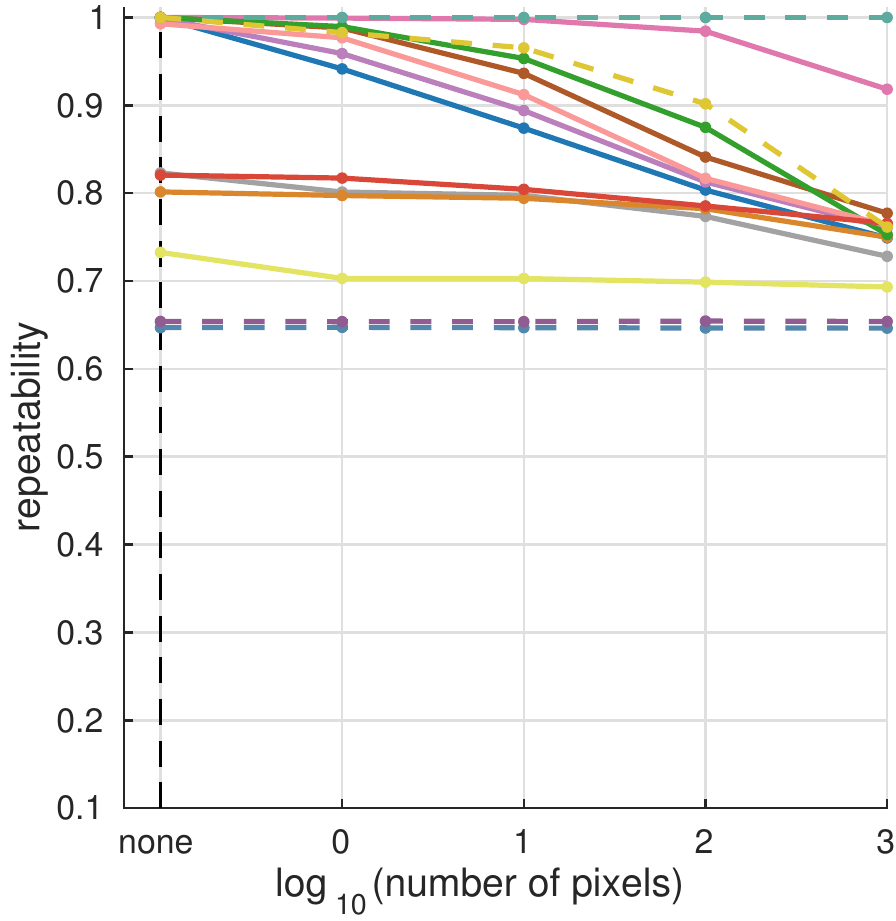}
\par\end{centering}

} & \subfloat{\raggedright{}\includegraphics[bb=0bp -15bp 152bp 209bp,width=0.2\textwidth]{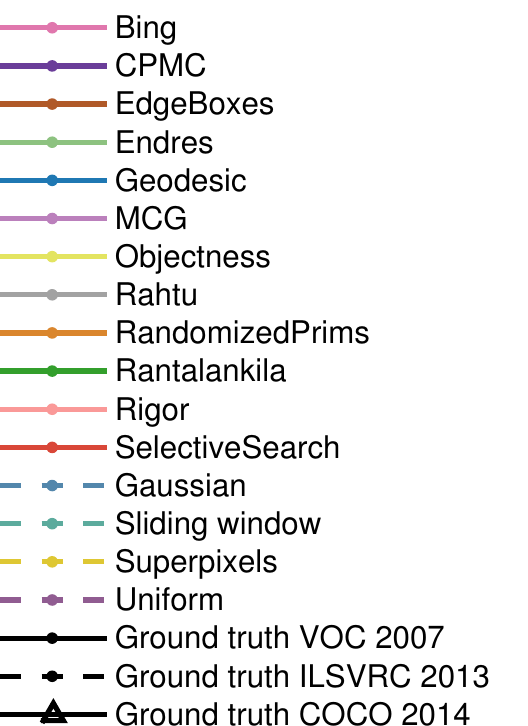}\hspace*{0.1\columnwidth}}\tabularnewline
\end{tabular}
\par\end{centering}

\protect\caption{\label{fig:repeatability}Repeatability results under various perturbations.}

\centering{}\vspace{-1em}
\end{figure*}

There are some salient aspects of the result curves in figure \ref{fig:repeatability}
that need additional explanation. First, not all methods have $100\%$
repeatability when there is no perturbation. This is due to random
components in the selection of proposals for several methods. Attempting
to remove a method's random component is beyond the scope of this
work and could potentially considerably alter the method. A second
important aspect is the large drop of repeatability for most methods,
even for subtle image changes. We observed that many of the methods
based on superpixels are particularly prone to such perturbations.
Indeed the \texttt{Superpixels} baseline itself shows high sensitivity
to perturbations, so the instability of the superpixels likely explains
much of this effect. Inversely we notice that methods that are not
based on superpixels are most robust to small image changes (e.g.
\texttt{Bing} and also the baselines that ignore image content).

We now discuss the details and effects of each perturbation on repeatability,
shown in figure \ref{fig:repeatability}:

\paragraph*{Scale (\ref{fig:repeatiblity-scale-change}\textmd{)}}

We uniformly sample the scale factor from $.5\times$ to $2\times$,
and test additional scales near the original resolution ($.9\times,\,.95\times,\,.99\times,\,1.01\times,\,1.05\times,\,1.1\times$).
Upscaling is done with bicubic interpolation and downscaling with
anti-aliasing. All methods except \texttt{Bing} show a drastic drop
with small scale changes, but suffer only minor degradation for larger
changes. \texttt{Bing} is more robust to small scale changes; however,
it is more sensitive to larger changes due to its use of a coarse
set of box sizes while searching for candidates (this also accounts
for its dip in repeatability at half scales). The \texttt{SlidingWindow}
baseline suffers from the same effect.

\paragraph*{JPEG artefacts (\ref{fig:repeatiblity-jpeg-change}\textmd{)}}

To create JPEG artefacts we write the target image to disk with the
Matlab function imwrite and specify a quality settings ranging from
5\% to 100\%, see figure~\ref{fig:pertubations-range}. Even the
100\% quality setting is lossy, so we also include a lossless setting
for comparison. Similar to scale change, even slight compression has
a large effect and more aggressive compression shows monotonic degradation.
Despite using gradient information, \texttt{Bing} is most robust to
these kind of changes.

\paragraph*{Rotation (\ref{fig:repeatiblity-rotation-change}\textmd{)}}

We rotate the image in $5{}^{\circ}$ steps between $-20{}^{\circ}$
and $20{}^{\circ}$. To ensure roughly the same content is visible
under all rotations, we construct the largest box with the same aspect
as the original image that fits into the image under a $20{}^{\circ}$
rotation and use this crop for all other rotations, see figure~\ref{fig:rotation-examples}.
All proposal methods are equally affected by image rotation. The drop
of the \texttt{Uniform} and \texttt{Gaussian} baselines indicate the
repeatability loss due to the fact that we are matching rotated bounding
boxes.

\paragraph*{Illumination (\ref{fig:repeatiblity-illumination-change})}

To synthetically alter illumination of an image we changed its brightness
channel in HSB colour space. We vary the brightness between $50\%$
and $150\%$ of the original image so that some over and under saturation
occurs, see figure~\ref{fig:pertubations-range}. Repeatability under
illumination changes shows a similar trend as under JPEG artefacts.
Methods based on superpixels are heavily affected. \texttt{Bing} is
more robust, likely due to use of gradient information which is known
to be fairly robust to illumination changes.

\paragraph*{Blur (\ref{fig:repeatiblity-blurring})}

We blur the images with a Gaussian kernel with standard deviations
$0\le\sigma\le8$, see figure~\ref{fig:pertubations-range}. The
repeatability results again exhibit a similar trend although the drop
is stronger for a small $\sigma$.

\paragraph*{Salt and pepper noise (\ref{fig:repeatiblity-saltnpepper})}

We sample between 1 and $1000$ random locations in the image and
change the colour of the pixel to white if it is a dark and to black
otherwise, see figure~\ref{fig:pertubations-range}. Surprisingly,
most methods already lose some repeatability when even a single pixel
is changed. Significant degradation in repeatability for the majority
of the methods occurs when merely ten pixels are modified.

\paragraph*{Discussion}

Small changes to an image cause noticeable differences in the set
of detection proposals for all methods except \texttt{Bing}. The higher
repeatability of \texttt{Bing} is explained by its sliding window
pattern, which has been designed to cover almost all possible annotations
with $\mbox{IoU}=0.5$ (see also \texttt{Cracking Bing}\,\cite{Zhao2014Bmvc}).
As one cause for poor repeatability we identify the segmentation algorithm
on which many methods build. Among all proposal methods, \texttt{EdgeBoxes}
also performs favourably, possibly because it avoids the hard decision
of grouping pixels into superpixels.

We also experimented with repeatability of boxes that touch annotations
sufficiently ($\text{{IoU}}\geq0.5$), which showed very similar trends,
indicating that the issue of repeatability also applies to proposals
that partially cover objects.

Different applications will be more or less sensitive to repeatability.
Our results indicate that if repeatability is a concern, the proposal
method should be selected with care. For object detection, another
aspect of interest is recall, which we explore in the next section.

\section{\label{sec:Proposal-recall}Proposal recall}

\begin{figure*}
\begin{centering}
\subfloat[\label{fig:recall-vs-iuo-at-100-windows}$100$ proposals per image.]{\begin{centering}
\includegraphics[width=0.271\textwidth]{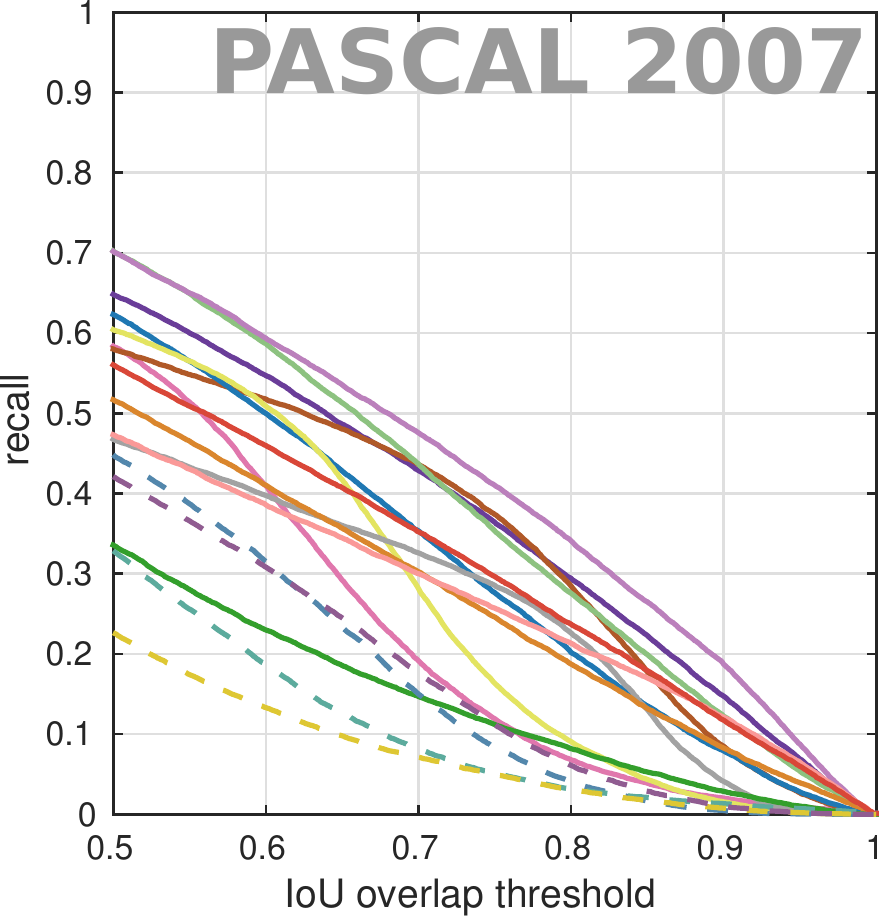}
\par\end{centering}

}\subfloat[\label{fig:recall-vs-iuo-at-1000-windows}$1\,000$ proposals per
image. ]{\begin{centering}
\includegraphics[width=0.278\textwidth]{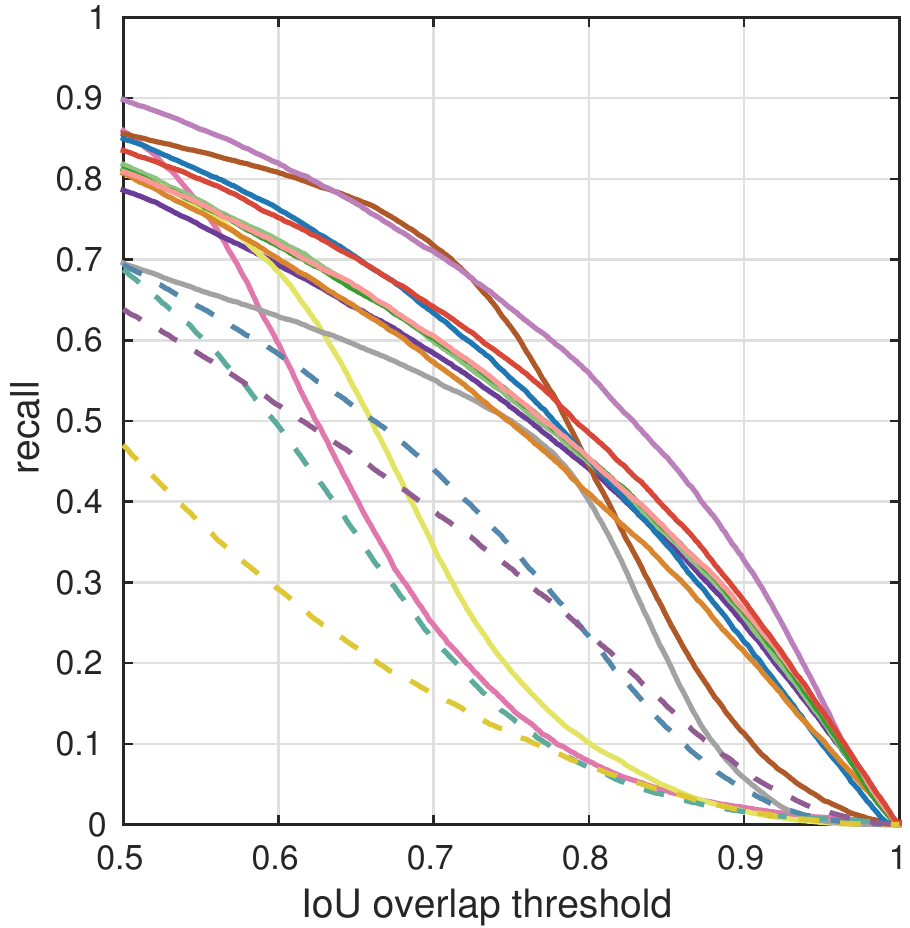}
\par\end{centering}

}\subfloat[\label{fig:recall-vs-iuo-at-10000-windows}$10\,000$ proposals per
image. ]{\begin{centering}
\includegraphics[width=0.275\textwidth]{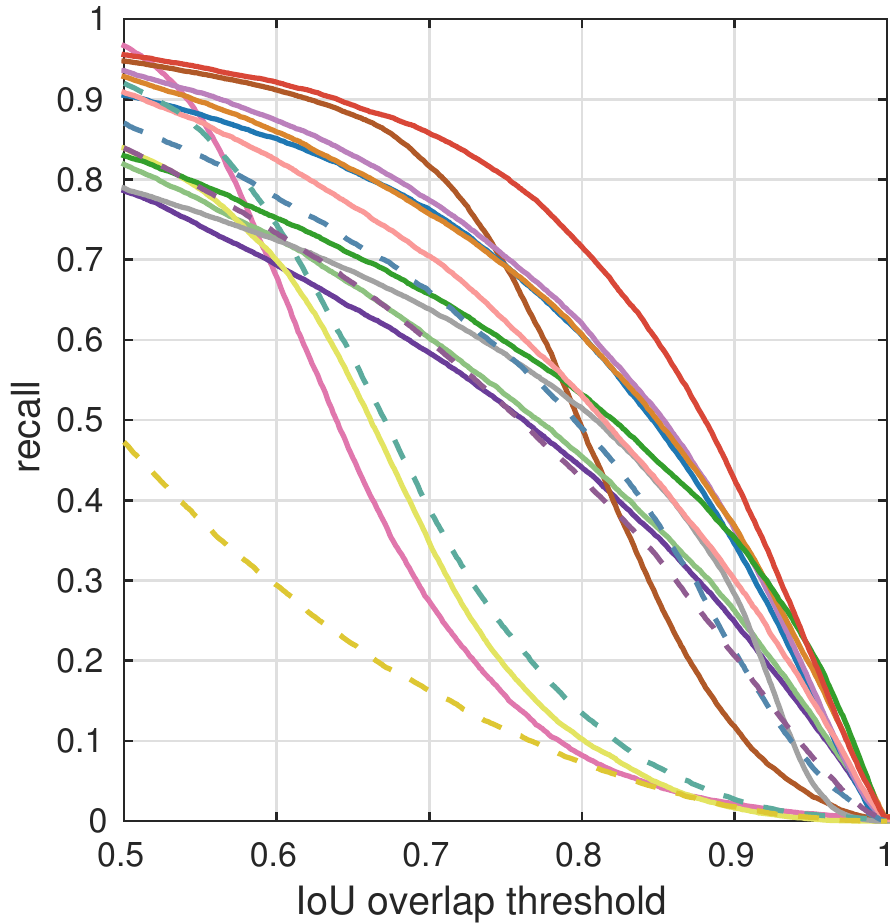}
\par\end{centering}

}\includegraphics[bb=0bp -30bp 114bp 175bp,width=0.148\textwidth]{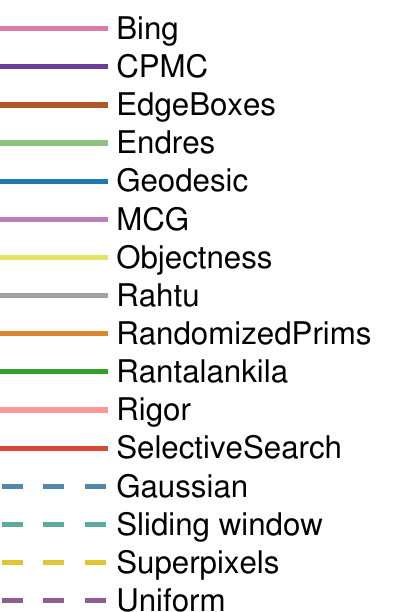}
\par\end{centering}

\protect\caption{\label{fig:recall-versus-iou-threshold-pascal}Recall versus IoU threshold
on the PASCAL VOC 2007 test set.}
\vspace{-1.5em}
\end{figure*}
\begin{figure*}
\centering{}\subfloat[\label{fig:recall-at-iou-0.5-vs-number-of-proposals-pascal}Recall
at 0.5 IoU.]{\begin{centering}
\includegraphics[width=0.272\textwidth]{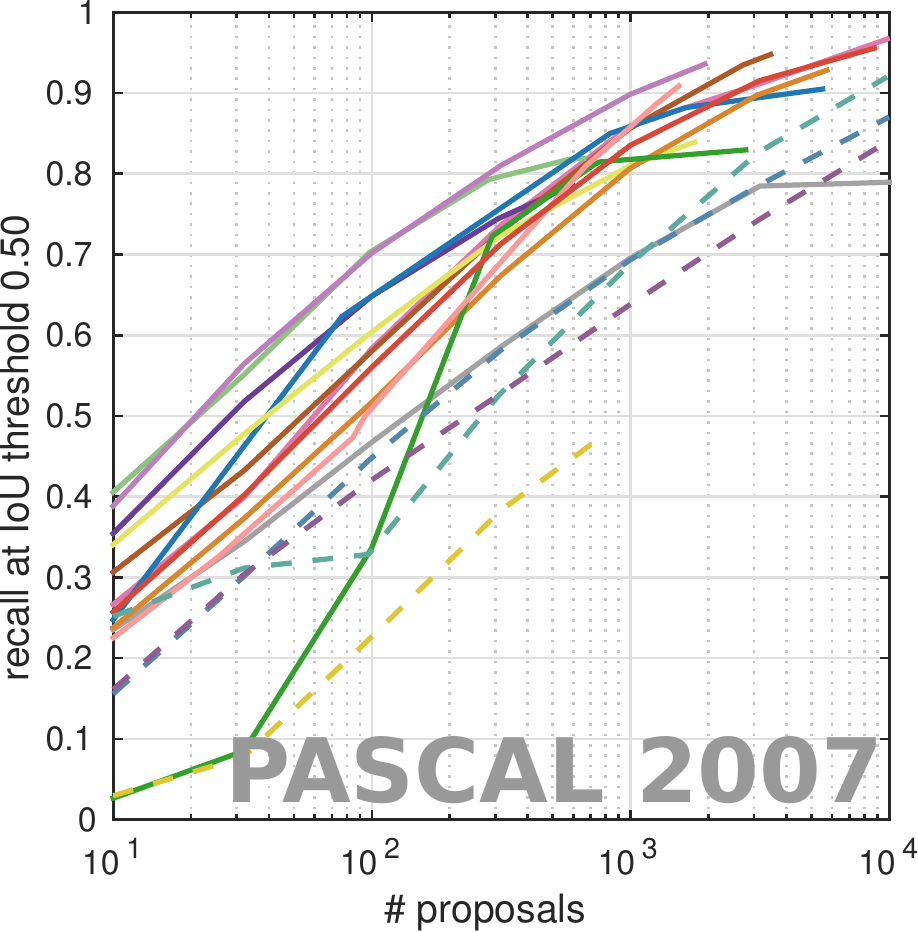}
\par\end{centering}

}\subfloat[\label{fig:recall-at-iou-0.8-vs-number-of-proposals-pascal}Recall
at 0.8 IoU.]{\begin{centering}
\includegraphics[width=0.275\textwidth]{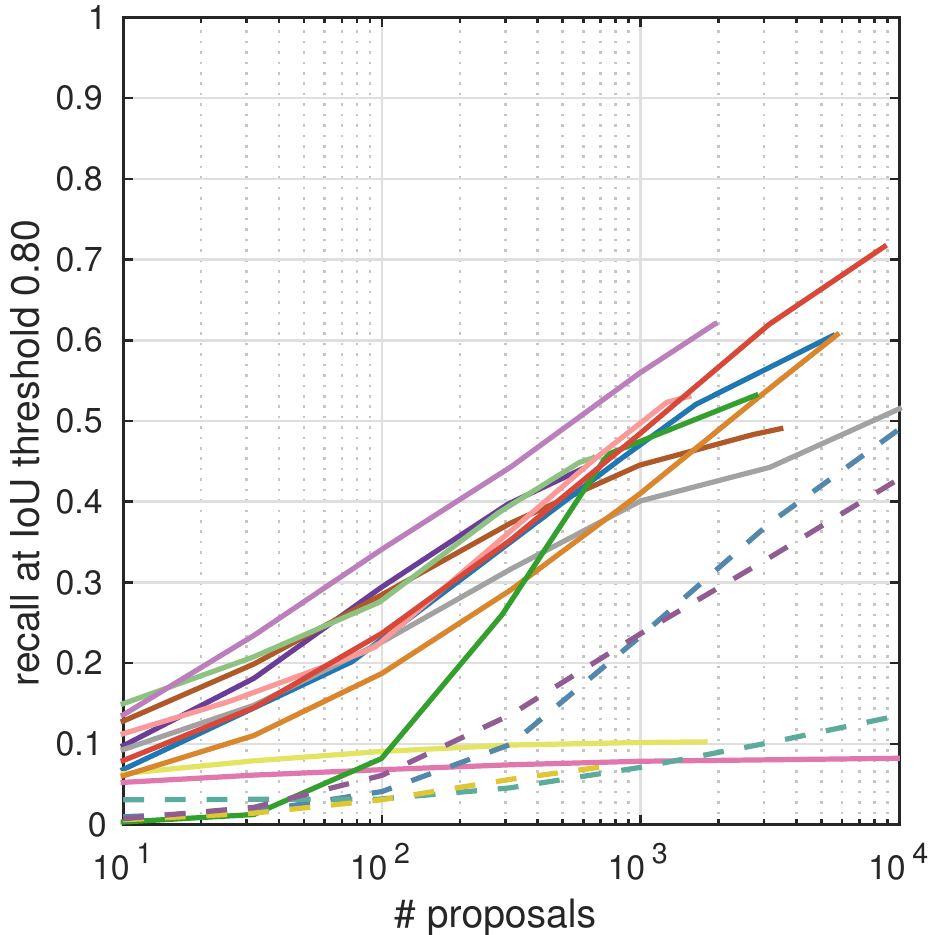}
\par\end{centering}

}\subfloat[{\label{fig:recall-vs-iuo-area-vs-number-of-proposals-pascal}Average
recall (between $[0.5,\,1]\ \mbox{IoU}$).}]{\begin{centering}
\includegraphics[width=0.275\textwidth]{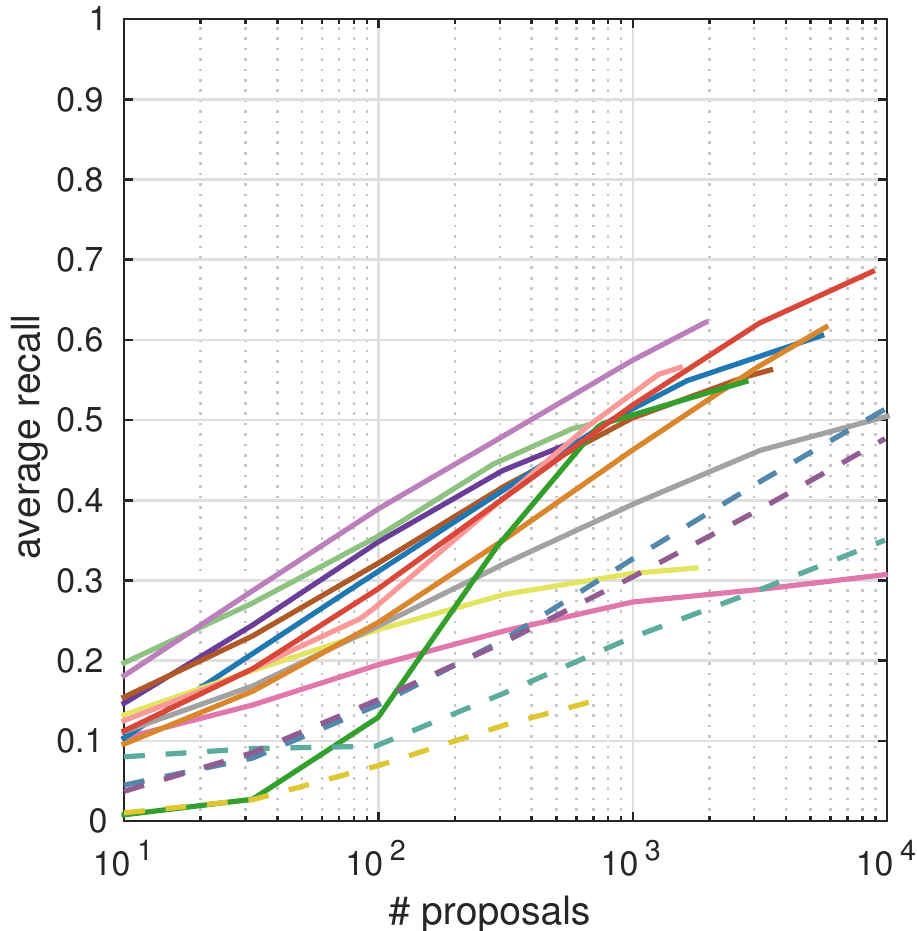}
\par\end{centering}

}\includegraphics[bb=0bp -30bp 114bp 175bp,width=0.148\textwidth]{figures/recall_legend}\protect\caption{\label{fig:recall-versus-num-windows-pascal}Recall versus number
of proposal windows on the PASCAL VOC 2007 test set.}
\vspace{-1.5em}
\end{figure*}
\begin{figure*}
\begin{centering}
\subfloat[\label{fig:recall-vs-iuo-at-1000-windows-imagenet}$1\,000$ proposals
per image.]{\begin{centering}
\includegraphics[width=0.271\textwidth]{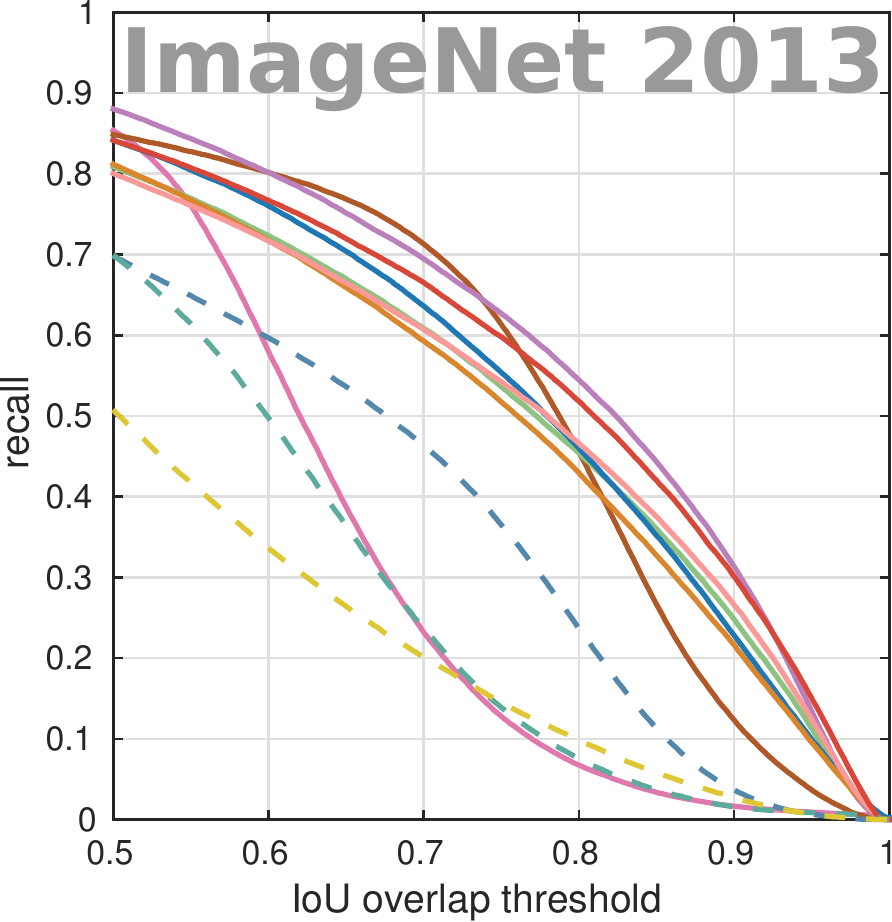}
\par\end{centering}

}\subfloat[\label{fig:recall-at-iou-0.8-vs-number-of-proposals-imagenet}Recall
at 0.8 IoU.]{\begin{centering}
\includegraphics[width=0.275\textwidth]{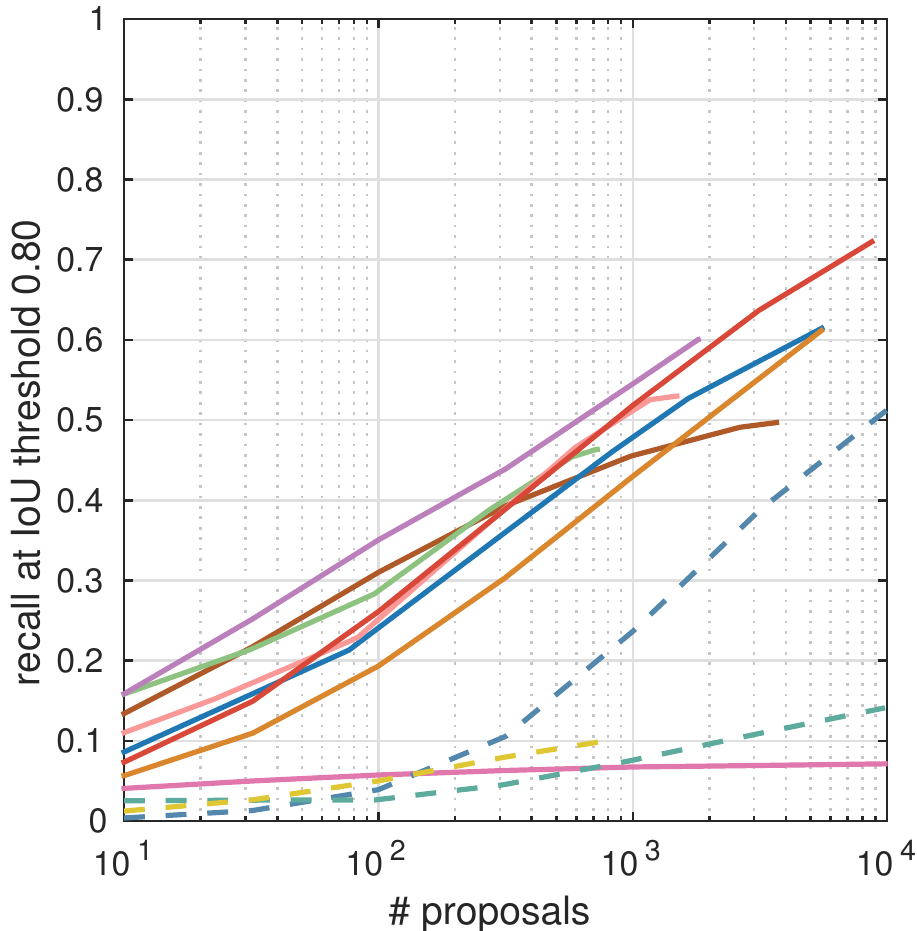}
\par\end{centering}

}\subfloat[{\label{fig:recall-vs-iuo-area-vs-number-of-proposals-imagenet}Average
recall (between $[0.5,\,1]\ \mbox{IoU}$).}]{\begin{centering}
\includegraphics[width=0.275\textwidth]{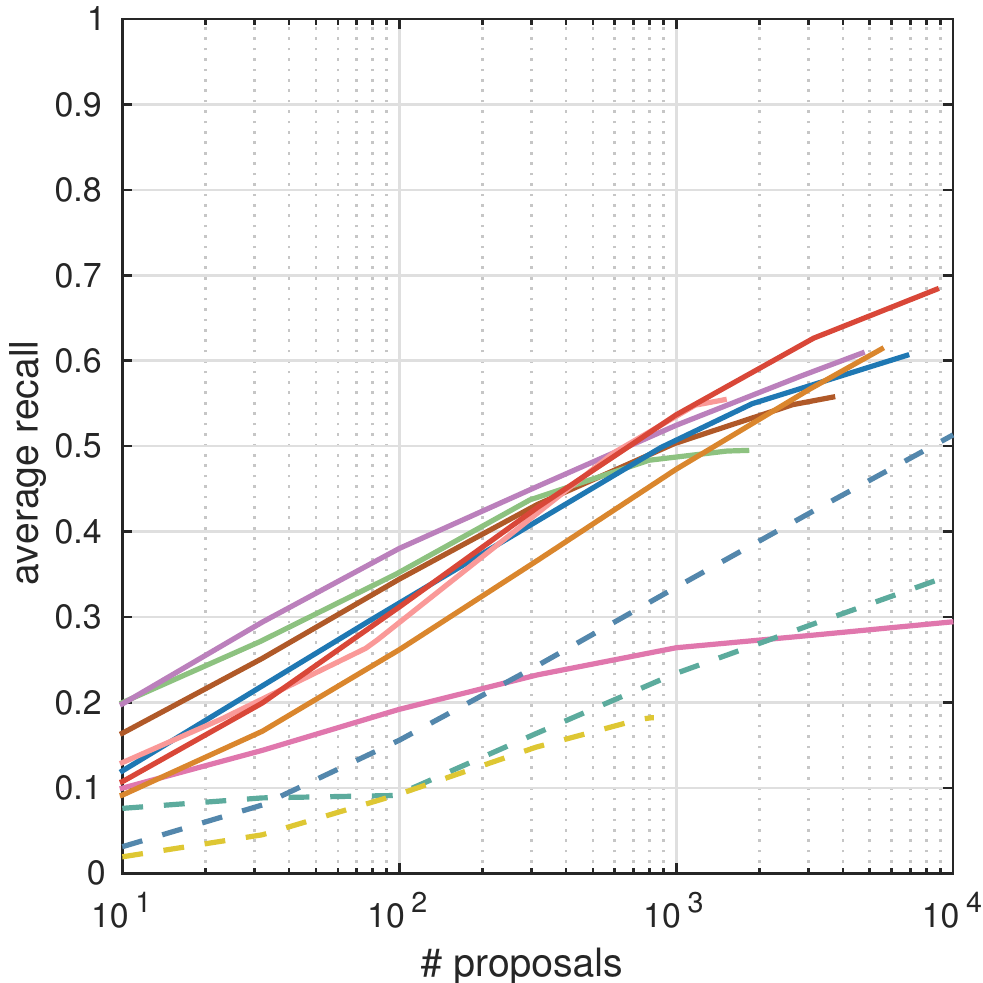}
\par\end{centering}

}\includegraphics[bb=0bp -75bp 114bp 121bp,width=0.148\textwidth]{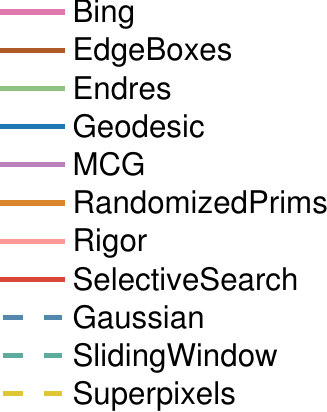}
\par\end{centering}

\protect\caption{\label{fig:imagenet-quality-results}Recall on the ImageNet 2013 validation
set.}

\centering{}\vspace{-0.5em}
\end{figure*}
\begin{figure*}
\begin{centering}
\subfloat[\label{fig:recall-vs-iuo-at-1000-windows-coco}$1\,000$ proposals
per image.]{\begin{centering}
\includegraphics[bb=0bp 0bp 258bp 266bp,width=0.272\textwidth]{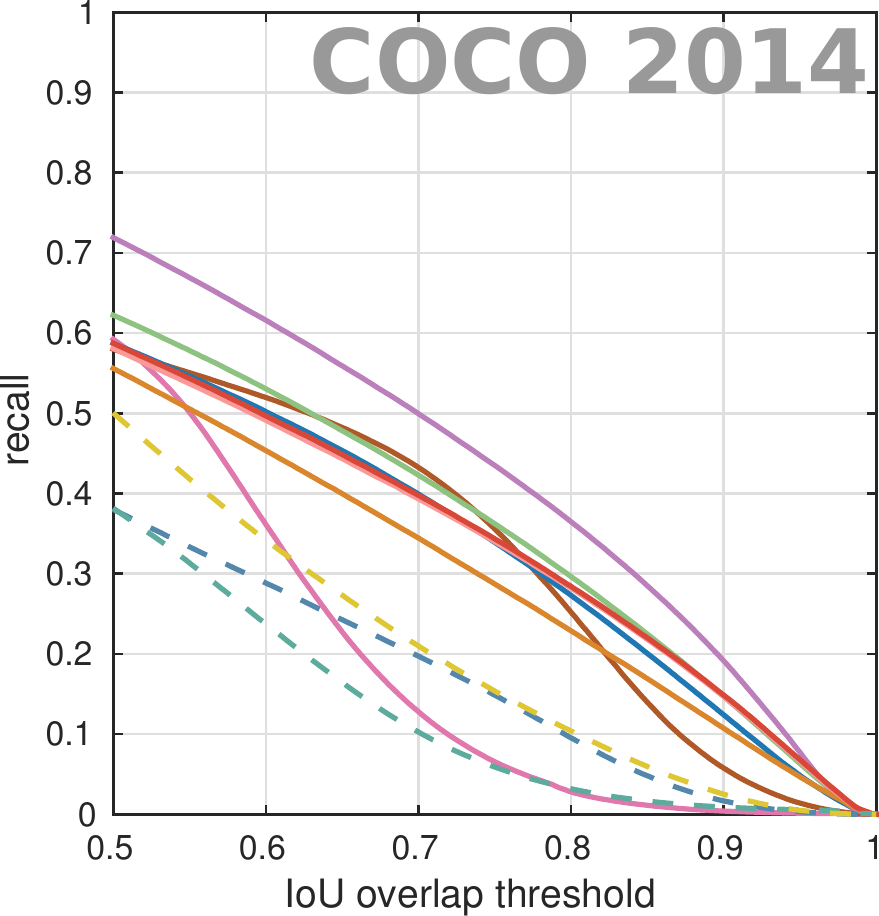}
\par\end{centering}

}\subfloat[\label{fig:recall-at-iou-0.8-vs-number-of-proposals-coco}Recall at
0.8 IoU.]{\begin{centering}
\includegraphics[width=0.275\textwidth]{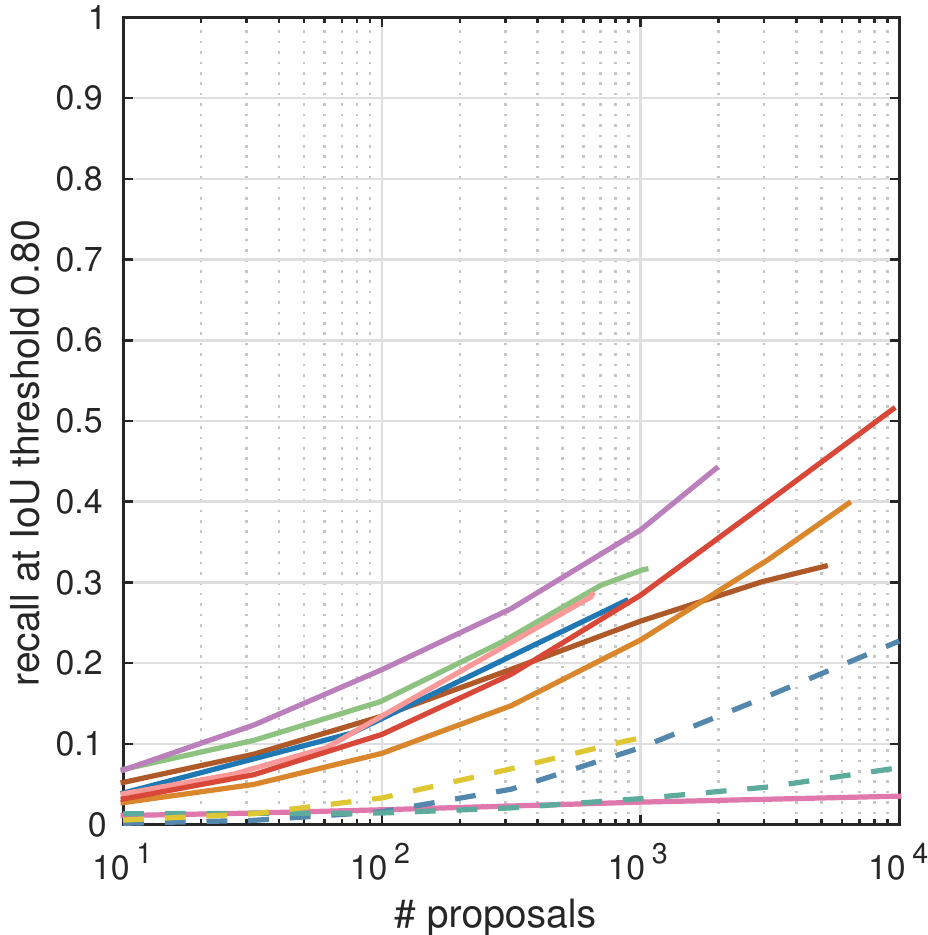}
\par\end{centering}

}\subfloat[{\label{fig:recall-vs-iuo-area-vs-number-of-proposals-coco}Average
recall (between $[0.5,\,1]\ \mbox{IoU}$).}]{\begin{centering}
\includegraphics[width=0.275\textwidth]{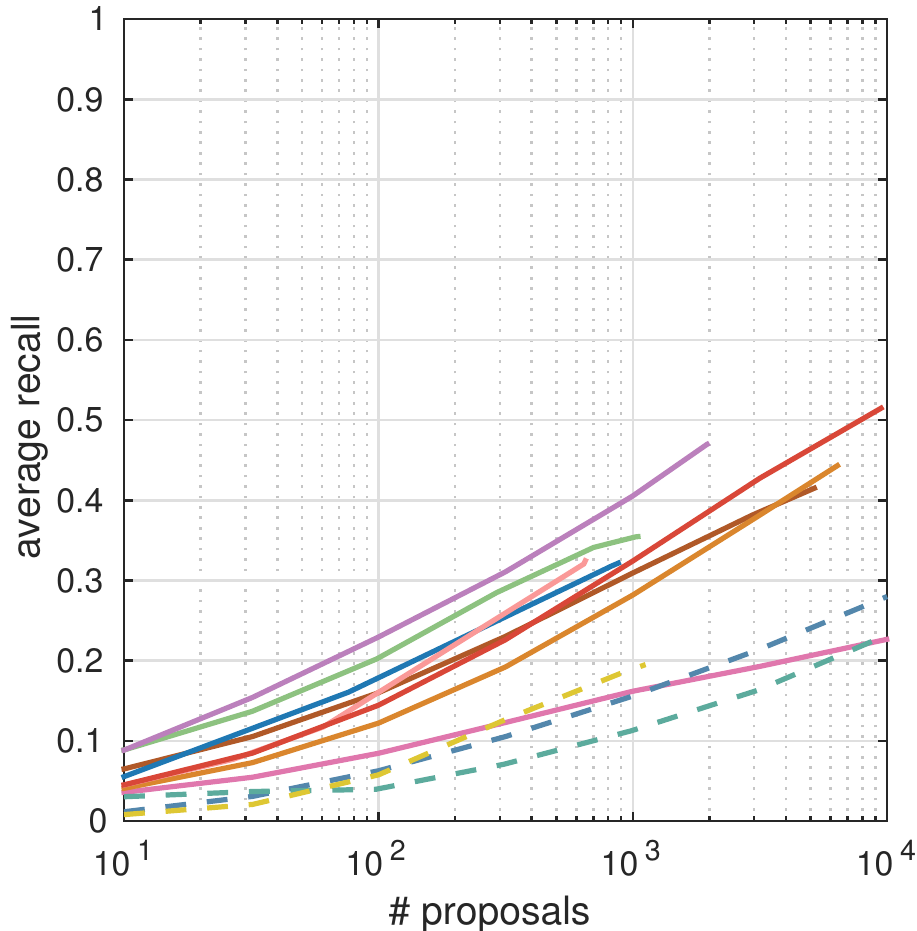}
\par\end{centering}

}\includegraphics[bb=0bp -75bp 114bp 121bp,width=0.148\textwidth]{figures/imagenet_recall_legend}
\par\end{centering}

\protect\caption{\label{fig:imagenet-quality-results-1}Recall on the MS COCO 2014
validation set.}

\centering{}\vspace{-0.5em}
\end{figure*}
When using detection proposals for detection it is important to have
a good coverage of the objects of interest in the test image, since
missed objects cannot be recovered in the subsequent classification
stage. Thus it is common practice to evaluate the quality of proposals
based on the recall of the ground truth annotations.

\subsection{\label{sub:proposals-recall-evaluation-protocol}Evaluation protocol
for recall}

The protocol introduced in \cite{Alexe2010Cvpr} (using the PASCAL
VOC 2007 dataset \cite{Everingham2014Ijcv}) has served as a guideline
for most evaluations in the literature. While previous papers do show
various comparisons on PASCAL, the train and test sets vary amongst
papers, and the metrics shown tend to favour different methods. We
provide an extensive and unified evaluation and show that different
metrics result in different rankings of proposal methods (e.g. see
figure \ref{fig:recall-vs-iuo-at-1000-windows} versus \ref{fig:recall-at-iou-0.8-vs-number-of-proposals-pascal}).

\paragraph*{Metrics}

Evaluating (class agnostic) detection proposals is quite different
from traditional class-specific detection \cite{Hoiem2012Eccv} since
most metrics (class confusion, background confusion, precision, etc.)
do not apply. Instead, one of the primary metrics for evaluating proposals
is, for a fixed number of proposals, the fraction of ground truth
annotations covered as the intersection over union (IoU) threshold
is varied (figure \ref{fig:recall-versus-iou-threshold-pascal}).
Another common and complementary metric is, for a fixed IoU threshold,
proposal recall as the number of proposals is varied (figure \ref{fig:recall-at-iou-0.5-vs-number-of-proposals-pascal},
\ref{fig:recall-at-iou-0.8-vs-number-of-proposals-pascal}). Finally,
we define and report a novel metric, the average recall (AR) between
IoU 0.5 to 1, and plot AR versus number of proposals (figure \ref{fig:recall-vs-iuo-area-vs-number-of-proposals-pascal}).

\paragraph*{PASCAL}

We evaluate recall on the full PASCAL VOC 2007 test set \cite{Everingham2014Ijcv},
which includes $20$ object categories present in $\sim\negmedspace5\,000$
unconstrained images. For the purpose of proposal evaluation we include
all $20$ object categories and all ground truth bounding boxes, including
``difficult'' ones, since our goal is to measure maximum recall.
In contrast to \cite{Alexe2010Cvpr}, we compute a matching between
proposals and ground truth, so one proposal cannot cover two objects.
Note that while different methods may be trained on different sets
of object categories and subsets of data, we believe evaluating on
all categories at test time is appropriate as we care about absolute
proposal quality. Such an evaluation strategy is further supported
as many methods have no training stage, yet provide competitive results
(e.g. \texttt{SelectiveSearch}).

\paragraph*{ImageNet}

The PASCAL VOC 2007 test set, on which most proposal methods have
been previously evaluated, has only $20$ categories, yet detection
proposal methods claim to predict proposals for \emph{any} object
category. Thus there is some concern that the proposal methods may
be tuned to the PASCAL categories and not generalise well to novel
categories. To investigate this potential bias, we also evaluate methods
on the larger ImageNet \cite{Deng2009Cvpr} 2013 validation set, which
contains annotations for $200$ categories in over $\sim\negmedspace20\,000$
images. It should be noted that these $200$ categories are \emph{not}
fine grained versions of the PASCAL ones. They include additional
types of animals (e.g. crustaceans), food items (e.g. hot-dogs), household
items (e.g. diapers), and other diverse object categories.

\paragraph*{MS COCO}

Although ImageNet has $180$ more classes than PASCAL, it is still
similar in statistics like number of objects per image and size of
objects. Microsoft Common Objects in Context (MS COCO) \cite{mscoco2015}
has more objects per image, smaller objects, but also fewer object
classes ($80$ object categories). We evaluate the recall of this
dataset to further investigate potential biases of proposal methods.
We evaluate the recall on all annotations excluding the ``crowd''
annotations which may mark large image areas including a lot of background.

\subsection{\label{sub:Recall-results}Recall results}

PASCAL Results in figure \ref{fig:recall-versus-iou-threshold-pascal}
and \ref{fig:recall-versus-num-windows-pascal} present a consistent
trend across the different metrics. \texttt{MCG}, \texttt{Edge\-Boxes},
\texttt{Selective\-Search}, \texttt{Rigor}, and \texttt{Geodesic}
are the best methods across different numbers of proposals. \texttt{Selective\-Search}
is surprisingly effective despite being fully hand-crafted (no machine
learning involved). When considering less than $10^{3}$ proposals,
\texttt{MCG}, \texttt{En\-dres}, and \texttt{CPMC} provide strong
results.

Overall, the methods fall into two groups: well localised methods
that gradually lose recall as the IoU threshold increases and methods
that only provide coarse bounding box locations, so their recall drops
rapidly. All baseline methods, as well as \texttt{Bing}, \texttt{Rahtu},
\texttt{Object\-ness}, and \texttt{Edge\-Boxes} fall into the latter
category. \texttt{Bing} in particular, while providing high repeatability,
only provides high recall at $\mbox{IoU}=0.5$ and drops dramatically
when requiring higher overlap (the reason for this is identified in
\cite{Zhao2014Bmvc}).

\paragraph*{Baselines}

When inspecting figure \ref{fig:recall-versus-iou-threshold-pascal}
from left to right, one notices that with few proposals the baselines
provide relatively low recall (figure \ref{fig:recall-vs-iuo-at-100-windows}).
However as the number of proposals increases, \texttt{Gau\-ssian}
and \texttt{Uniform} become more competitive (figure \ref{fig:recall-vs-iuo-at-1000-windows}).
In relative gain, detection proposal methods have most to offer for
low numbers of windows.

\paragraph*{Average Recall}

Rather than reporting recall at particular IoU thresholds, we also
report the average recall (AR) between IoU 0.5 to 1 (which is related
to the ABO metric, see \S\ref{sec:additional-metrics}), and plot
AR for varying number of proposals in figure \ref{fig:recall-vs-iuo-area-vs-number-of-proposals-pascal}.
Much like the average precision (AP) metric for (class specific) object
detection, AR summarises proposal performance across IoU thresholds
(for a given number of proposals). In fact, in \S\ref{sec:Using-detection-proposals}
we will show that AR correlates well with detection performance. As
can be seen in figure \ref{fig:recall-vs-iuo-area-vs-number-of-proposals-pascal},\texttt{
MCG} performs well across the entire range of number of proposals.
\texttt{Endres} and \texttt{EdgeBoxes} work well for a low number
of proposals while for a higher number of proposals \texttt{Rigor}
and \texttt{Selective\-Search} perform best.

\begin{figure}
\hfill{}\subfloat[\label{fig:all-datasets-AR}AR with 1000 proposals]{\includegraphics[width=0.48\columnwidth]{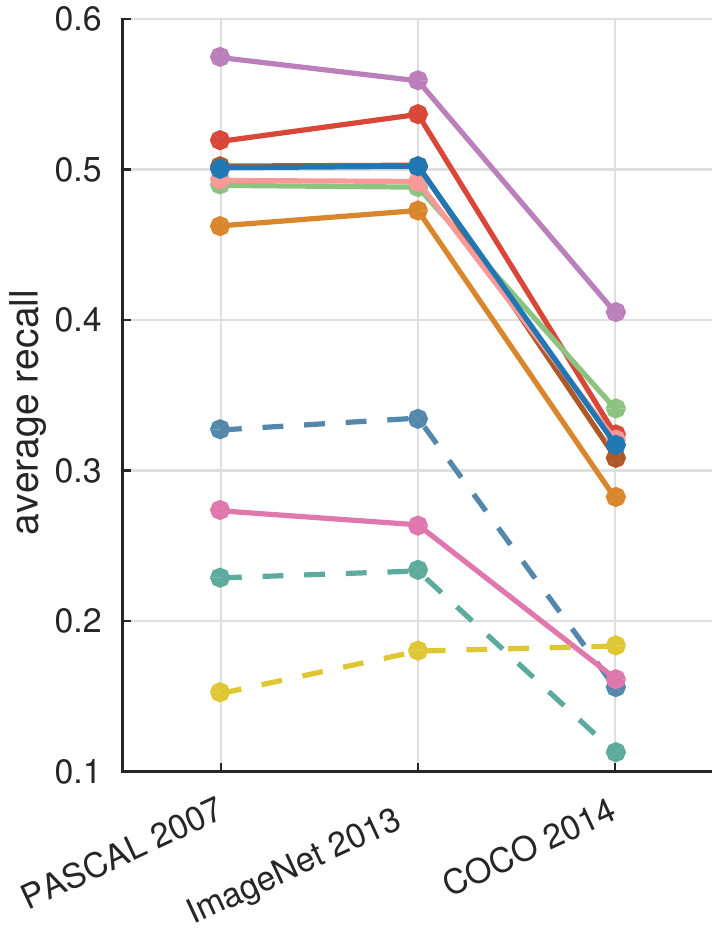}

}\hfill{}\subfloat[\label{fig:all-datasets-annotation-size}Ground truth size]{\includegraphics[bb=0bp -20bp 214bp 266bp,width=0.47\columnwidth]{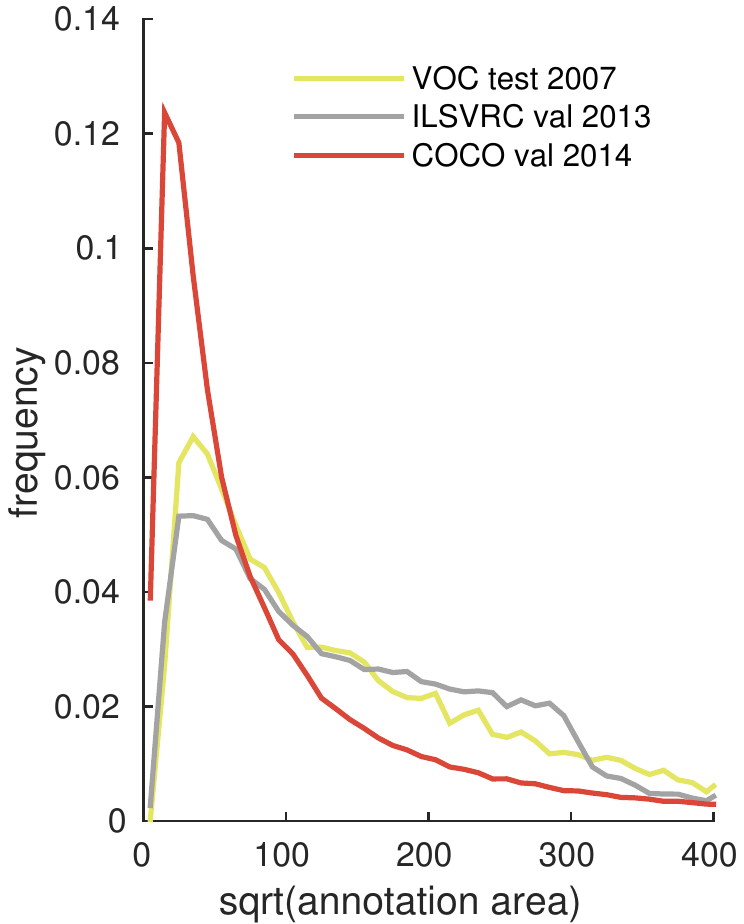}

}\hfill{}

\protect\caption{Comparison between all considered datasets: PASCAL VOC 2007 test set,
ImageNet 2013 validation set, MS COCO 2014 validation set (see methods
legend fig. \ref{fig:recall-vs-iuo-area-vs-number-of-proposals-pascal}).}
\end{figure}

\paragraph*{ImageNet}

As discussed, compared to PASCAL, ImageNet includes $10\times$ ground
truth classes and $4\times$ images. Somewhat surprisingly the ImageNet
results in figure \ref{fig:imagenet-quality-results} are almost identical
to the ones in figures \ref{fig:recall-vs-iuo-at-1000-windows}, \ref{fig:recall-at-iou-0.8-vs-number-of-proposals-pascal},
and \ref{fig:recall-vs-iuo-area-vs-number-of-proposals-pascal}. To
understand this phenomenon, we note that the statistics of ImageNet
match the ones of PASCAL. In particular the typical image size and
the mean number of object annotation per image (three) is similar
in both datasets. This helps explain why the recall behaviour is similar,
and why methods tuned on PASCAL still perform well on ImageNet.

\paragraph*{MS COCO}

We present the same results for MS COCO in figure \ref{fig:imagenet-quality-results-1}.
We see different absolute numbers, yet similar trends with some notable
exceptions as can be seen in figure \ref{fig:all-datasets-AR}. \texttt{EdgeBoxes}
no longer ranks significantly better than \texttt{Selective\-Search},
\texttt{Geodesic} and \texttt{Rigor} for few proposals. \texttt{MCG}
and \texttt{Endres} improve relative to the other methods, in particular
for a higher number of proposals. We attribute these difference to
different statistics of the dataset, particularly the different size
distribution, see figure \ref{fig:all-datasets-annotation-size}.

Overall, \texttt{MCG} is the top performing method across all datasets
in terms of both recall and AR at all settings. This is readily apparent
in figure \ref{fig:all-datasets-AR}.

\paragraph*{Generalisation}

We emphasise that although the results on PASCAL, ImageNet, and MS
COCO are quite similar (see figure \ref{fig:all-datasets-AR}), ImageNet
covers $200$ object categories, many of them unrelated to the $20$
PASCAL categories and COCO has significantly different statistics.
In other words, there is no measurable over-fitting of the detection
proposal methods towards the PASCAL categories. This suggests that
proposal methods transfer adequately amongst object classes, and can
thus be considered true ``objectness'' measures.

\section{\label{sec:Using-detection-proposals}Using the detection proposals}

In this section we analyse detection proposals for use with object
detectors. We consider two well known and quite distinct approaches
to object detection. First we use a variant of the popular DPM part-based
sliding window detector \cite{Felzenszwalb2010Pami}, specifically
the LM-LLDA detector\cite{Girshick2013ICCV}. We also test the state
of the art R-CNN \cite{Girshick2014Cvpr} and Fast R-CNN \cite{Girshick2015arXiv}
detectors which couple object proposals with a convolutional neural
network classification stage. Our goals are twofold. First, we aim
to measure the performance of different proposal methods for object
detection. Second, we are interested in evaluating how well the proposal
metrics reported in the previous sections can serve as a proxy for
predicting final detection performance. All following experiments
involving proposals use $1\,000$ proposals.

\subsection{\label{sub:miss-aligned-detectors}Detector responses around objects}

\begin{figure}[t]
\begin{centering}
\subfloat[\label{fig:detector-scoremap-scale-pos-average}Average R-CNN score
map across all ground truth annotations.]{\includegraphics[bb=70bp 0bp 650bp 110bp,clip,width=0.95\columnwidth]{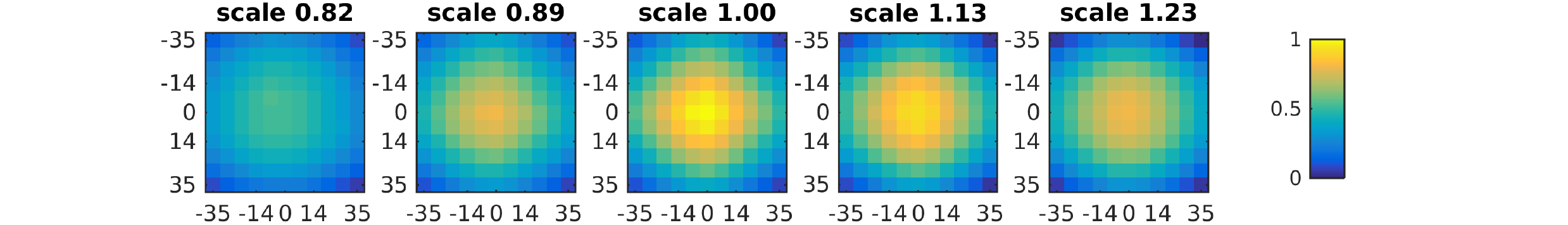}

}\vspace{-3mm}

\par\end{centering}

\begin{centering}
\subfloat[\label{fig:detector-scoremap-scale-pos-good}A score map similar to
the mean (around a correct detection).]{\includegraphics[bb=50bp 0bp 615bp 110bp,clip,width=0.95\columnwidth]{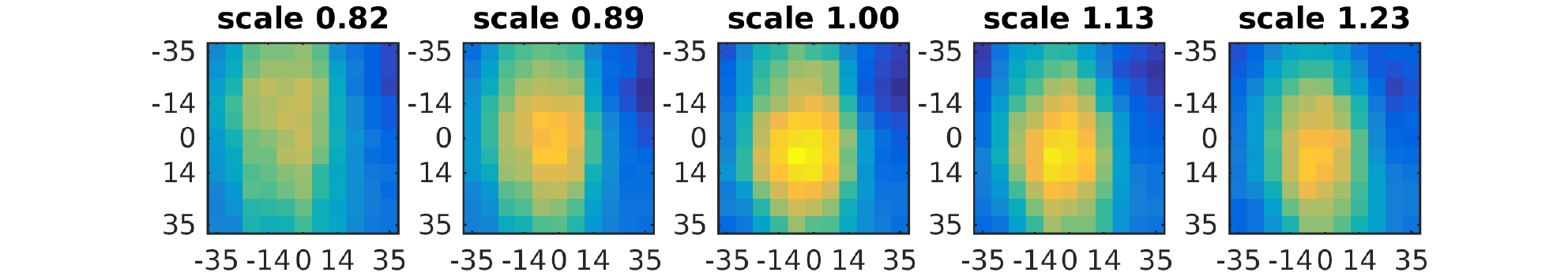}

}
\par\end{centering}

\begin{centering}
\vspace{-3mm}
\subfloat[\label{fig:detector-scoremap-scale-pos-bad}A score map different
than the mean (around a missed detection).]{\includegraphics[bb=50bp 0bp 615bp 110bp,clip,width=0.95\columnwidth]{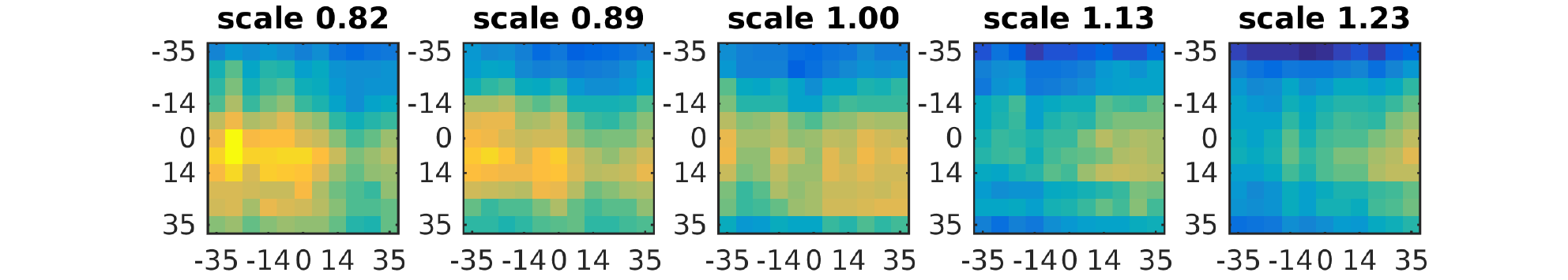}

}
\par\end{centering}

\begin{centering}
\vspace{0em}

\par\end{centering}

\protect\caption{\label{fig:detector-scoremaps-scale-pos}Normalised score maps of
the R-CNN around ground truth annotations on the PASCAL 2007 test
set. One grid cell in each map has width and height of $\sim\negmedspace7$px
after the object height has been resized to the detector window of
$227\negthinspace\times\negthinspace227\ \mbox{px}$ ($3\%$ of the
object height). }
\end{figure}
\begin{figure}[t]
\begin{centering}
\hspace*{\fill}\includegraphics[width=0.95\columnwidth]{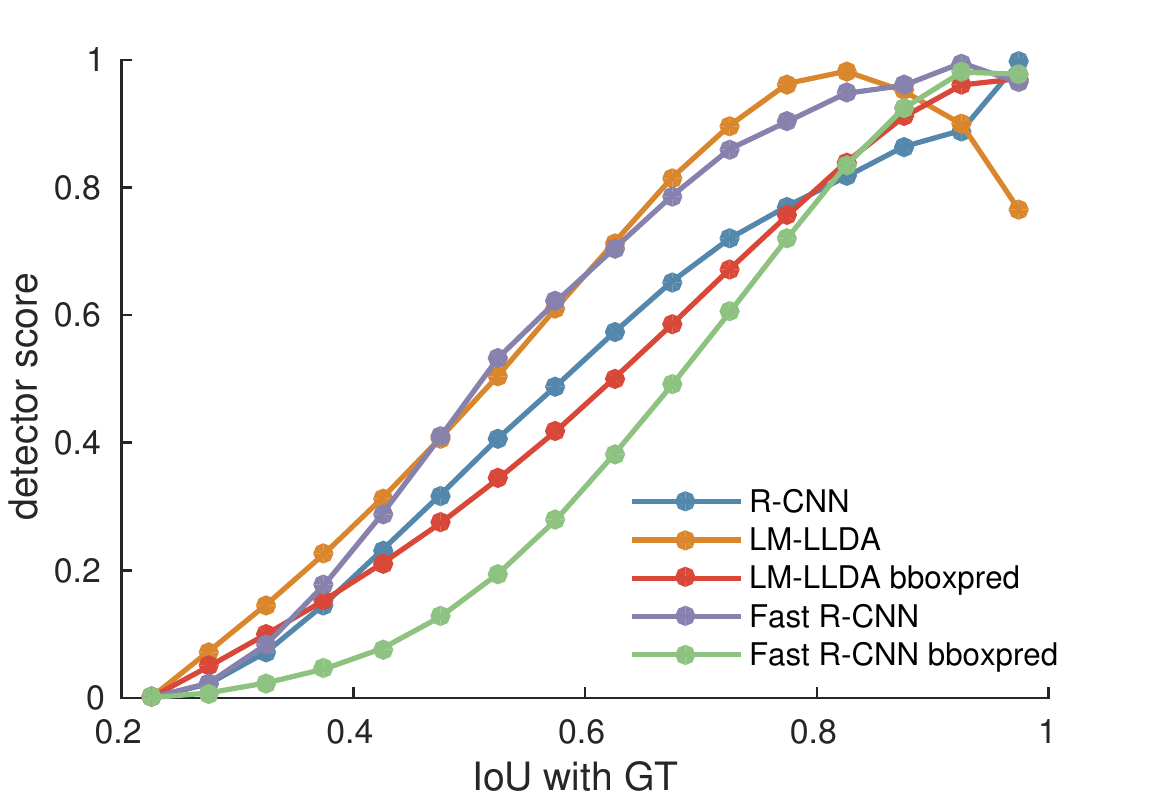}\hspace*{\fill}\vspace{-1mm}

\par\end{centering}

\protect\caption{\label{fig:detector-scoremaps}Normalised detector scores as a function
of the overlap between the detector window and the ground truth.}
\vspace{-2mm}
\end{figure}
As a preliminary experiment, we aim to quantify the importance of
having well localised proposals for object detection. We begin by
measuring how detection scores are affected by the overlap between
the detector window and the ground truth annotation on the PASCAL
2007 test set \cite{Everingham2014Ijcv}. When considering the detectors'
bounding box prediction, we use the refined position to compute the
overlap.

\paragraph*{Score map}

Figure \ref{fig:detector-scoremap-scale-pos-average} shows the average
R-CNN detection score around the ground truth annotations. We notice
that the score map is symmetric and attains a maximum at the ground
truth object location. In other words, the detector has no systematic
spatial or scale bias. However, averaging the score maps removes small
details and imperfections of individual score maps. When considering
individual activations instead of the average, we observe a high variance
in the quality of the score maps, see figures \ref{fig:detector-scoremap-scale-pos-good}
and \ref{fig:detector-scoremap-scale-pos-bad}.

\begin{table*}
\setlength\tabcolsep{2pt}

\hspace*{\fill}%
\begin{tabular}{ccccccccccccccccccccccc}
 & {\scriptsize{}aero} & {\scriptsize{}bicycle} & {\scriptsize{}bird} & {\scriptsize{}boat} & {\scriptsize{}bottle} & {\scriptsize{}bus} & {\scriptsize{}car} & {\scriptsize{}cat} & {\scriptsize{}chair} & {\scriptsize{}cow} & {\scriptsize{}table} & {\scriptsize{}dog} & {\scriptsize{}horse} & {\scriptsize{}mbike} & {\scriptsize{}person} & {\scriptsize{}plant} & {\scriptsize{}sheep} & {\scriptsize{}sofa} & {\scriptsize{}train} & {\scriptsize{}tv} & {\scriptsize{}\quad{}} & {\scriptsize{}mean}\tabularnewline
\hline 
{\scriptsize{}LM-LLDA }\texttt{\scriptsize{}Dense} & {\scriptsize{}33.7} & {\scriptsize{}61.3} & {\scriptsize{}12.4} & {\scriptsize{}18.5} & {\scriptsize{}26.7} & {\scriptsize{}53.0} & {\scriptsize{}57.2} & {\scriptsize{}22.4} & {\scriptsize{}22.7} & {\scriptsize{}25.6} & {\scriptsize{}25.1} & {\scriptsize{}14.0} & {\scriptsize{}59.2} & {\scriptsize{}51.0} & {\scriptsize{}39.1} & {\scriptsize{}13.6} & {\scriptsize{}21.7} & {\scriptsize{}38.0} & {\scriptsize{}48.8} & {\scriptsize{}44.0} &  & {\scriptsize{}34.4}\tabularnewline
\hline 
\texttt{\scriptsize{}Bing} & {\scriptsize{}\celln-7.5} & {\scriptsize{}\celln-23.2} & {\scriptsize{}\celln-6.2} & {\scriptsize{}\celln-8.1} & {\scriptsize{}\celln-10.6} & {\scriptsize{}\celln-13.3} & {\scriptsize{}\celln-17.5} & {\scriptsize{}\celln-6.8} & {\scriptsize{}\celln-9.8} & {\scriptsize{}\celln-15.4} & {\scriptsize{}\celln-7.5} & {\scriptsize{}\cellz-1.4} & {\scriptsize{}\celln-19.6} & {\scriptsize{}\celln-19.0} & {\scriptsize{}\celln-16.1} & {\scriptsize{}\celln-3.4} & {\scriptsize{}\celln-6.6} & {\scriptsize{}\celln-18.1} & {\scriptsize{}\celln-18.8} & {\scriptsize{}\celln-10.0} &  & {\scriptsize{}\celln-11.9}\tabularnewline
\texttt{\scriptsize{}CPMC} & {\scriptsize{}\cellz-1.0} & {\scriptsize{}\celln-15.0} & {\scriptsize{}\cellz-0.2} & {\scriptsize{}\celln-4.4} & {\scriptsize{}\celln-13.5} & {\scriptsize{}\cellz-1.8} & {\scriptsize{}\celln-9.2} & {\scriptsize{}\cellp3.2} & {\scriptsize{}\celln-9.1} & {\scriptsize{}\celln-2.6} & {\scriptsize{}\cellp5.1} & {\scriptsize{}\cellp2.2} & {\scriptsize{}\celln-4.2} & {\scriptsize{}\celln-4.8} & {\scriptsize{}\celln-7.0} & {\scriptsize{}\cellz-2.0} & {\scriptsize{}\celln-2.6} & {\scriptsize{}\cellz1.2} & {\scriptsize{}\celln-4.1} & {\scriptsize{}\celln-4.9} &  & {\scriptsize{}\celln-3.7}\tabularnewline
\texttt{\textbf{\scriptsize{}EdgeBoxes}} & {\scriptsize{}\cellz-2.0} & {\scriptsize{}\celln-6.1} & {\scriptsize{}\cellz-0.7} & {\scriptsize{}\celln-3.8} & {\scriptsize{}\celln}\textbf{\scriptsize{}-6.7} & {\scriptsize{}\cellz}\textbf{\scriptsize{}0.6} & {\scriptsize{}\celln}\textbf{\scriptsize{}-5.8} & {\scriptsize{}\cellz-1.1} & {\scriptsize{}\cellz}\textbf{\scriptsize{}-2.0} & {\scriptsize{}\cellz-1.8} & {\scriptsize{}\celln-4.6} & {\scriptsize{}\cellz0.4} & {\scriptsize{}\cellz-1.3} & {\scriptsize{}\cellz}\textbf{\scriptsize{}-1.3} & {\scriptsize{}\celln-3.0} & {\scriptsize{}\cellz-1.7} & {\scriptsize{}\cellz}\textbf{\scriptsize{}-0.1} & {\scriptsize{}\cellz-0.9} & {\scriptsize{}\cellz-0.2} & {\scriptsize{}\cellz-1.1} &  & {\scriptsize{}\celln}\textbf{\scriptsize{}-2.2}\tabularnewline
\texttt{\scriptsize{}Endres} & {\scriptsize{}\cellz}\foreignlanguage{english}{{\scriptsize{}-1.5}} & {\scriptsize{}\celln}\foreignlanguage{english}{\textbf{\scriptsize{}-5.8}} & {\scriptsize{}\cellz}\foreignlanguage{english}{{\scriptsize{}-0.6}} & {\scriptsize{}\celln}\foreignlanguage{english}{{\scriptsize{}-4.8}} & {\scriptsize{}\celln}\foreignlanguage{english}{{\scriptsize{}-12.7}} & {\scriptsize{}\cellz}\foreignlanguage{english}{{\scriptsize{}-1.1}} & {\scriptsize{}\celln}\foreignlanguage{english}{{\scriptsize{}-7.1}} & {\scriptsize{}\cellp}\foreignlanguage{english}{{\scriptsize{}3.4}} & {\scriptsize{}\celln}\foreignlanguage{english}{{\scriptsize{}-6.9}} & {\scriptsize{}\celln}\foreignlanguage{english}{{\scriptsize{}-3.2}} & {\scriptsize{}\cellp}\foreignlanguage{english}{{\scriptsize{}4.7}} & {\scriptsize{}\cellz}\foreignlanguage{english}{{\scriptsize{}1.9}} & {\scriptsize{}\celln}\foreignlanguage{english}{{\scriptsize{}-2.4}} & {\scriptsize{}\celln}\foreignlanguage{english}{{\scriptsize{}-2.4}} & {\scriptsize{}\celln}\foreignlanguage{english}{{\scriptsize{}-7.7}} & {\scriptsize{}\celln}\foreignlanguage{english}{{\scriptsize{}-2.8}} & {\scriptsize{}\cellz}\foreignlanguage{english}{{\scriptsize{}-1.9}} & {\scriptsize{}\cellz}\foreignlanguage{english}{{\scriptsize{}1.5}} & {\scriptsize{}\cellz}\foreignlanguage{english}{{\scriptsize{}0.4}} & {\scriptsize{}\celln}\foreignlanguage{english}{{\scriptsize{}-4.2}} &  & {\scriptsize{}\celln}\foreignlanguage{english}{{\scriptsize{}-2.7}}\tabularnewline
\texttt{\textbf{\scriptsize{}Geodesic}} & {\scriptsize{}\cellz-1.9} & {\scriptsize{}\celln-8.1} & {\scriptsize{}\cellz-0.2} & {\scriptsize{}\celln-4.6} & {\scriptsize{}\celln-14.4} & {\scriptsize{}\cellz0.6} & {\scriptsize{}\celln-6.5} & {\scriptsize{}\cellp2.6} & {\scriptsize{}\celln-7.3} & {\scriptsize{}\cellz}\textbf{\scriptsize{}-1.3} & {\scriptsize{}\cellp4.7} & {\scriptsize{}\cellp2.4} & {\scriptsize{}\celln-2.5} & {\scriptsize{}\celln-2.7} & {\scriptsize{}\celln-4.7} & {\scriptsize{}\cellz-1.2} & {\scriptsize{}\cellz-0.7} & {\scriptsize{}\cellz-0.1} & {\scriptsize{}\cellz}\textbf{\scriptsize{}1.9} & {\scriptsize{}\cellz0.2} &  & {\scriptsize{}\celln}\textbf{\scriptsize{}-2.2}\tabularnewline
\texttt{\textbf{\scriptsize{}MCG}} & {\scriptsize{}\cellz-0.7} & {\scriptsize{}\celln-7.2} & {\scriptsize{}\cellz0.1} & {\scriptsize{}\celln-3.6} & {\scriptsize{}\celln}\textbf{\scriptsize{}-6.7} & {\scriptsize{}\cellz-1.2} & {\scriptsize{}\celln-7.0} & {\scriptsize{}\cellp3.4} & {\scriptsize{}\celln-3.2} & {\scriptsize{}\celln-2.3} & {\scriptsize{}\cellp5.0} & {\scriptsize{}\cellz1.9} & {\scriptsize{}\celln-3.5} & {\scriptsize{}\cellz}\textbf{\scriptsize{}-1.3} & {\scriptsize{}\cellz}\textbf{\scriptsize{}-1.5} & {\scriptsize{}\cellz}\textbf{\scriptsize{}-1.1} & {\scriptsize{}\cellz-1.3} & {\scriptsize{}\cellp2.2} & {\scriptsize{}\cellz0.3} & {\scriptsize{}\cellz}\textbf{\scriptsize{}0.5} &  & {\scriptsize{}\cellz}\textbf{\scriptsize{}-1.4}\tabularnewline
\texttt{\scriptsize{}Objectness} & {\scriptsize{}\celln-10.3} & {\scriptsize{}\celln-15.1} & {\scriptsize{}\cellz-2.0} & {\scriptsize{}\celln-6.2} & {\scriptsize{}\celln-11.0} & {\scriptsize{}\celln-9.5} & {\scriptsize{}\celln-13.0} & {\scriptsize{}\celln-3.6} & {\scriptsize{}\celln-10.0} & {\scriptsize{}\celln-6.4} & {\scriptsize{}\celln-7.8} & {\scriptsize{}\cellz-1.0} & {\scriptsize{}\celln-11.6} & {\scriptsize{}\celln-15.9} & {\scriptsize{}\celln-13.0} & {\scriptsize{}\celln-2.7} & {\scriptsize{}\celln-5.8} & {\scriptsize{}\celln-11.2} & {\scriptsize{}\celln-10.9} & {\scriptsize{}\celln-12.9} &  & {\scriptsize{}\celln-9.0}\tabularnewline
\texttt{\scriptsize{}Rahtu} & {\scriptsize{}\cellz-0.3} & {\scriptsize{}\celln-13.2} & {\scriptsize{}\cellz-0.3} & {\scriptsize{}\cellz}\textbf{\scriptsize{}-1.2} & {\scriptsize{}\celln-13.0} & {\scriptsize{}\cellz-0.6} & {\scriptsize{}\celln-12.0} & {\scriptsize{}\cellp3.3} & {\scriptsize{}\celln-10.5} & {\scriptsize{}\celln-4.3} & {\scriptsize{}\cellp2.0} & {\scriptsize{}\cellp2.1} & {\scriptsize{}\celln-3.2} & {\scriptsize{}\celln-4.9} & {\scriptsize{}\celln-7.9} & {\scriptsize{}\celln-2.8} & {\scriptsize{}\celln-4.9} & {\scriptsize{}\celln-5.0} & {\scriptsize{}\cellz0.0} & {\scriptsize{}\celln-3.7} &  & {\scriptsize{}\celln-4.0}\tabularnewline
\texttt{\scriptsize{}Rand.Prim} & {\scriptsize{}\cellp}\textbf{\scriptsize{}2.1} & {\scriptsize{}\celln-10.4} & {\scriptsize{}\cellz-0.5} & {\scriptsize{}\celln-4.5} & {\scriptsize{}\celln-13.2} & {\scriptsize{}\cellz-1.9} & {\scriptsize{}\celln-10.1} & {\scriptsize{}\cellp5.0} & {\scriptsize{}\celln-6.7} & {\scriptsize{}\celln-3.5} & {\scriptsize{}\cellp2.0} & {\scriptsize{}\cellp2.4} & {\scriptsize{}\celln-4.4} & {\scriptsize{}\celln-5.1} & {\scriptsize{}\celln-10.0} & {\scriptsize{}\celln-2.3} & {\scriptsize{}\cellz-1.8} & {\scriptsize{}\cellz1.2} & {\scriptsize{}\celln-3.8} & {\scriptsize{}\celln-4.4} &  & {\scriptsize{}\celln-3.5}\tabularnewline
\texttt{\scriptsize{}Rantalankila} & {\scriptsize{}\cellz}\foreignlanguage{english}{{\scriptsize{}0.5}} & {\scriptsize{}\celln}\foreignlanguage{english}{{\scriptsize{}-13.6}} & {\scriptsize{}\cellz}\foreignlanguage{english}{{\scriptsize{}0.3}} & {\scriptsize{}\celln}\foreignlanguage{english}{{\scriptsize{}-3.0}} & {\scriptsize{}\celln}\foreignlanguage{english}{{\scriptsize{}-12.9}} & {\scriptsize{}\celln}\foreignlanguage{english}{{\scriptsize{}-3.6}} & {\scriptsize{}\celln}\foreignlanguage{english}{{\scriptsize{}-9.0}} & {\scriptsize{}\cellp}\foreignlanguage{english}{{\scriptsize{}4.4}} & {\scriptsize{}\celln}\foreignlanguage{english}{{\scriptsize{}-5.6}} & {\scriptsize{}\celln}\foreignlanguage{english}{{\scriptsize{}-3.7}} & {\scriptsize{}\cellp}\foreignlanguage{english}{{\scriptsize{}4.1}} & {\scriptsize{}\cellp}\foreignlanguage{english}{{\scriptsize{}2.5}} & {\scriptsize{}\celln}\foreignlanguage{english}{{\scriptsize{}-2.2}} & {\scriptsize{}\celln}\foreignlanguage{english}{{\scriptsize{}-4.0}} & {\scriptsize{}\celln}\foreignlanguage{english}{{\scriptsize{}-7.8}} & {\scriptsize{}\celln}\foreignlanguage{english}{{\scriptsize{}-2.5}} & {\scriptsize{}\celln}\foreignlanguage{english}{{\scriptsize{}-3.8}} & {\scriptsize{}\cellp}\foreignlanguage{english}{\textbf{\scriptsize{}2.1}} & {\scriptsize{}\cellz}\foreignlanguage{english}{{\scriptsize{}-1.5}} & {\scriptsize{}\cellz}\foreignlanguage{english}{{\scriptsize{}-0.7}} &  & {\scriptsize{}\celln}\foreignlanguage{english}{{\scriptsize{}-3.0}}\tabularnewline
\texttt{\textbf{\scriptsize{}Rigor}} & {\scriptsize{}\cellz1.7} & {\scriptsize{}\celln-7.9} & {\scriptsize{}\cellz0.5} & {\scriptsize{}\celln-4.1} & {\scriptsize{}\celln-12.4} & {\scriptsize{}\cellz-0.8} & {\scriptsize{}\celln-9.0} & {\scriptsize{}\cellp}\textbf{\scriptsize{}6.3} & {\scriptsize{}\celln-6.9} & {\scriptsize{}\cellz-1.7} & {\scriptsize{}\cellz1.8} & {\scriptsize{}\cellp}\textbf{\scriptsize{}2.9} & {\scriptsize{}\cellz}\textbf{\scriptsize{}-0.9} & {\scriptsize{}\celln-3.3} & {\scriptsize{}\celln-7.7} & {\scriptsize{}\cellz-1.8} & {\scriptsize{}\cellz-1.3} & {\scriptsize{}\cellz1.6} & {\scriptsize{}\cellz-1.2} & {\scriptsize{}\cellz-1.7} &  & {\scriptsize{}\celln}\textbf{\scriptsize{}-2.3}\tabularnewline
\texttt{\textbf{\scriptsize{}SelectiveSearch}} & {\scriptsize{}\cellz1.3} & {\scriptsize{}\celln-7.7} & {\scriptsize{}\cellz}\textbf{\scriptsize{}1.0} & {\scriptsize{}\celln-4.3} & {\scriptsize{}\celln-11.1} & {\scriptsize{}\cellz-1.7} & {\scriptsize{}\celln-7.8} & {\scriptsize{}\cellp3.9} & {\scriptsize{}\celln-4.8} & {\scriptsize{}\cellz-1.5} & {\scriptsize{}\cellp}\textbf{\scriptsize{}5.4} & {\scriptsize{}\cellp2.2} & {\scriptsize{}\cellz-1.4} & {\scriptsize{}\celln-3.8} & {\scriptsize{}\celln-6.0} & {\scriptsize{}\cellz-1.5} & {\scriptsize{}\cellz-0.8} & {\scriptsize{}\cellz0.6} & {\scriptsize{}\celln-2.4} & {\scriptsize{}\celln-2.1} &  & {\scriptsize{}\celln}\textbf{\scriptsize{}-2.1}\tabularnewline
\hline 
\texttt{\scriptsize{}Gaussian} & {\scriptsize{}\celln-6.6} & {\scriptsize{}\celln-13.4} & {\scriptsize{}\cellz-0.7} & {\scriptsize{}\celln-4.4} & {\scriptsize{}\celln-15.0} & {\scriptsize{}\celln-6.1} & {\scriptsize{}\celln-16.0} & {\scriptsize{}\cellz0.9} & {\scriptsize{}\celln-9.1} & {\scriptsize{}\celln-8.0} & {\scriptsize{}\cellz0.3} & {\scriptsize{}\cellz1.2} & {\scriptsize{}\celln-4.2} & {\scriptsize{}\celln-6.9} & {\scriptsize{}\celln-10.3} & {\scriptsize{}\celln-2.3} & {\scriptsize{}\celln-6.5} & {\scriptsize{}\celln-4.5} & {\scriptsize{}\celln-3.6} & {\scriptsize{}\celln-12.1} &  & {\scriptsize{}\celln}\emph{\scriptsize{}-6.4}\tabularnewline
\texttt{\scriptsize{}SlidingWindow} & {\scriptsize{}\celln-21.8} & {\scriptsize{}\celln-20.7} & {\scriptsize{}\celln-3.2} & {\scriptsize{}\celln-8.1} & {\scriptsize{}\celln-16.6} & {\scriptsize{}\celln-14.7} & {\scriptsize{}\celln-22.1} & {\scriptsize{}\cellz-0.7} & {\scriptsize{}\celln-9.8} & {\scriptsize{}\celln-11.7} & {\scriptsize{}\celln-10.2} & {\scriptsize{}\cellz-1.4} & {\scriptsize{}\celln-14.7} & {\scriptsize{}\celln-20.1} & {\scriptsize{}\celln-14.8} & {\scriptsize{}\celln-3.8} & {\scriptsize{}\celln-7.7} & {\scriptsize{}\celln-21.0} & {\scriptsize{}\celln-20.8} & {\scriptsize{}\celln-14.8} &  & {\scriptsize{}\celln-12.9}\tabularnewline
\texttt{\scriptsize{}Superpixels} & {\scriptsize{}\celln-23.9} & {\scriptsize{}\celln-52.2} & {\scriptsize{}\celln-3.1} & {\scriptsize{}\celln-9.4} & {\scriptsize{}\celln-17.4} & {\scriptsize{}\celln-43.9} & {\scriptsize{}\celln-42.3} & {\scriptsize{}\celln-10.2} & {\scriptsize{}\celln-11.3} & {\scriptsize{}\celln-12.6} & {\scriptsize{}\celln-15.8} & {\scriptsize{}\celln-8.5} & {\scriptsize{}\celln-50.1} & {\scriptsize{}\celln-41.7} & {\scriptsize{}\celln-30.9} & {\scriptsize{}\celln-4.4} & {\scriptsize{}\celln-10.6} & {\scriptsize{}\celln-25.2} & {\scriptsize{}\celln-39.7} & {\scriptsize{}\celln-8.2} &  & {\scriptsize{}\celln-23.1}\tabularnewline
\texttt{\scriptsize{}Uniform} & {\scriptsize{}\celln}\foreignlanguage{english}{{\scriptsize{}-3.2}} & {\scriptsize{}\celln}\foreignlanguage{english}{{\scriptsize{}-18.8}} & {\scriptsize{}\celln}\foreignlanguage{english}{{\scriptsize{}-4.0}} & {\scriptsize{}\celln}\foreignlanguage{english}{{\scriptsize{}-4.8}} & {\scriptsize{}\celln}\foreignlanguage{english}{{\scriptsize{}-15.2}} & {\scriptsize{}\celln}\foreignlanguage{english}{{\scriptsize{}-8.6}} & {\scriptsize{}\celln}\foreignlanguage{english}{{\scriptsize{}-16.6}} & {\scriptsize{}\cellz}\foreignlanguage{english}{{\scriptsize{}0.2}} & {\scriptsize{}\celln}\foreignlanguage{english}{{\scriptsize{}-10.4}} & {\scriptsize{}\celln}\foreignlanguage{english}{{\scriptsize{}-8.8}} & {\scriptsize{}\cellp}\foreignlanguage{english}{{\scriptsize{}3.7}} & {\scriptsize{}\cellz}\foreignlanguage{english}{{\scriptsize{}1.3}} & {\scriptsize{}\celln}\foreignlanguage{english}{{\scriptsize{}-6.6}} & {\scriptsize{}\celln}\foreignlanguage{english}{{\scriptsize{}-11.3}} & {\scriptsize{}\celln}\foreignlanguage{english}{{\scriptsize{}-10.2}} & {\scriptsize{}\celln}\foreignlanguage{english}{{\scriptsize{}-3.6}} & {\scriptsize{}\celln}\foreignlanguage{english}{{\scriptsize{}-8.9}} & {\scriptsize{}\celln}\foreignlanguage{english}{{\scriptsize{}-5.8}} & {\scriptsize{}\celln}\foreignlanguage{english}{{\scriptsize{}-5.1}} & {\scriptsize{}\celln}\foreignlanguage{english}{{\scriptsize{}-20.2}} &  & {\scriptsize{}\celln}\foreignlanguage{english}{{\scriptsize{}-7.8}}\tabularnewline
\hline 
{\scriptsize{}Top methods avg.} & {\scriptsize{}\cellz-0.3} & {\scriptsize{}\celln-7.4} & {\scriptsize{}\cellz0.1} & {\scriptsize{}\celln-4.1} & {\scriptsize{}\celln-10.2} & {\scriptsize{}\cellz-0.5} & {\scriptsize{}\celln-7.2} & {\scriptsize{}\cellp3.0} & {\scriptsize{}\celln-4.8} & {\scriptsize{}\cellz-1.7} & {\scriptsize{}\cellp2.5} & {\scriptsize{}\cellp2.0} & {\scriptsize{}\cellz-1.9} & {\scriptsize{}\celln-2.5} & {\scriptsize{}\celln-4.6} & {\scriptsize{}\cellz-1.5} & {\scriptsize{}\cellz-0.8} & {\scriptsize{}\cellz0.7} & {\scriptsize{}\cellz-0.3} & {\scriptsize{}\cellz-0.8} &  & {\scriptsize{}\cellz-2.0}\tabularnewline
\end{tabular}\hspace*{\fill}

\protect\caption{\label{tab:pascal-llda-dpm-per-class}LM-LLDA detection results on
PASCAL 2007 (with bounding box regression). The top row indicates
the average precision (AP) of LM-LLDA alone, while the other rows
show the difference in AP when adding proposal methods. Green indicates
improvement of at least $2\,\mbox{AP}$, blue indicates minor change
($-2\le\text{AP}<2$), and white indicates a decrease by more than
$2\,\mbox{AP}$. \texttt{Edge\-Boxes} achieves top results on 6 of
the 20 categories; \texttt{MCG} performs best overall with -1.4 mAP
loss.}
\end{table*}
\begin{table}
\begin{centering}
{\scriptsize{}}%
\begin{tabular}{ccccc}
{\scriptsize{}Proposals} & {\scriptsize{}LM-LLDA} & {\scriptsize{}R-CNN} & {\scriptsize{}Fast R-CNN} & {\scriptsize{}$\ensuremath{\Delta}$Train}\tabularnewline
\hline 
\texttt{\footnotesize{}Dense} & 33.5/34.4 & -- & -- & --\tabularnewline
\hline 
\selectlanguage{english}%
\texttt{Bing}\selectlanguage{british}%
 & \selectlanguage{english}%
21.8/22.4\selectlanguage{british}%
 & \selectlanguage{english}%
36.7\selectlanguage{british}%
 & \selectlanguage{english}%
37.3/49.0\selectlanguage{british}%
 & \textbf{\cellz}\foreignlanguage{english}{+6.3}\tabularnewline
\selectlanguage{english}%
\texttt{CPMC}\selectlanguage{british}%
 & \selectlanguage{english}%
30.0/30.7\selectlanguage{british}%
 & \selectlanguage{english}%
51.7\selectlanguage{british}%
 & \selectlanguage{english}%
53.7/57.1\selectlanguage{british}%
 & \selectlanguage{english}%
-1.3\selectlanguage{british}%
\tabularnewline
\selectlanguage{english}%
\texttt{EdgeBoxes}\selectlanguage{british}%
 & \textbf{\cellp}\foreignlanguage{english}{31.8/32.2} & \selectlanguage{english}%
53.0\selectlanguage{british}%
 & \textbf{\cellp}\foreignlanguage{english}{55.4/\textbf{60.4}} & \textbf{\cellz}\foreignlanguage{english}{+3.3}\tabularnewline
\selectlanguage{english}%
\texttt{Endres}\selectlanguage{british}%
 & \selectlanguage{english}%
31.2/31.7\selectlanguage{british}%
 & \selectlanguage{english}%
52.8\selectlanguage{british}%
 & \selectlanguage{english}%
54.2/57.4\selectlanguage{british}%
 & \selectlanguage{english}%
-0.2\selectlanguage{british}%
\tabularnewline
\selectlanguage{english}%
\texttt{Geodesic}\selectlanguage{british}%
 & \textbf{\cellp}\foreignlanguage{english}{31.8/32.2} & \textbf{\cellp}\foreignlanguage{english}{53.8} & \selectlanguage{english}%
53.6/57.5\selectlanguage{british}%
 & \selectlanguage{english}%
-0.4\selectlanguage{british}%
\tabularnewline
\selectlanguage{english}%
\texttt{MCG}\selectlanguage{british}%
 & \textbf{\cellp}\foreignlanguage{english}{\textbf{32.5}/\textbf{33.0}} & \textbf{\cellp}\foreignlanguage{english}{\textbf{56.5}} & \textbf{\cellp}\foreignlanguage{english}{\textbf{58.1}/60.3} & \selectlanguage{english}%
+1.8\selectlanguage{british}%
\tabularnewline
\selectlanguage{english}%
\texttt{Objectness}\selectlanguage{british}%
 & \selectlanguage{english}%
25.0/25.4\selectlanguage{british}%
 & \selectlanguage{english}%
39.7\selectlanguage{british}%
 & \selectlanguage{english}%
41.5/51.4\selectlanguage{british}%
 & \textbf{\cellz}\foreignlanguage{english}{+9.1}\tabularnewline
\selectlanguage{english}%
\texttt{Rahtu}\selectlanguage{british}%
 & \selectlanguage{english}%
29.6/30.4\selectlanguage{british}%
 & \selectlanguage{english}%
46.1\selectlanguage{british}%
 & \selectlanguage{english}%
48.9/53.6\selectlanguage{british}%
 & \selectlanguage{english}%
+0.7\selectlanguage{british}%
\tabularnewline
\selectlanguage{english}%
\texttt{RandomizedPrims}\selectlanguage{british}%
 & \selectlanguage{english}%
30.5/30.9\selectlanguage{british}%
 & \selectlanguage{english}%
51.6\selectlanguage{british}%
 & \selectlanguage{english}%
53.2/57.6\selectlanguage{british}%
 & \selectlanguage{english}%
-0.6\selectlanguage{british}%
\tabularnewline
\selectlanguage{english}%
\texttt{Rantalankila}\selectlanguage{british}%
 & \selectlanguage{english}%
30.9/31.4\selectlanguage{british}%
 & \textbf{\cellp}\foreignlanguage{english}{53.1} & \textbf{\cellp}\foreignlanguage{english}{55.0/57.9} & \selectlanguage{english}%
-0.5\selectlanguage{british}%
\tabularnewline
\selectlanguage{english}%
\texttt{Rigor}\selectlanguage{british}%
 & \textbf{\cellp}\foreignlanguage{english}{31.5/32.1} & \textbf{\cellp}\foreignlanguage{english}{54.1} & \textbf{\cellp}\foreignlanguage{english}{55.4/58.4} & \selectlanguage{english}%
-0.2\selectlanguage{british}%
\tabularnewline
\selectlanguage{english}%
\texttt{SelectiveSearch}\selectlanguage{british}%
 & \textbf{\cellp}\foreignlanguage{english}{31.7/32.3} & \textbf{\cellp}\foreignlanguage{english}{54.6} & \textbf{\cellp}\foreignlanguage{english}{56.3/59.5} & \selectlanguage{english}%
+0.0\selectlanguage{british}%
\tabularnewline
\hline 
\selectlanguage{english}%
\texttt{Gaussian}\selectlanguage{british}%
 & \selectlanguage{english}%
\textit{27.3}/\textit{28.0}\selectlanguage{british}%
 & \selectlanguage{english}%
\textit{40.6}\selectlanguage{british}%
 & \selectlanguage{english}%
\textit{44.6}/\textit{50.8}\selectlanguage{british}%
 & \selectlanguage{english}%
+0.8\selectlanguage{british}%
\tabularnewline
\selectlanguage{english}%
\texttt{Sliding window}\selectlanguage{british}%
 & \selectlanguage{english}%
20.7/21.5\selectlanguage{british}%
 & \selectlanguage{english}%
32.7\selectlanguage{british}%
 & \selectlanguage{english}%
32.7/44.8\selectlanguage{british}%
 & \textbf{\cellz}\foreignlanguage{english}{+3.3}\tabularnewline
\selectlanguage{english}%
\texttt{Superpixels}\selectlanguage{british}%
 & \selectlanguage{english}%
11.2/11.3\selectlanguage{british}%
 & \selectlanguage{english}%
17.6\selectlanguage{british}%
 & \selectlanguage{english}%
15.4/20.3\selectlanguage{british}%
 & \selectlanguage{english}%
-2.0\selectlanguage{british}%
\tabularnewline
\selectlanguage{english}%
\texttt{Uniform}\selectlanguage{british}%
 & \selectlanguage{english}%
26.0/26.6\selectlanguage{british}%
 & \selectlanguage{english}%
37.3\selectlanguage{british}%
 & \selectlanguage{english}%
39.5/46.9\selectlanguage{british}%
 & \selectlanguage{english}%
-0.1\selectlanguage{british}%
\tabularnewline
\end{tabular}
\par\end{centering}{\scriptsize \par}

\protect\caption{\label{tab:pascal-mAP}Mean average precision (mAP) on PASCAL 2007
for multiple detectors and proposal methods (using $1\,000$ proposals).
LM-LLDA and Fast R-CNN results shown before/after bounding box regression.
The final column shows the change in mAP obtained from re-training
Fast R-CNN (with box regression) for the specific proposal method.}
\end{table}

\paragraph*{Score vs IoU}

In figure \ref{fig:detector-scoremaps} we show average detection
scores for proposals with varying IoU overlap with the ground truth.
The scores have been scaled between zero and one per class before
averaging across classes. The drop of the LM-LLDA scores at high overlaps
is due to a bias introduced during training by the latent location
estimation on positive samples; this bias is compensated for by the
subsequent bounding box prediction stage of LM-LLDA. For Fast R-CNN,
the bounding box prediction effectively improves proposal IoU with
the ground truth and results in a substantial shift of the curve to
the right.

\paragraph*{Localisation}

We observe from figure \ref{fig:detector-scoremaps} that both LM-LLDA
and R-CNN exhibit an almost linear increase in detection score as
IoU increases (especially between 0.4 and 0.8 IoU). From this we conclude
that there is no IoU threshold that is ``sufficiently good'' for
obtaining top detection quality. We thus consider that improving localisation
of proposals is as important as increasing ground truth recall, and
the linear relation helps motivate us to linearly reward localisation
in the average recall metric (see \S\ref{sub:Recall-results}). For
Fast R-CNN there is also an almost linear relation, but performance
saturates earlier. Thus, Fast R-CNN is likely to also benefit from
better localisation, but up to a point.

\subsection{\label{sub:LLDA-DPM}LM-LLDA detection performance}

\begin{table*}
\setlength\tabcolsep{2pt}

\hspace*{\fill}%
\begin{tabular}{ccccccccccccccccccccccc}
 & {\scriptsize{}aero} & {\scriptsize{}bicycle} & {\scriptsize{}bird} & {\scriptsize{}boat} & {\scriptsize{}bottle} & {\scriptsize{}bus} & {\scriptsize{}car} & {\scriptsize{}cat} & {\scriptsize{}chair} & {\scriptsize{}cow} & {\scriptsize{}table} & {\scriptsize{}dog} & {\scriptsize{}horse} & {\scriptsize{}mbike} & {\scriptsize{}person} & {\scriptsize{}plant} & {\scriptsize{}sheep} & {\scriptsize{}sofa} & {\scriptsize{}train} & {\scriptsize{}tv} & {\scriptsize{}\quad{}} & {\scriptsize{}mean}\tabularnewline
\hline 
\texttt{\scriptsize{}Bing} & \selectlanguage{english}%
{\scriptsize{}56.6}\selectlanguage{british}%
 & \selectlanguage{english}%
{\scriptsize{}54.9}\selectlanguage{british}%
 & \selectlanguage{english}%
{\scriptsize{}45.0}\selectlanguage{british}%
 & \selectlanguage{english}%
{\scriptsize{}28.6}\selectlanguage{british}%
 & \selectlanguage{english}%
{\scriptsize{}24.6}\selectlanguage{british}%
 & \selectlanguage{english}%
{\scriptsize{}53.9}\selectlanguage{british}%
 & \selectlanguage{english}%
{\scriptsize{}63.5}\selectlanguage{british}%
 & \selectlanguage{english}%
{\scriptsize{}72.5}\selectlanguage{british}%
 & \selectlanguage{english}%
{\scriptsize{}15.6}\selectlanguage{british}%
 & \selectlanguage{english}%
{\scriptsize{}59.4}\selectlanguage{british}%
 & \selectlanguage{english}%
{\scriptsize{}49.0}\selectlanguage{british}%
 & \selectlanguage{english}%
{\scriptsize{}59.7}\selectlanguage{british}%
 & \selectlanguage{english}%
{\scriptsize{}68.5}\selectlanguage{british}%
 & \selectlanguage{english}%
{\scriptsize{}60.3}\selectlanguage{british}%
 & \selectlanguage{english}%
{\scriptsize{}50.7}\selectlanguage{british}%
 & \selectlanguage{english}%
{\scriptsize{}16.5}\selectlanguage{british}%
 & \selectlanguage{english}%
{\scriptsize{}49.0}\selectlanguage{british}%
 & \selectlanguage{english}%
{\scriptsize{}42.8}\selectlanguage{british}%
 & \selectlanguage{english}%
{\scriptsize{}64.8}\selectlanguage{british}%
 & \selectlanguage{english}%
{\scriptsize{}44.9}\selectlanguage{british}%
 &  & \selectlanguage{english}%
{\scriptsize{}49.0}\selectlanguage{british}%
\tabularnewline
\texttt{\scriptsize{}CPMC} & \selectlanguage{english}%
{\scriptsize{}65.2}\selectlanguage{british}%
 & \selectlanguage{english}%
{\scriptsize{}61.8}\selectlanguage{british}%
 & \selectlanguage{english}%
{\scriptsize{}58.2}\selectlanguage{british}%
 & \selectlanguage{english}%
{\scriptsize{}37.2}\selectlanguage{british}%
 & \selectlanguage{english}%
{\scriptsize{}17.9}\selectlanguage{british}%
 & \selectlanguage{english}%
{\scriptsize{}71.0}\selectlanguage{british}%
 & \selectlanguage{english}%
{\scriptsize{}67.3}\selectlanguage{british}%
 & \selectlanguage{english}%
{\scriptsize{}\cellp76.7}\selectlanguage{british}%
 & \selectlanguage{english}%
{\scriptsize{}22.9}\selectlanguage{british}%
 & \selectlanguage{english}%
{\scriptsize{}61.2}\selectlanguage{british}%
 & \selectlanguage{english}%
{\scriptsize{}64.6}\selectlanguage{british}%
 & \selectlanguage{english}%
{\scriptsize{}\cellp70.1}\selectlanguage{british}%
 & \selectlanguage{english}%
{\scriptsize{}77.0}\selectlanguage{british}%
 & \selectlanguage{english}%
{\scriptsize{}69.2}\selectlanguage{british}%
 & \selectlanguage{english}%
{\scriptsize{}54.8}\selectlanguage{british}%
 & \selectlanguage{english}%
{\scriptsize{}18.5}\selectlanguage{british}%
 & \selectlanguage{english}%
{\scriptsize{}52.6}\selectlanguage{british}%
 & \selectlanguage{english}%
{\scriptsize{}63.4}\selectlanguage{british}%
 & \selectlanguage{english}%
{\scriptsize{}71.7}\selectlanguage{british}%
 & \selectlanguage{english}%
{\scriptsize{}\cellp61.5}\selectlanguage{british}%
 &  & \selectlanguage{english}%
{\scriptsize{}57.1}\selectlanguage{british}%
\tabularnewline
\texttt{\scriptsize{}EdgeBoxes} & \selectlanguage{english}%
{\scriptsize{}67.0}\selectlanguage{british}%
 & \selectlanguage{english}%
{\scriptsize{}\cellp69.9}\selectlanguage{british}%
 & \selectlanguage{english}%
{\scriptsize{}\cellp59.8}\selectlanguage{british}%
 & \selectlanguage{english}%
{\scriptsize{}\cellp}\textbf{\scriptsize{}46.1}\selectlanguage{british}%
 & \selectlanguage{english}%
{\scriptsize{}\cellp28.3}\selectlanguage{british}%
 & \selectlanguage{english}%
{\scriptsize{}\cellp72.9}\selectlanguage{british}%
 & \selectlanguage{english}%
{\scriptsize{}\cellp}\textbf{\scriptsize{}72.3}\selectlanguage{british}%
 & \selectlanguage{english}%
{\scriptsize{}73.8}\selectlanguage{british}%
 & \selectlanguage{english}%
{\scriptsize{}\cellp28.8}\selectlanguage{british}%
 & \selectlanguage{english}%
{\scriptsize{}\cellp}\textbf{\scriptsize{}68.1}\selectlanguage{british}%
 & \selectlanguage{english}%
{\scriptsize{}62.4}\selectlanguage{british}%
 & \selectlanguage{english}%
{\scriptsize{}67.6}\selectlanguage{british}%
 & \selectlanguage{english}%
{\scriptsize{}\cellp}\textbf{\scriptsize{}79.2}\selectlanguage{british}%
 & \selectlanguage{english}%
{\scriptsize{}\cellp}\textbf{\scriptsize{}73.6}\selectlanguage{british}%
 & \selectlanguage{english}%
{\scriptsize{}\cellp}\textbf{\scriptsize{}62.4}\selectlanguage{british}%
 & \selectlanguage{english}%
{\scriptsize{}\cellp}\textbf{\scriptsize{}28.2}\selectlanguage{british}%
 & \selectlanguage{english}%
{\scriptsize{}\cellp55.8}\selectlanguage{british}%
 & \selectlanguage{english}%
{\scriptsize{}61.2}\selectlanguage{british}%
 & \selectlanguage{english}%
{\scriptsize{}70.4}\selectlanguage{british}%
 & \selectlanguage{english}%
{\scriptsize{}59.7}\selectlanguage{british}%
 &  & \selectlanguage{english}%
{\scriptsize{}\cellp}\textbf{\scriptsize{}60.4}\selectlanguage{british}%
\tabularnewline
\texttt{\scriptsize{}Endres} & \selectlanguage{english}%
{\scriptsize{}61.5}\selectlanguage{british}%
 & \selectlanguage{english}%
{\scriptsize{}\cellp}\textbf{\scriptsize{}70.8}\selectlanguage{british}%
 & \selectlanguage{english}%
{\scriptsize{}57.1}\selectlanguage{british}%
 & \selectlanguage{english}%
{\scriptsize{}33.5}\selectlanguage{british}%
 & \selectlanguage{english}%
{\scriptsize{}18.0}\selectlanguage{british}%
 & \selectlanguage{english}%
{\scriptsize{}\cellp72.5}\selectlanguage{british}%
 & \selectlanguage{english}%
{\scriptsize{}68.8}\selectlanguage{british}%
 & \selectlanguage{english}%
{\scriptsize{}\cellp77.3}\selectlanguage{british}%
 & \selectlanguage{english}%
{\scriptsize{}21.7}\selectlanguage{british}%
 & \selectlanguage{english}%
{\scriptsize{}61.8}\selectlanguage{british}%
 & \selectlanguage{english}%
{\scriptsize{}64.5}\selectlanguage{british}%
 & \selectlanguage{english}%
{\scriptsize{}68.2}\selectlanguage{british}%
 & \selectlanguage{english}%
{\scriptsize{}\cellp78.0}\selectlanguage{british}%
 & \selectlanguage{english}%
{\scriptsize{}69.9}\selectlanguage{british}%
 & \selectlanguage{english}%
{\scriptsize{}56.2}\selectlanguage{british}%
 & \selectlanguage{english}%
{\scriptsize{}21.4}\selectlanguage{british}%
 & \selectlanguage{english}%
{\scriptsize{}54.5}\selectlanguage{british}%
 & \selectlanguage{english}%
{\scriptsize{}63.2}\selectlanguage{british}%
 & \selectlanguage{english}%
{\scriptsize{}72.4}\selectlanguage{british}%
 & \selectlanguage{english}%
{\scriptsize{}56.9}\selectlanguage{british}%
 &  & \selectlanguage{english}%
{\scriptsize{}57.4}\selectlanguage{british}%
\tabularnewline
\texttt{\scriptsize{}Geodesic} & \selectlanguage{english}%
{\scriptsize{}63.2}\selectlanguage{british}%
 & \selectlanguage{english}%
{\scriptsize{}68.0}\selectlanguage{british}%
 & \selectlanguage{english}%
{\scriptsize{}55.9}\selectlanguage{british}%
 & \selectlanguage{english}%
{\scriptsize{}39.2}\selectlanguage{british}%
 & \selectlanguage{english}%
{\scriptsize{}19.8}\selectlanguage{british}%
 & \selectlanguage{english}%
{\scriptsize{}71.1}\selectlanguage{british}%
 & \selectlanguage{english}%
{\scriptsize{}\cellp70.4}\selectlanguage{british}%
 & \selectlanguage{english}%
{\scriptsize{}74.4}\selectlanguage{british}%
 & \selectlanguage{english}%
{\scriptsize{}24.8}\selectlanguage{british}%
 & \selectlanguage{english}%
{\scriptsize{}65.0}\selectlanguage{british}%
 & \selectlanguage{english}%
{\scriptsize{}63.5}\selectlanguage{british}%
 & \selectlanguage{english}%
{\scriptsize{}65.6}\selectlanguage{british}%
 & \selectlanguage{english}%
{\scriptsize{}\cellp78.7}\selectlanguage{british}%
 & \selectlanguage{english}%
{\scriptsize{}69.2}\selectlanguage{british}%
 & \selectlanguage{english}%
{\scriptsize{}58.0}\selectlanguage{british}%
 & \selectlanguage{english}%
{\scriptsize{}20.4}\selectlanguage{british}%
 & \selectlanguage{english}%
{\scriptsize{}54.5}\selectlanguage{british}%
 & \selectlanguage{english}%
{\scriptsize{}57.8}\selectlanguage{british}%
 & \selectlanguage{english}%
{\scriptsize{}70.2}\selectlanguage{british}%
 & \selectlanguage{english}%
{\scriptsize{}\cellp60.9}\selectlanguage{british}%
 &  & \selectlanguage{english}%
{\scriptsize{}57.5}\selectlanguage{british}%
\tabularnewline
\texttt{\scriptsize{}MCG} & \selectlanguage{english}%
{\scriptsize{}66.6}\selectlanguage{british}%
 & \selectlanguage{english}%
{\scriptsize{}\cellp69.1}\selectlanguage{british}%
 & \selectlanguage{english}%
{\scriptsize{}\cellp60.1}\selectlanguage{british}%
 & \selectlanguage{english}%
{\scriptsize{}42.0}\selectlanguage{british}%
 & \selectlanguage{english}%
{\scriptsize{}\cellp}\textbf{\scriptsize{}28.5}\selectlanguage{british}%
 & \selectlanguage{english}%
{\scriptsize{}71.9}\selectlanguage{british}%
 & \selectlanguage{english}%
{\scriptsize{}\cellp}\textbf{\scriptsize{}72.3}\selectlanguage{british}%
 & \selectlanguage{english}%
{\scriptsize{}\cellp77.3}\selectlanguage{british}%
 & \selectlanguage{english}%
{\scriptsize{}\cellp}\textbf{\scriptsize{}30.2}\selectlanguage{british}%
 & \selectlanguage{english}%
{\scriptsize{}61.3}\selectlanguage{british}%
 & \selectlanguage{english}%
{\scriptsize{}62.4}\selectlanguage{british}%
 & \selectlanguage{english}%
{\scriptsize{}\cellp69.8}\selectlanguage{british}%
 & \selectlanguage{english}%
{\scriptsize{}\cellp77.4}\selectlanguage{british}%
 & \selectlanguage{english}%
{\scriptsize{}68.2}\selectlanguage{british}%
 & \selectlanguage{english}%
{\scriptsize{}\cellp62.2}\selectlanguage{british}%
 & \selectlanguage{english}%
{\scriptsize{}\cellp27.5}\selectlanguage{british}%
 & \selectlanguage{english}%
{\scriptsize{}\cellp}\textbf{\scriptsize{}57.6}\selectlanguage{british}%
 & \selectlanguage{english}%
{\scriptsize{}\cellp}\textbf{\scriptsize{}66.0}\selectlanguage{british}%
 & \selectlanguage{english}%
{\scriptsize{}\cellp75.8}\selectlanguage{british}%
 & \selectlanguage{english}%
{\scriptsize{}59.4}\selectlanguage{british}%
 &  & \selectlanguage{english}%
{\scriptsize{}\cellp60.3}\selectlanguage{british}%
\tabularnewline
\texttt{\scriptsize{}Objectness} & \selectlanguage{english}%
{\scriptsize{}62.4}\selectlanguage{british}%
 & \selectlanguage{english}%
{\scriptsize{}61.5}\selectlanguage{british}%
 & \selectlanguage{english}%
{\scriptsize{}51.0}\selectlanguage{british}%
 & \selectlanguage{english}%
{\scriptsize{}32.0}\selectlanguage{british}%
 & \selectlanguage{english}%
{\scriptsize{}19.3}\selectlanguage{british}%
 & \selectlanguage{english}%
{\scriptsize{}65.8}\selectlanguage{british}%
 & \selectlanguage{english}%
{\scriptsize{}64.3}\selectlanguage{british}%
 & \selectlanguage{english}%
{\scriptsize{}69.5}\selectlanguage{british}%
 & \selectlanguage{english}%
{\scriptsize{}18.0}\selectlanguage{british}%
 & \selectlanguage{english}%
{\scriptsize{}55.4}\selectlanguage{british}%
 & \selectlanguage{english}%
{\scriptsize{}51.4}\selectlanguage{british}%
 & \selectlanguage{english}%
{\scriptsize{}60.1}\selectlanguage{british}%
 & \selectlanguage{english}%
{\scriptsize{}74.1}\selectlanguage{british}%
 & \selectlanguage{english}%
{\scriptsize{}64.7}\selectlanguage{british}%
 & \selectlanguage{english}%
{\scriptsize{}50.9}\selectlanguage{british}%
 & \selectlanguage{english}%
{\scriptsize{}17.3}\selectlanguage{british}%
 & \selectlanguage{english}%
{\scriptsize{}41.9}\selectlanguage{british}%
 & \selectlanguage{english}%
{\scriptsize{}50.9}\selectlanguage{british}%
 & \selectlanguage{english}%
{\scriptsize{}67.8}\selectlanguage{british}%
 & \selectlanguage{english}%
{\scriptsize{}49.0}\selectlanguage{british}%
 &  & \selectlanguage{english}%
{\scriptsize{}51.4}\selectlanguage{british}%
\tabularnewline
\texttt{\scriptsize{}Rahtu} & \selectlanguage{english}%
{\scriptsize{}62.8}\selectlanguage{british}%
 & \selectlanguage{english}%
{\scriptsize{}60.9}\selectlanguage{british}%
 & \selectlanguage{english}%
{\scriptsize{}53.3}\selectlanguage{british}%
 & \selectlanguage{english}%
{\scriptsize{}35.1}\selectlanguage{british}%
 & \selectlanguage{english}%
{\scriptsize{}15.3}\selectlanguage{british}%
 & \selectlanguage{english}%
{\scriptsize{}\cellp72.6}\selectlanguage{british}%
 & \selectlanguage{english}%
{\scriptsize{}60.5}\selectlanguage{british}%
 & \selectlanguage{english}%
{\scriptsize{}75.1}\selectlanguage{british}%
 & \selectlanguage{english}%
{\scriptsize{}15.4}\selectlanguage{british}%
 & \selectlanguage{english}%
{\scriptsize{}56.9}\selectlanguage{british}%
 & \selectlanguage{english}%
{\scriptsize{}61.6}\selectlanguage{british}%
 & \selectlanguage{english}%
{\scriptsize{}66.3}\selectlanguage{british}%
 & \selectlanguage{english}%
{\scriptsize{}76.3}\selectlanguage{british}%
 & \selectlanguage{english}%
{\scriptsize{}65.2}\selectlanguage{british}%
 & \selectlanguage{english}%
{\scriptsize{}51.2}\selectlanguage{british}%
 & \selectlanguage{english}%
{\scriptsize{}14.1}\selectlanguage{british}%
 & \selectlanguage{english}%
{\scriptsize{}44.6}\selectlanguage{british}%
 & \selectlanguage{english}%
{\scriptsize{}58.1}\selectlanguage{british}%
 & \selectlanguage{english}%
{\scriptsize{}72.0}\selectlanguage{british}%
 & \selectlanguage{english}%
{\scriptsize{}54.3}\selectlanguage{british}%
 &  & \selectlanguage{english}%
{\scriptsize{}53.6}\selectlanguage{british}%
\tabularnewline
\texttt{\scriptsize{}RandomizedPrims} & \selectlanguage{english}%
{\scriptsize{}\cellp70.2}\selectlanguage{british}%
 & \selectlanguage{english}%
{\scriptsize{}68.2}\selectlanguage{british}%
 & \selectlanguage{english}%
{\scriptsize{}55.5}\selectlanguage{british}%
 & \selectlanguage{english}%
{\scriptsize{}39.5}\selectlanguage{british}%
 & \selectlanguage{english}%
{\scriptsize{}18.5}\selectlanguage{british}%
 & \selectlanguage{english}%
{\scriptsize{}\cellp72.3}\selectlanguage{british}%
 & \selectlanguage{english}%
{\scriptsize{}63.7}\selectlanguage{british}%
 & \selectlanguage{english}%
{\scriptsize{}\cellp76.8}\selectlanguage{british}%
 & \selectlanguage{english}%
{\scriptsize{}25.7}\selectlanguage{british}%
 & \selectlanguage{english}%
{\scriptsize{}62.4}\selectlanguage{british}%
 & \selectlanguage{english}%
{\scriptsize{}64.2}\selectlanguage{british}%
 & \selectlanguage{english}%
{\scriptsize{}\cellp68.7}\selectlanguage{british}%
 & \selectlanguage{english}%
{\scriptsize{}76.6}\selectlanguage{british}%
 & \selectlanguage{english}%
{\scriptsize{}68.5}\selectlanguage{british}%
 & \selectlanguage{english}%
{\scriptsize{}51.0}\selectlanguage{british}%
 & \selectlanguage{english}%
{\scriptsize{}22.4}\selectlanguage{british}%
 & \selectlanguage{english}%
{\scriptsize{}53.1}\selectlanguage{british}%
 & \selectlanguage{english}%
{\scriptsize{}62.9}\selectlanguage{british}%
 & \selectlanguage{english}%
{\scriptsize{}72.4}\selectlanguage{british}%
 & \selectlanguage{english}%
{\scriptsize{}59.7}\selectlanguage{british}%
 &  & \selectlanguage{english}%
{\scriptsize{}57.6}\selectlanguage{british}%
\tabularnewline
\texttt{\scriptsize{}Rantalankila} & \selectlanguage{english}%
{\scriptsize{}64.7}\selectlanguage{british}%
 & \selectlanguage{english}%
{\scriptsize{}66.1}\selectlanguage{british}%
 & \selectlanguage{english}%
{\scriptsize{}57.2}\selectlanguage{british}%
 & \selectlanguage{english}%
{\scriptsize{}37.8}\selectlanguage{british}%
 & \selectlanguage{english}%
{\scriptsize{}19.7}\selectlanguage{british}%
 & \selectlanguage{english}%
{\scriptsize{}\cellp}\textbf{\scriptsize{}74.2}\selectlanguage{british}%
 & \selectlanguage{english}%
{\scriptsize{}67.5}\selectlanguage{british}%
 & \selectlanguage{english}%
{\scriptsize{}\cellp}\textbf{\scriptsize{}78.2}\selectlanguage{british}%
 & \selectlanguage{english}%
{\scriptsize{}23.0}\selectlanguage{british}%
 & \selectlanguage{english}%
{\scriptsize{}63.6}\selectlanguage{british}%
 & \selectlanguage{english}%
{\scriptsize{}63.4}\selectlanguage{british}%
 & \selectlanguage{english}%
{\scriptsize{}\cellp}\textbf{\scriptsize{}70.3}\selectlanguage{british}%
 & \selectlanguage{english}%
{\scriptsize{}\cellp78.6}\selectlanguage{british}%
 & \selectlanguage{english}%
{\scriptsize{}69.8}\selectlanguage{british}%
 & \selectlanguage{english}%
{\scriptsize{}55.9}\selectlanguage{british}%
 & \selectlanguage{english}%
{\scriptsize{}21.4}\selectlanguage{british}%
 & \selectlanguage{english}%
{\scriptsize{}50.8}\selectlanguage{british}%
 & \selectlanguage{english}%
{\scriptsize{}\cellp64.3}\selectlanguage{british}%
 & \selectlanguage{english}%
{\scriptsize{}\cellp74.1}\selectlanguage{british}%
 & \selectlanguage{english}%
{\scriptsize{}58.3}\selectlanguage{british}%
 &  & \selectlanguage{english}%
{\scriptsize{}57.9}\selectlanguage{british}%
\tabularnewline
\texttt{\scriptsize{}Rigor} & \selectlanguage{english}%
{\scriptsize{}62.6}\selectlanguage{british}%
 & \selectlanguage{english}%
{\scriptsize{}\cellp70.5}\selectlanguage{british}%
 & \selectlanguage{english}%
{\scriptsize{}57.5}\selectlanguage{british}%
 & \selectlanguage{english}%
{\scriptsize{}40.1}\selectlanguage{british}%
 & \selectlanguage{english}%
{\scriptsize{}15.9}\selectlanguage{british}%
 & \selectlanguage{english}%
{\scriptsize{}\cellp72.9}\selectlanguage{british}%
 & \selectlanguage{english}%
{\scriptsize{}65.7}\selectlanguage{british}%
 & \selectlanguage{english}%
{\scriptsize{}\cellp77.9}\selectlanguage{british}%
 & \selectlanguage{english}%
{\scriptsize{}\cellp28.6}\selectlanguage{british}%
 & \selectlanguage{english}%
{\scriptsize{}65.1}\selectlanguage{british}%
 & \selectlanguage{english}%
{\scriptsize{}63.7}\selectlanguage{british}%
 & \selectlanguage{english}%
{\scriptsize{}\cellp68.6}\selectlanguage{british}%
 & \selectlanguage{english}%
{\scriptsize{}\cellp77.9}\selectlanguage{british}%
 & \selectlanguage{english}%
{\scriptsize{}68.9}\selectlanguage{british}%
 & \selectlanguage{english}%
{\scriptsize{}54.8}\selectlanguage{british}%
 & \selectlanguage{english}%
{\scriptsize{}23.3}\selectlanguage{british}%
 & \selectlanguage{english}%
{\scriptsize{}\cellp56.3}\selectlanguage{british}%
 & \selectlanguage{english}%
{\scriptsize{}63.8}\selectlanguage{british}%
 & \selectlanguage{english}%
{\scriptsize{}73.7}\selectlanguage{british}%
 & \selectlanguage{english}%
{\scriptsize{}60.3}\selectlanguage{british}%
 &  & \selectlanguage{english}%
{\scriptsize{}\cellp58.4}\selectlanguage{british}%
\tabularnewline
\texttt{\scriptsize{}SelectiveSearch} & \selectlanguage{english}%
{\scriptsize{}\cellp}\textbf{\scriptsize{}70.3}\selectlanguage{british}%
 & \selectlanguage{english}%
{\scriptsize{}66.9}\selectlanguage{british}%
 & \selectlanguage{english}%
{\scriptsize{}\cellp}\textbf{\scriptsize{}61.5}\selectlanguage{british}%
 & \selectlanguage{english}%
{\scriptsize{}42.2}\selectlanguage{british}%
 & \selectlanguage{english}%
{\scriptsize{}21.7}\selectlanguage{british}%
 & \selectlanguage{english}%
{\scriptsize{}68.3}\selectlanguage{british}%
 & \selectlanguage{english}%
{\scriptsize{}68.7}\selectlanguage{british}%
 & \selectlanguage{english}%
{\scriptsize{}\cellp76.3}\selectlanguage{british}%
 & \selectlanguage{english}%
{\scriptsize{}27.5}\selectlanguage{british}%
 & \selectlanguage{english}%
{\scriptsize{}65.9}\selectlanguage{british}%
 & \selectlanguage{english}%
{\scriptsize{}\cellp}\textbf{\scriptsize{}67.0}\selectlanguage{british}%
 & \selectlanguage{english}%
{\scriptsize{}\cellp69.8}\selectlanguage{british}%
 & \selectlanguage{english}%
{\scriptsize{}75.5}\selectlanguage{british}%
 & \selectlanguage{english}%
{\scriptsize{}68.9}\selectlanguage{british}%
 & \selectlanguage{english}%
{\scriptsize{}57.9}\selectlanguage{british}%
 & \selectlanguage{english}%
{\scriptsize{}24.6}\selectlanguage{british}%
 & \selectlanguage{english}%
{\scriptsize{}53.6}\selectlanguage{british}%
 & \selectlanguage{english}%
{\scriptsize{}63.7}\selectlanguage{british}%
 & \selectlanguage{english}%
{\scriptsize{}\cellp}\textbf{\scriptsize{}76.0}\selectlanguage{british}%
 & \selectlanguage{english}%
{\scriptsize{}\cellp}\textbf{\scriptsize{}62.4}\selectlanguage{british}%
 &  & \selectlanguage{english}%
{\scriptsize{}\cellp59.5}\selectlanguage{british}%
\tabularnewline
\hline 
\texttt{\scriptsize{}Gaussian} & \selectlanguage{english}%
{\scriptsize{}53.9}\selectlanguage{british}%
 & \selectlanguage{english}%
{\scriptsize{}66.1}\selectlanguage{british}%
 & \selectlanguage{english}%
{\scriptsize{}46.6}\selectlanguage{british}%
 & \selectlanguage{english}%
{\scriptsize{}24.6}\selectlanguage{british}%
 & \selectlanguage{english}%
{\scriptsize{}10.0}\selectlanguage{british}%
 & \selectlanguage{english}%
{\scriptsize{}66.6}\selectlanguage{british}%
 & \selectlanguage{english}%
{\scriptsize{}52.2}\selectlanguage{british}%
 & \selectlanguage{english}%
{\scriptsize{}77.1}\selectlanguage{british}%
 & \selectlanguage{english}%
{\scriptsize{}20.6}\selectlanguage{british}%
 & \selectlanguage{english}%
{\scriptsize{}48.7}\selectlanguage{british}%
 & \selectlanguage{english}%
{\scriptsize{}64.1}\selectlanguage{british}%
 & \selectlanguage{english}%
{\scriptsize{}65.5}\selectlanguage{british}%
 & \selectlanguage{english}%
{\scriptsize{}75.6}\selectlanguage{british}%
 & \selectlanguage{english}%
{\scriptsize{}64.2}\selectlanguage{british}%
 & \selectlanguage{english}%
{\scriptsize{}47.0}\selectlanguage{british}%
 & \selectlanguage{english}%
{\scriptsize{}14.2}\selectlanguage{british}%
 & \selectlanguage{english}%
{\scriptsize{}38.1}\selectlanguage{british}%
 & \selectlanguage{english}%
{\scriptsize{}58.2}\selectlanguage{british}%
 & \selectlanguage{english}%
{\scriptsize{}70.5}\selectlanguage{british}%
 & \selectlanguage{english}%
{\scriptsize{}53.0}\selectlanguage{british}%
 &  & \selectlanguage{english}%
{\scriptsize{}50.8}\selectlanguage{british}%
\tabularnewline
\texttt{\scriptsize{}SlidingWindow} & \selectlanguage{english}%
{\scriptsize{}42.0}\selectlanguage{british}%
 & \selectlanguage{english}%
{\scriptsize{}57.7}\selectlanguage{british}%
 & \selectlanguage{english}%
{\scriptsize{}40.1}\selectlanguage{british}%
 & \selectlanguage{english}%
{\scriptsize{}23.7}\selectlanguage{british}%
 & \selectlanguage{english}%
{\scriptsize{}9.3}\selectlanguage{british}%
 & \selectlanguage{english}%
{\scriptsize{}60.8}\selectlanguage{british}%
 & \selectlanguage{english}%
{\scriptsize{}47.8}\selectlanguage{british}%
 & \selectlanguage{english}%
{\scriptsize{}72.8}\selectlanguage{british}%
 & \selectlanguage{english}%
{\scriptsize{}12.5}\selectlanguage{british}%
 & \selectlanguage{english}%
{\scriptsize{}42.1}\selectlanguage{british}%
 & \selectlanguage{english}%
{\scriptsize{}44.7}\selectlanguage{british}%
 & \selectlanguage{english}%
{\scriptsize{}63.7}\selectlanguage{british}%
 & \selectlanguage{english}%
{\scriptsize{}72.8}\selectlanguage{british}%
 & \selectlanguage{english}%
{\scriptsize{}62.5}\selectlanguage{british}%
 & \selectlanguage{english}%
{\scriptsize{}44.5}\selectlanguage{british}%
 & \selectlanguage{english}%
{\scriptsize{}8.5}\selectlanguage{british}%
 & \selectlanguage{english}%
{\scriptsize{}34.3}\selectlanguage{british}%
 & \selectlanguage{english}%
{\scriptsize{}47.7}\selectlanguage{british}%
 & \selectlanguage{english}%
{\scriptsize{}62.3}\selectlanguage{british}%
 & \selectlanguage{english}%
{\scriptsize{}46.6}\selectlanguage{british}%
 &  & \selectlanguage{english}%
{\scriptsize{}44.8}\selectlanguage{british}%
\tabularnewline
\texttt{\scriptsize{}Superpixels} & \selectlanguage{english}%
{\scriptsize{}29.7}\selectlanguage{british}%
 & \selectlanguage{english}%
{\scriptsize{}5.5}\selectlanguage{british}%
 & \selectlanguage{english}%
{\scriptsize{}19.8}\selectlanguage{british}%
 & \selectlanguage{english}%
{\scriptsize{}10.4}\selectlanguage{british}%
 & \selectlanguage{english}%
{\scriptsize{}9.0}\selectlanguage{british}%
 & \selectlanguage{english}%
{\scriptsize{}7.4}\selectlanguage{british}%
 & \selectlanguage{english}%
{\scriptsize{}24.4}\selectlanguage{british}%
 & \selectlanguage{english}%
{\scriptsize{}42.0}\selectlanguage{british}%
 & \selectlanguage{english}%
{\scriptsize{}15.1}\selectlanguage{british}%
 & \selectlanguage{english}%
{\scriptsize{}39.9}\selectlanguage{british}%
 & \selectlanguage{english}%
{\scriptsize{}6.6}\selectlanguage{british}%
 & \selectlanguage{english}%
{\scriptsize{}30.3}\selectlanguage{british}%
 & \selectlanguage{english}%
{\scriptsize{}10.7}\selectlanguage{british}%
 & \selectlanguage{english}%
{\scriptsize{}13.7}\selectlanguage{british}%
 & \selectlanguage{english}%
{\scriptsize{}12.8}\selectlanguage{british}%
 & \selectlanguage{english}%
{\scriptsize{}8.9}\selectlanguage{british}%
 & \selectlanguage{english}%
{\scriptsize{}40.7}\selectlanguage{british}%
 & \selectlanguage{english}%
{\scriptsize{}18.1}\selectlanguage{british}%
 & \selectlanguage{english}%
{\scriptsize{}4.9}\selectlanguage{british}%
 & \selectlanguage{english}%
{\scriptsize{}55.6}\selectlanguage{british}%
 &  & \selectlanguage{english}%
{\scriptsize{}20.3}\selectlanguage{british}%
\tabularnewline
\texttt{\scriptsize{}Uniform} & \selectlanguage{english}%
{\scriptsize{}51.0}\selectlanguage{british}%
 & \selectlanguage{english}%
{\scriptsize{}58.0}\selectlanguage{british}%
 & \selectlanguage{english}%
{\scriptsize{}38.6}\selectlanguage{british}%
 & \selectlanguage{english}%
{\scriptsize{}24.6}\selectlanguage{british}%
 & \selectlanguage{english}%
{\scriptsize{}11.7}\selectlanguage{british}%
 & \selectlanguage{english}%
{\scriptsize{}64.3}\selectlanguage{british}%
 & \selectlanguage{english}%
{\scriptsize{}50.9}\selectlanguage{british}%
 & \selectlanguage{english}%
{\scriptsize{}72.3}\selectlanguage{british}%
 & \selectlanguage{english}%
{\scriptsize{}14.8}\selectlanguage{british}%
 & \selectlanguage{english}%
{\scriptsize{}43.4}\selectlanguage{british}%
 & \selectlanguage{english}%
{\scriptsize{}62.6}\selectlanguage{british}%
 & \selectlanguage{english}%
{\scriptsize{}63.4}\selectlanguage{british}%
 & \selectlanguage{english}%
{\scriptsize{}73.9}\selectlanguage{british}%
 & \selectlanguage{english}%
{\scriptsize{}59.3}\selectlanguage{british}%
 & \selectlanguage{english}%
{\scriptsize{}43.4}\selectlanguage{british}%
 & \selectlanguage{english}%
{\scriptsize{}10.8}\selectlanguage{british}%
 & \selectlanguage{english}%
{\scriptsize{}27.5}\selectlanguage{british}%
 & \selectlanguage{english}%
{\scriptsize{}60.4}\selectlanguage{british}%
 & \selectlanguage{english}%
{\scriptsize{}69.0}\selectlanguage{british}%
 & \selectlanguage{english}%
{\scriptsize{}38.3}\selectlanguage{british}%
 &  & \selectlanguage{english}%
{\scriptsize{}46.9}\selectlanguage{british}%
\tabularnewline
\hline 
{\scriptsize{}best per class} & \selectlanguage{english}%
{\scriptsize{}70.3}\selectlanguage{british}%
 & \selectlanguage{english}%
{\scriptsize{}70.8}\selectlanguage{british}%
 & \selectlanguage{english}%
{\scriptsize{}61.5}\selectlanguage{british}%
 & \selectlanguage{english}%
{\scriptsize{}46.1}\selectlanguage{british}%
 & \selectlanguage{english}%
{\scriptsize{}28.5}\selectlanguage{british}%
 & \selectlanguage{english}%
{\scriptsize{}74.2}\selectlanguage{british}%
 & \selectlanguage{english}%
{\scriptsize{}72.3}\selectlanguage{british}%
 & \selectlanguage{english}%
{\scriptsize{}78.2}\selectlanguage{british}%
 & \selectlanguage{english}%
{\scriptsize{}30.2}\selectlanguage{british}%
 & \selectlanguage{english}%
{\scriptsize{}68.1}\selectlanguage{british}%
 & \selectlanguage{english}%
{\scriptsize{}67.0}\selectlanguage{british}%
 & \selectlanguage{english}%
{\scriptsize{}70.3}\selectlanguage{british}%
 & \selectlanguage{english}%
{\scriptsize{}79.2}\selectlanguage{british}%
 & \selectlanguage{english}%
{\scriptsize{}73.6}\selectlanguage{british}%
 & \selectlanguage{english}%
{\scriptsize{}62.4}\selectlanguage{british}%
 & \selectlanguage{english}%
{\scriptsize{}28.2}\selectlanguage{british}%
 & \selectlanguage{english}%
{\scriptsize{}57.6}\selectlanguage{british}%
 & \selectlanguage{english}%
{\scriptsize{}66.0}\selectlanguage{british}%
 & \selectlanguage{english}%
{\scriptsize{}76.0}\selectlanguage{british}%
 & \selectlanguage{english}%
{\scriptsize{}62.4}\selectlanguage{british}%
 &  & \selectlanguage{english}%
{\scriptsize{}62.1}\selectlanguage{british}%
\tabularnewline
\end{tabular}\hspace*{\fill}

\protect\caption{\label{tab:pascal-FRCN-per-class}Fast R-CNN (model M) detection results
(AP) on PASCAL VOC 2007. Bold numbers indicate the best proposal method
per class, green numbers are within 2 AP of the best result. The ``best
per class'' row shows the best performance when choosing the optimal
proposals per class, improving from 60.4 mAP (\texttt{EdgeBoxes})
to 62.1 mAP.}
\end{table*}

We use pre-trained LM-LLDA~\cite{Girshick2013ICCV} models to generate
dense detections using the standard sliding window setup and subsequently
apply different proposals to filter these detections at test time.
This does not speed-up detection, but enables evaluating the effect
of proposals on detection quality. A priori we may expect that detection
results will deteriorate due to lost recall, but conversely, they
may improve if the proposals filter out windows that would otherwise
be false positives.

\paragraph*{Implementation}

We take the raw detections of LM-LLDA before non-maximum suppression
(nms) and filter them with the detection proposals of each method.
We keep all detections that overlap more than 0.8 IoU with a candidate
proposal and subsequently apply nms to the surviving detections. As
a final step we do bounding box regression, as is common for DPM models
\cite{Felzenszwalb2010Pami}. Note that this procedure returns predictions
near to, but distinct from, each proposal.

\paragraph*{Results}

Table~\ref{tab:pascal-mAP}, LM-LLDA columns, show that using $1\,000$
proposals decreases detection quality compared with the original sliding
window setup%
\footnote{Not to be confused with the \texttt{SlidingWindow} proposals baseline.%
} by about 1-2 mAP for the best performing methods, see top row (\texttt{Dense})
versus the rows below. The five top performing methods all have mAP
between 32.0 and 33.0 and are marked in green: \texttt{MCG}, \texttt{Selective\-Search},
\texttt{Edge\-Boxes}, \texttt{Geo\-desic}, and \texttt{Rigor}. Note
that the difference between these methods and the \texttt{Gaussian}
baseline is fairly small ($33.0$ versus $28.0$ mAP). 

When we compare these results with figure~\ref{fig:recall-vs-iuo-area-vs-number-of-proposals-pascal}
at $1\,000$ proposals, we see that methods are ranked similarly.
Methods with high average recall (AR) also have high mAP, and methods
with lower AR also have lower mAP. We analyse the correlation between
AR and mAP more closely in \S\ref{sub:mAP-vs-recall}.

From table~\ref{tab:pascal-llda-dpm-per-class} we see that the per-class
performance can be grouped into three cases: classes on which the
best proposals (1) clearly hurt performance (bicycle, boat, bottle,
car, chair, horse, mbike, person), (2) improve performance (cat, table,
dog), (3) do not show significant change (all remaining classes).
In the case of (1) we observe both reduced recall and reduced precision
in the detection curves, probably because bad localisation decreases
the scores of strong detections.

\subsection{\label{sub:R-CNN} R-CNN detection performance}

The highly successful and widely used R-CNN detector \cite{Girshick2014Cvpr}
couples detection proposals with a convolutional neural network classification
stage. It was designed from the ground up to rely on proposals, making
it a perfect candidate for our case study. We report results for both
the original R-CNN detector and also the improved Fast R-CNN \cite{Girshick2015arXiv}.
We focus primarily on Fast R-CNN due to its efficiency and higher
detection accuracy.

\paragraph*{Implementation}

For each proposal method we re-train and test Fast R-CNN (using the
medium model M for efficiency). Unlike Fast R-CNN, the original R-CNN
is fairly slow to train; we therefore experiment with the R-CNN model
that is published with the code and which has been trained on $2\,000$
\texttt{SelectiveSearch} proposals.

\paragraph*{Results}

Although the absolute mAP numbers are considerably higher for Fast
R-CNN (nearly double mAP), the results (Fast R-CNN and R-CNN) in table~\ref{tab:pascal-mAP}
show a similar trend than the LM-LLDA results. As expected, \texttt{Selective\-Search},
with which Fast R-CNN was developed, performs well, but multiple other
proposal methods get similar results. The five top performing methods
are similar to the top methods for LM-LLDA: \texttt{Rantalankila}
edges out \texttt{EdgeBoxes} for R-CNN and \texttt{Geodesic} for Fast
R-CNN. \texttt{EdgeBoxes} and \texttt{MCG} provide the best results.
The gap between \texttt{Gaussian} and the top result is more pronounced
($60.4$ versus $50.8$ mAP), but this baseline still performs surprisingly
well considering it disregards the image content. We show per-class
Fast R-CNN results in table \ref{tab:pascal-FRCN-per-class}.

\paragraph*{Retraining}

To provide a fair comparison amongst proposal methods, the ``Fast
R-CNN'' column in table~\ref{tab:pascal-mAP} reports results after
re-training for each method. The rightmost column of table~\ref{tab:pascal-mAP}
shows the change in mAP when comparing Fast R-CNN (with bounding box
regression) trained with $1\,000$ \texttt{SelectiveSearch} proposals
and applied at test time with a given proposal method, versus Fast
R-CNN trained for the test time proposal method. 

Most methods improve from re-training, although the performance of
a few degrades. While in most cases the change in mAP is within 1-2
points, re-training provided substantial benefits for \texttt{Bing},
\texttt{Edge\-Boxes}, \texttt{Objectness}, and \texttt{SlidingWindow}.
These methods all have poor localisation at high IoU (see figure \ref{fig:recall-versus-iou-threshold-pascal});
re-training likely allows Fast R-CNN to compensate for their inferior
localisation.

\paragraph*{Summary}

We emphasise that the various proposal methods exhibit similar ordering
with all tested detectors (LM-LLDA, R-CNN, and Fast R-CNN). Our experiments
did not reveal any proposal methods as being particularly well-adapted
for certain detectors; rather, for object detection some proposals
methods are strictly better than others.

\begin{figure*}[t]
\begin{centering}
\hspace*{\fill}\subfloat[\label{fig:correlation-recall-IoU-mAP-LM-LLDA}LM-LLDA with bounding
box regression]{\begin{centering}
\includegraphics[width=0.45\columnwidth]{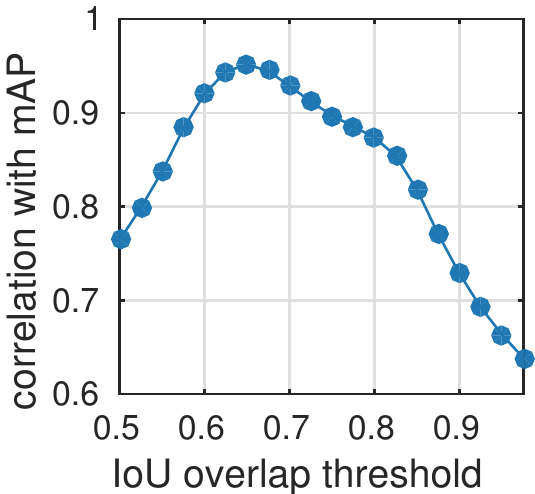}\quad{}\includegraphics[width=0.45\columnwidth]{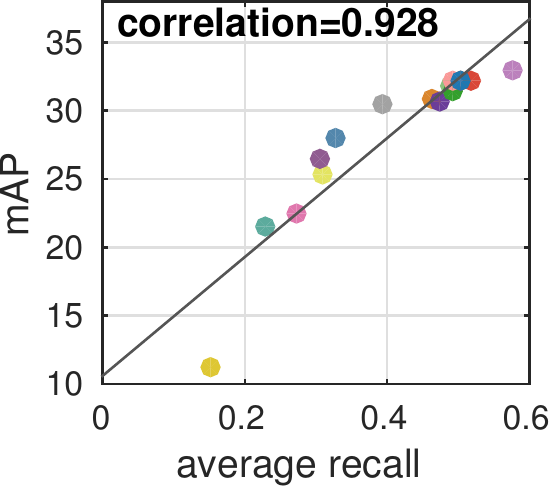}
\par\end{centering}

}\hspace*{\fill}\subfloat[\label{fig:correlation-recall-IoU-mAP-LM-LLDA-2-1-1}R-CNN without
bounding box regression]{\begin{centering}
\includegraphics[width=0.45\columnwidth]{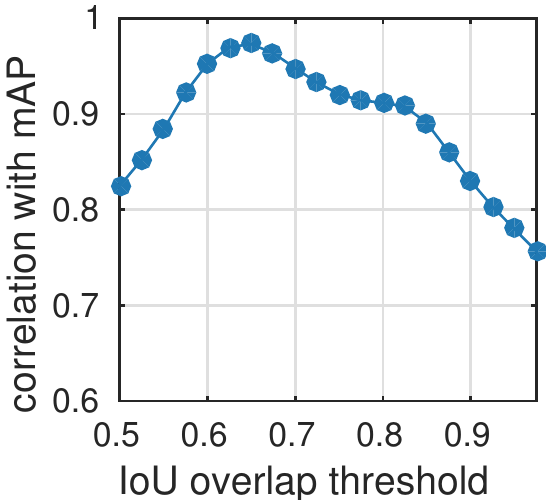}\quad{}\includegraphics[width=0.45\columnwidth]{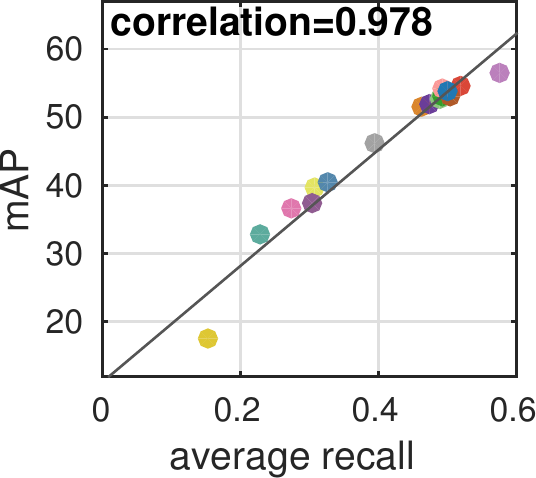}
\par\end{centering}

}\hspace*{\fill}
\par\end{centering}

\begin{centering}
\hspace*{\fill}\subfloat[\label{fig:correlation-recall-IoU-mAP-FRCN-noregr}Fast R-CNN without
bounding box regression]{\begin{centering}
\includegraphics[width=0.45\columnwidth]{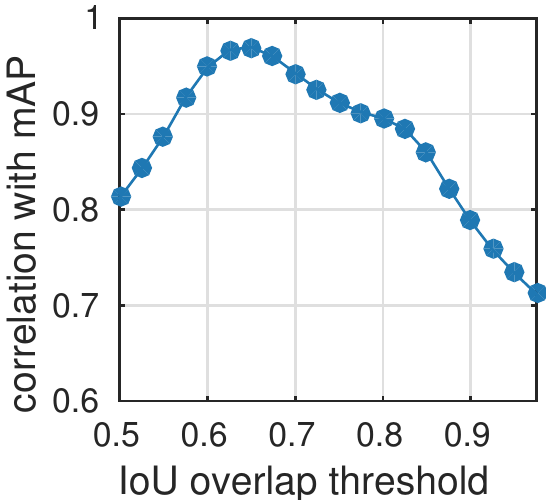}\quad{}\includegraphics[width=0.45\columnwidth]{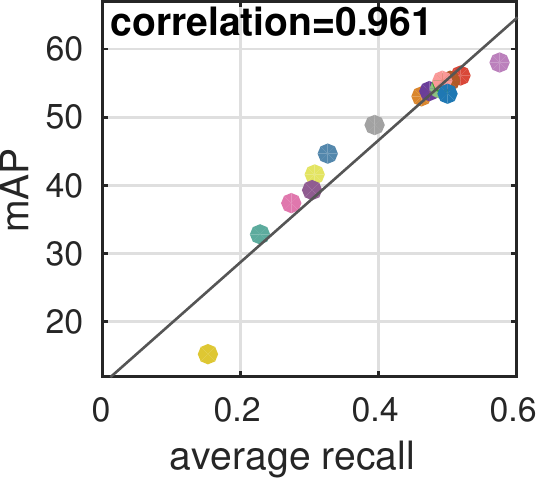}
\par\end{centering}

}\hspace*{\fill}\subfloat[\label{fig:correlation-recall-IoU-mAP-FRCN}Fast R-CNN with bounding
box regression]{\begin{centering}
\includegraphics[width=0.45\columnwidth]{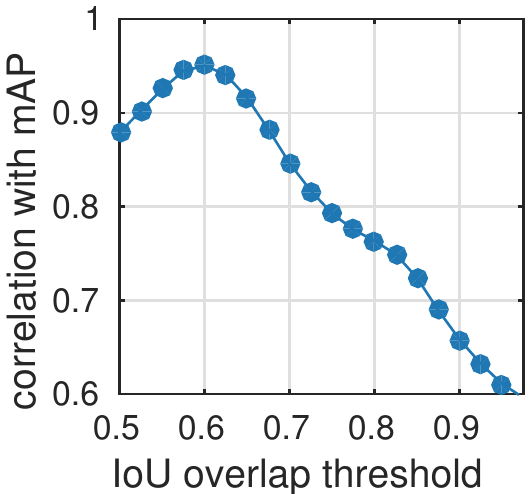}\quad{}\includegraphics[bb=0bp 0bp 155bp 142bp,width=0.45\columnwidth]{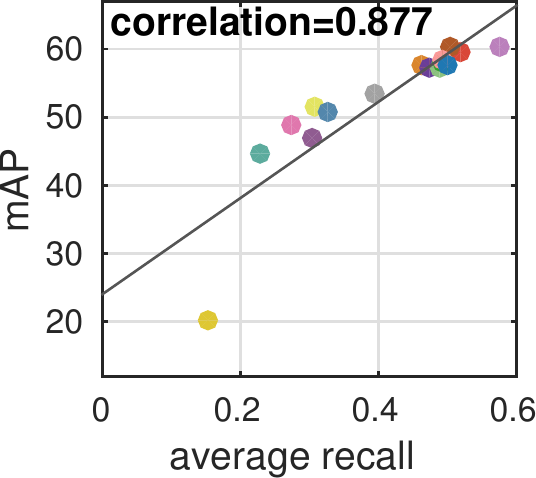}
\par\end{centering}

}\hspace*{\fill}
\par\end{centering}

\centering{}\protect\caption{\label{fig:correlation-recall-mAP}Correlation between detector performance
on PASCAL 07 and different proposal metrics. Left columns: correlation
between mAP and recall at different IoU thresholds. Right columns:
correlation between mAP and AR.}
\end{figure*}

\subsection{\label{sub:mAP-vs-recall}Predicting detection performance}

We aim to determine which recall metrics from section \ref{sec:Proposal-recall}
(figures \ref{fig:recall-versus-iou-threshold-pascal} and \ref{fig:recall-versus-num-windows-pascal})
serve as the best predictor for detector performance. In figure~\ref{fig:correlation-recall-mAP}
we show the Pearson correlation coefficient between detector performance
and two recall metrics: recall at different overlap thresholds (left
columns) and the \emph{average recall} (AR) between IoU of 0.5 to
1.0 (right columns)%
\footnote{We compute the average between 0.5 and 1 IoU (and not between 0 and
1 as in \S\ref{sec:Repeatibility}), because we are interested in
recall above the PASCAL evaluation criterion of 0.5 IoU. Proposals
with worse overlap than 0.5 are not only harder to classify correctly,
but require a potentially large subsequent location refinement to
become a successful detection.%
}. As before, we use $1\,000$ proposals per method.

We begin by examining correlation between detection performance and
recall at various IoU thresholds (figure~\ref{fig:correlation-recall-mAP},
left columns). All detectors show a strong correlation ($>0.9$) at
an IoU range of roughly 0.6 to 0.8, with the exception of Fast R-CNN
with bounding box prediction, which correlates better for lower overlap.
Note that recall at IoU of 0.5 is actually only weakly correlated
with detection performance, and methods that optimise for IoU of 0.5,
such as \texttt{Bing}, are not well suited for use with object detectors
(see table \ref{tab:pascal-mAP}). Thus, although recall at IoU of
0.5 has been traditionally used to evaluate object proposals, our
analysis shows that it is \emph{not} a good metric for predicting
detection performance.

The correlation between detection performance and AR is quite strong,
see figure~\ref{fig:correlation-recall-mAP}, right columns. Computing
the AR over a partial IoU range (e.g. 0.6 to 0.8) can further increase
the correlation; however, since the effect is generally minor, we
opted to use AR over the entire range from 0.5 to 1.0 for simplicity.
While the strong correlation does not imply that the AR can perfectly
predict detection performance, as figure~\ref{fig:correlation-recall-mAP}
shows, the relationship is surprisingly linear. AR over the full range
of 0 to 1 IoU (which is similar to ABO, see appendix \ref{sec:additional-metrics})
has weaker correlation with mAP, since proposals with low overlap
are not sufficient for a successful detection under the PASCAL criterion
and are also harder to classify.

For detectors with bounding box regression, the AR computation can
be restricted to a tighter IoU range. In figure \ref{fig:detector-scoremaps},
we can observe that detection score of Fast R-CNN saturates earlier.
Thus there is little benefit in proposals that are perfectly localised
as the bounding box refinement improves the localisation of those
proposals. If we restrict the AR to IoU from 0.5 to 0.7, we obtain
a higher correlation of 0.949 for Fast R-CNN with bounding box regression
(compared to 0.877 in figure \ref{fig:correlation-recall-IoU-mAP-FRCN}).

\begin{figure}
\includegraphics[width=1\columnwidth]{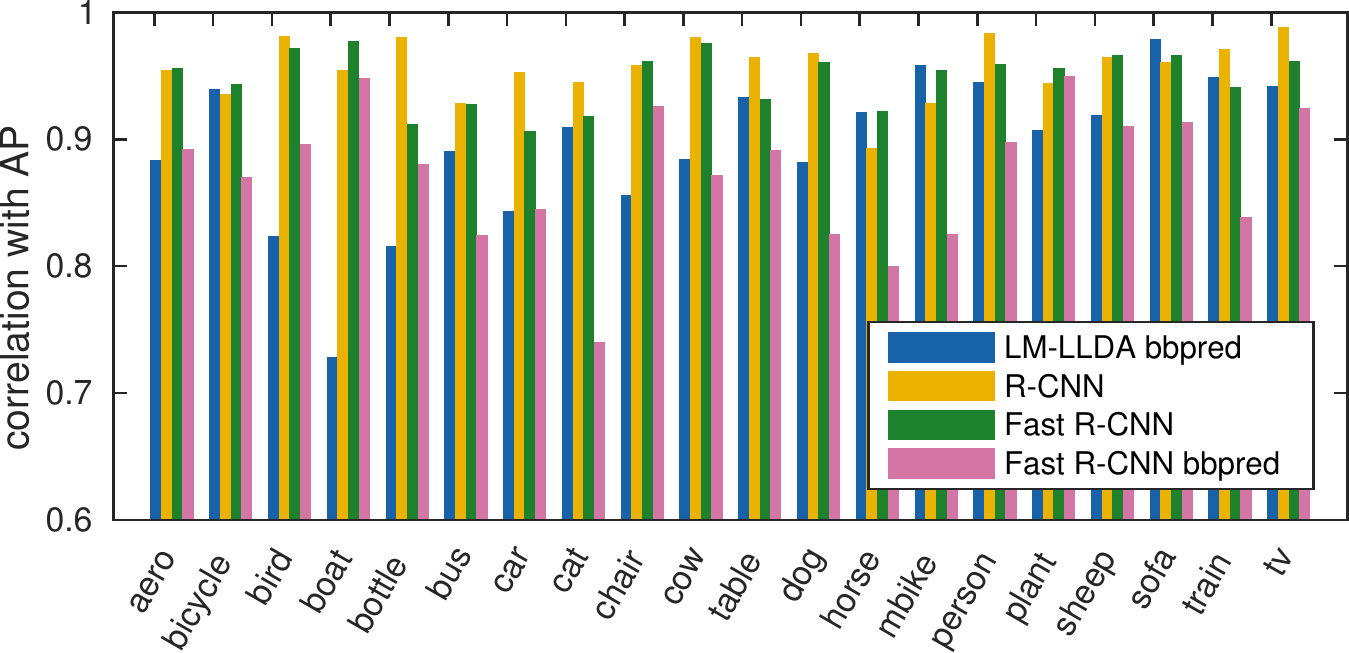}

\protect\caption{\label{fig:correlation-AR-AP-per-class}Correlation between AR and
AP for each PASCAL VOC class and detector across all proposal methods.
}
\end{figure}

For a more detailed analysis of the correlation between mAP and AR
we show the correlation for each class for different detectors in
figure \ref{fig:correlation-AR-AP-per-class}. The per-class correlation
is highest for R-CNN and Fast R-CNN without regression.

We conclude that AR allows us to identify good proposal methods for
object detection. The AR metric is simple, easy to justify, and is
strongly correlated with detection performance. Note that our analysis
only covers the case in which all methods produce the same number
of proposals. As Girshick \cite{Girshick2015arXiv} points out, as
the number of proposals increases, AR will necessarily increase but
resulting detector performance saturates and may even degrade. For
a fixed number of proposals, however, AR is a good predictor of detection
performance. We suggest that future proposal methods should aim to
optimise this metric.

\begin{figure*}[t]
\begin{centering}
\subfloat[\label{fig:Recall-versus-IoU-edgeboxes-variants}Recall versus IoU]{\begin{centering}
\includegraphics[bb=10bp 0bp 185bp 200bp,width=0.45\columnwidth]{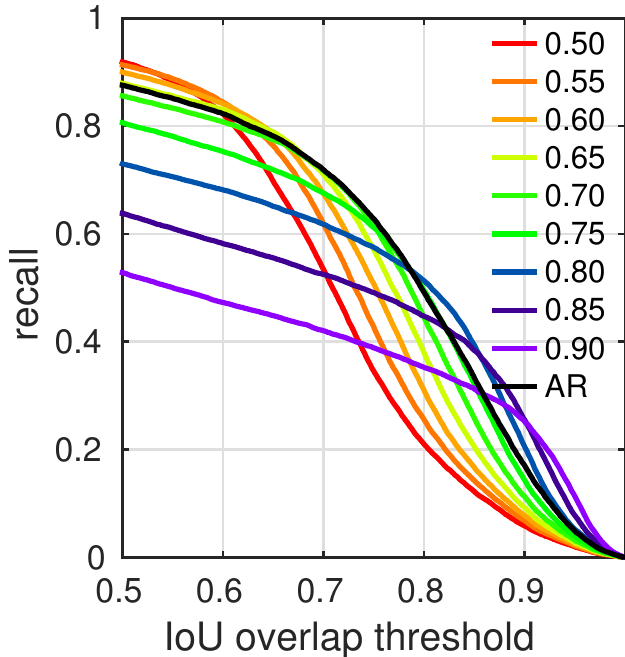}
\par\end{centering}

}\hfill{}\subfloat[\label{fig:mAP-vs-area-edgebox-versions-RCNN}R-CNN]{\begin{centering}
\includegraphics[bb=10bp 0bp 188bp 198bp,width=0.45\columnwidth]{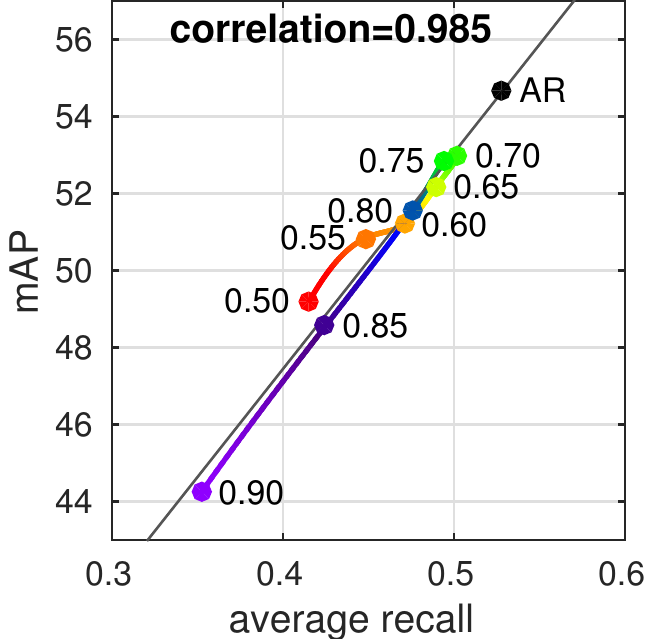}
\par\end{centering}

}\hfill{}\subfloat[\label{fig:mAP-vs-area-edgebox-versions-FRCN-noregr}Fast R-CNN without
regression]{\begin{centering}
\includegraphics[bb=10bp 0bp 188bp 198bp,width=0.45\columnwidth]{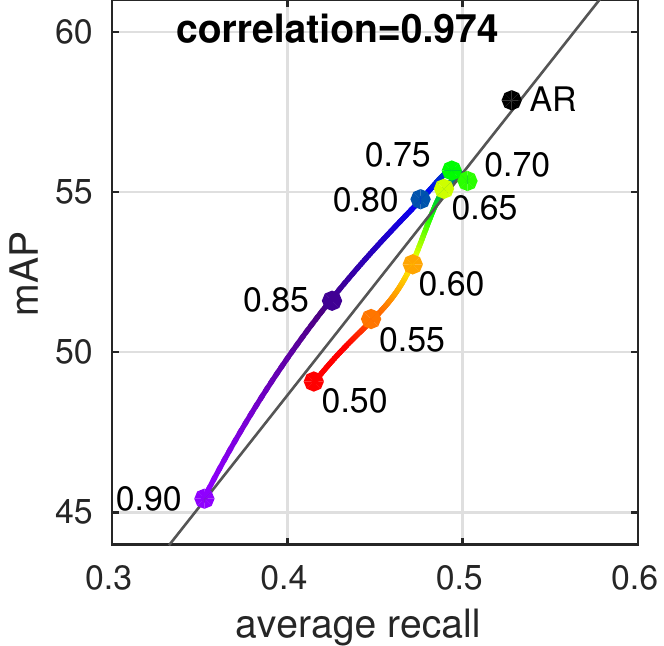}
\par\end{centering}

}\hfill{}\subfloat[\label{fig:mAP-vs-area-edgebox-versions-FRCN}Fast R-CNN with regression]{\begin{centering}
\includegraphics[bb=10bp 0bp 188bp 198bp,width=0.45\columnwidth]{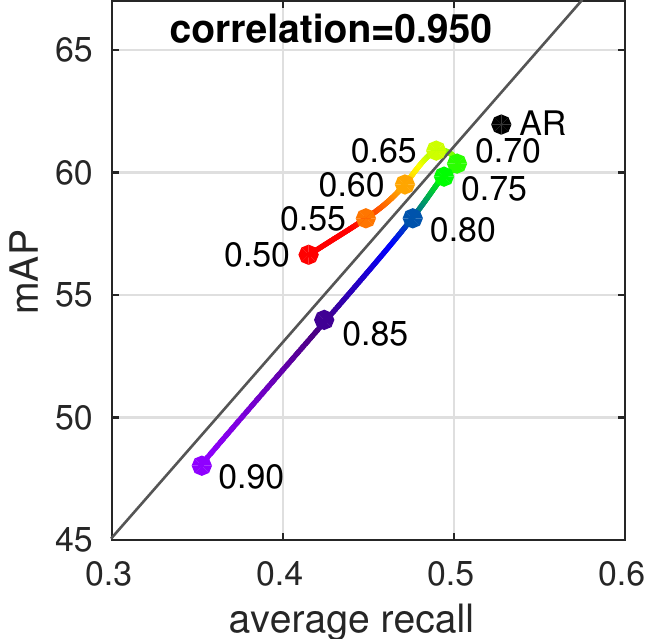}
\par\end{centering}

}
\par\end{centering}

\protect\caption{\label{fig:edge-box-tuning}Finetuning \texttt{EdgeBoxes} to optimise
AR results in top detector performance. These results further support
the conclusion that AR is a good predictor for mAP and suggest that
it can be used for fine-tuning proposal methods.}
\end{figure*}
\begin{table*}
\setlength\tabcolsep{2pt}

\hspace*{\fill}%
\begin{tabular}{lcccccccccccccccccccccc}
 & {\scriptsize{}aero} & {\scriptsize{}bicycle} & {\scriptsize{}bird} & {\scriptsize{}boat} & {\scriptsize{}bottle} & {\scriptsize{}bus} & {\scriptsize{}car} & {\scriptsize{}cat} & {\scriptsize{}chair} & {\scriptsize{}cow} & {\scriptsize{}table} & {\scriptsize{}dog} & {\scriptsize{}horse} & {\scriptsize{}mbike} & {\scriptsize{}person} & {\scriptsize{}plant} & {\scriptsize{}sheep} & {\scriptsize{}sofa} & {\scriptsize{}train} & {\scriptsize{}tv} & {\scriptsize{}\quad{}} & {\scriptsize{}mean}\tabularnewline
\hline 
\texttt{\scriptsize{}Edge\-Boxes\-AR$\quad$} & \selectlanguage{english}%
{\scriptsize{}69.6}\selectlanguage{british}%
 & \selectlanguage{english}%
{\scriptsize{}78.3}\selectlanguage{british}%
 & \selectlanguage{english}%
{\scriptsize{}66.2}\selectlanguage{british}%
 & \selectlanguage{english}%
{\scriptsize{}58.6}\selectlanguage{british}%
 & \selectlanguage{english}%
{\scriptsize{}42.5}\selectlanguage{british}%
 & \selectlanguage{english}%
{\scriptsize{}82.1}\selectlanguage{british}%
 & \selectlanguage{english}%
{\scriptsize{}78.1}\selectlanguage{british}%
 & \selectlanguage{english}%
{\scriptsize{}83.0}\selectlanguage{british}%
 & \selectlanguage{english}%
{\scriptsize{}42.7}\selectlanguage{british}%
 & \selectlanguage{english}%
{\scriptsize{}74.6}\selectlanguage{british}%
 & \selectlanguage{english}%
{\scriptsize{}66.4}\selectlanguage{british}%
 & \selectlanguage{english}%
{\scriptsize{}81.1}\selectlanguage{british}%
 & \selectlanguage{english}%
{\scriptsize{}82.0}\selectlanguage{british}%
 & \selectlanguage{english}%
{\scriptsize{}74.5}\selectlanguage{british}%
 & \selectlanguage{english}%
{\scriptsize{}68.3}\selectlanguage{british}%
 & \selectlanguage{english}%
{\scriptsize{}35.1}\selectlanguage{british}%
 & \selectlanguage{english}%
{\scriptsize{}66.1}\selectlanguage{british}%
 & \selectlanguage{english}%
{\scriptsize{}68.7}\selectlanguage{british}%
 & \selectlanguage{english}%
{\scriptsize{}75.2}\selectlanguage{british}%
 & \selectlanguage{english}%
{\scriptsize{}62.6}\selectlanguage{british}%
 &  & \selectlanguage{english}%
{\scriptsize{}67.8}\selectlanguage{british}%
\tabularnewline
\hline 
\selectlanguage{english}%
{\scriptsize{}+ gt oracle}\selectlanguage{british}%
 & \selectlanguage{english}%
{\scriptsize{}75.5}\selectlanguage{british}%
 & \selectlanguage{english}%
{\scriptsize{}79.4}\selectlanguage{british}%
 & \selectlanguage{english}%
{\scriptsize{}70.6}\selectlanguage{british}%
 & \selectlanguage{english}%
{\scriptsize{}63.1}\selectlanguage{british}%
 & \selectlanguage{english}%
{\scriptsize{}55.0}\selectlanguage{british}%
 & \selectlanguage{english}%
{\scriptsize{}82.6}\selectlanguage{british}%
 & \selectlanguage{english}%
{\scriptsize{}84.3}\selectlanguage{british}%
 & \selectlanguage{english}%
{\scriptsize{}83.9}\selectlanguage{british}%
 & \selectlanguage{english}%
{\scriptsize{}46.6}\selectlanguage{british}%
 & \selectlanguage{english}%
{\scriptsize{}75.1}\selectlanguage{british}%
 & \selectlanguage{english}%
{\scriptsize{}68.1}\selectlanguage{british}%
 & \selectlanguage{english}%
{\scriptsize{}82.5}\selectlanguage{british}%
 & \selectlanguage{english}%
{\scriptsize{}83.3}\selectlanguage{british}%
 & \selectlanguage{english}%
{\scriptsize{}75.9}\selectlanguage{british}%
 & \selectlanguage{english}%
{\scriptsize{}76.8}\selectlanguage{british}%
 & \selectlanguage{english}%
{\scriptsize{}41.2}\selectlanguage{british}%
 & \selectlanguage{english}%
{\scriptsize{}67.6}\selectlanguage{british}%
 & \selectlanguage{english}%
{\scriptsize{}70.1}\selectlanguage{british}%
 & \selectlanguage{english}%
{\scriptsize{}77.3}\selectlanguage{british}%
 & \selectlanguage{english}%
{\scriptsize{}65.7}\selectlanguage{british}%
 &  & \selectlanguage{english}%
{\scriptsize{}71.2}\selectlanguage{british}%
\tabularnewline
\selectlanguage{english}%
{\scriptsize{}+ nms oracle}\selectlanguage{british}%
 & \selectlanguage{english}%
{\scriptsize{}77.2}\selectlanguage{british}%
 & \selectlanguage{english}%
{\scriptsize{}87.1}\selectlanguage{british}%
 & \selectlanguage{english}%
{\scriptsize{}76.6}\selectlanguage{british}%
 & \selectlanguage{english}%
{\scriptsize{}67.6}\selectlanguage{british}%
 & \selectlanguage{english}%
{\scriptsize{}48.2}\selectlanguage{british}%
 & \selectlanguage{english}%
{\scriptsize{}84.8}\selectlanguage{british}%
 & \selectlanguage{english}%
{\scriptsize{}85.2}\selectlanguage{british}%
 & \selectlanguage{english}%
{\scriptsize{}87.1}\selectlanguage{british}%
 & \selectlanguage{english}%
{\scriptsize{}52.1}\selectlanguage{british}%
 & \selectlanguage{english}%
{\scriptsize{}83.9}\selectlanguage{british}%
 & \selectlanguage{english}%
{\scriptsize{}72.7}\selectlanguage{british}%
 & \selectlanguage{english}%
{\scriptsize{}86.7}\selectlanguage{british}%
 & \selectlanguage{english}%
{\scriptsize{}87.2}\selectlanguage{british}%
 & \selectlanguage{english}%
{\scriptsize{}84.2}\selectlanguage{british}%
 & \selectlanguage{english}%
{\scriptsize{}77.6}\selectlanguage{british}%
 & \selectlanguage{english}%
{\scriptsize{}44.2}\selectlanguage{british}%
 & \selectlanguage{english}%
{\scriptsize{}75.1}\selectlanguage{british}%
 & \selectlanguage{english}%
{\scriptsize{}73.5}\selectlanguage{british}%
 & \selectlanguage{english}%
{\scriptsize{}83.5}\selectlanguage{british}%
 & \selectlanguage{english}%
{\scriptsize{}65.4}\selectlanguage{british}%
 &  & \selectlanguage{english}%
{\scriptsize{}75.0}\selectlanguage{british}%
\tabularnewline
\selectlanguage{english}%
{\scriptsize{}+ both oracles}\selectlanguage{british}%
 & \selectlanguage{english}%
{\scriptsize{}83.7}\selectlanguage{british}%
 & \selectlanguage{english}%
{\scriptsize{}87.8}\selectlanguage{british}%
 & \selectlanguage{english}%
{\scriptsize{}79.6}\selectlanguage{british}%
 & \selectlanguage{english}%
{\scriptsize{}72.6}\selectlanguage{british}%
 & \selectlanguage{english}%
{\scriptsize{}61.6}\selectlanguage{british}%
 & \selectlanguage{english}%
{\scriptsize{}85.1}\selectlanguage{british}%
 & \selectlanguage{english}%
{\scriptsize{}88.2}\selectlanguage{british}%
 & \selectlanguage{english}%
{\scriptsize{}87.8}\selectlanguage{british}%
 & \selectlanguage{english}%
{\scriptsize{}55.7}\selectlanguage{british}%
 & \selectlanguage{english}%
{\scriptsize{}84.0}\selectlanguage{british}%
 & \selectlanguage{english}%
{\scriptsize{}74.8}\selectlanguage{british}%
 & \selectlanguage{english}%
{\scriptsize{}87.3}\selectlanguage{british}%
 & \selectlanguage{english}%
{\scriptsize{}87.5}\selectlanguage{british}%
 & \selectlanguage{english}%
{\scriptsize{}84.9}\selectlanguage{british}%
 & \selectlanguage{english}%
{\scriptsize{}86.8}\selectlanguage{british}%
 & \selectlanguage{english}%
{\scriptsize{}50.3}\selectlanguage{british}%
 & \selectlanguage{english}%
{\scriptsize{}76.2}\selectlanguage{british}%
 & \selectlanguage{english}%
{\scriptsize{}74.6}\selectlanguage{british}%
 & \selectlanguage{english}%
{\scriptsize{}85.3}\selectlanguage{british}%
 & \selectlanguage{english}%
{\scriptsize{}67.6}\selectlanguage{british}%
 &  & \selectlanguage{english}%
{\scriptsize{}78.1}\selectlanguage{british}%
\tabularnewline
\end{tabular}\hspace*{\fill}\vspace{-1mm}

\protect\caption{\label{tab:pascal-FRCN-L-per-class-oracles}Fast R-CNN (model L) detection
results on PASCAL 2007 test using \texttt{Edge\-Boxes\-AR} and given
access to ``oracles'' that provide additional information to the
detector. Given access to both oracles, the only way to further improve
detector performance would be to avoid proposals on background or
to learn a more discriminative classifier. See text for details. }
\end{table*}

\subsection{\label{sub:tuning-proposal-methods}Tuning proposal methods}

All previous experiments evaluate proposal methods using original
parameter settings. However many methods have free parameters that
allow for fine-tuning. For example, when adjusting window sampling
density and the non-maximum suppression (nms) in \texttt{Edge\-Boxes}
\cite{Zitnick2014Eccv}, it is possible to trade-off low recall with
good localisation for higher recall with worse localisation (a similar
observation was made in \cite{Blaschko2013Scia}). Figure~\ref{fig:edge-box-tuning}
compares different versions of \texttt{Edge\-Boxes} tuned to maximise
recall at different IoU points $\Delta$ (we set $\alpha=\max(0.65,\,\Delta-0.05),\,\beta=\Delta+0.05$,
see \cite{Zitnick2014Eccv} for details). \texttt{Edge\-Boxes} tuned
for $\Delta=0.70$ or $0.75$ maximises AR and also results in the
best detection results. 

While originally \texttt{Edge\-Boxes} allowed for optimising recall
for a particular IoU threshold, we consider a new variant that directly
maximises AR (marked `AR' in figure \ref{fig:edge-box-tuning}) to
further explore the link between AR and detection quality. To do so,
we alter its greedy nms procedure to make it \emph{adaptive}. We start
with a large nms threshold $\beta_{0}$ to encourage dense sampling
around the top scoring candidates (a window is suppressed if its IoU
with a higher scoring window exceeds this threshold). After selecting
each proposal, $\beta_{k+1}$ is decreased slightly via $\beta_{k+1}=\beta_{k}\cdot\eta$
to encourage greater proposal diversity. Setting $\beta_{0}=0.90$
and $\eta=0.9996$ gave best AR at $1\,000$ proposals on the PASCAL
validation set (we kept $\alpha=0.65$ fixed). This new adaptive \texttt{Edge\-Boxes}
variant is not optimal at any particular IoU threshold, but has best
overall AR and improves Fast R-CNN mAP by $1.6$ over the best previous
variant (reaching $62.0\ \mbox{mAP}$). 

The results in figure \ref{fig:edge-box-tuning} further support our
conclusion that AR is a good predictor for mAP and suggest that it
can be used for fine-tuning proposal methods. We expect other methods
to improve as well if optimised for AR instead of a particular IoU
threshold.

\subsection{\label{sub:oracle-experiments}Detection with oracles}

We finish by exploring the limits of proposal methods when coupled
with Fast R-CNN and given access to ``oracles'' that provide additional
information to the detector. For these experiments we use the \texttt{Edge\-Boxes\-AR
}proposals described in \S\ref{sub:tuning-proposal-methods} which
gave the best results of all evaluated methods when coupled with the
Fast R-CNN model M. Re-training the larger model L with \texttt{Edge\-Boxes\-AR}
proposals improves mAP to 67.8 (compared to 66.7 using \texttt{Se\-lec\-tive\-Search}
proposals as in \cite{Girshick2015arXiv}).

We test two oracles. First, we augment the set of proposals with all
ground truth annotations (\emph{gt} oracle), which results in AR of
1 (but contains many false positives). Second, we perform optimal,
per-class non-maximum suppression (\emph{nms} oracle) that suppresses
all false positives that overlap any true positives (without suppressing
any true positives, and keeping false positives in the background).
Results for the gt and nms oracles are shown in table \ref{tab:pascal-FRCN-L-per-class-oracles}. 

The gt oracle improves mAP by about $3\%$. The nms oracle has the
overall stronger effect with about $7\%$ mAP improvement. Combining
both oracles improves mAP by about $10\%$, indicating that their
effect is largely orthogonal. All remaining mistakes that prevent
perfect detection are confusions on the background or misclassifications.
Therefore, the only way to further improve detector performance would
be to avoid proposals on background or to learn a more discriminative
classifier.

\section{\label{sec:Discussion-Conclusion}Discussion}

In this work we have revisited the majority of existing detection
proposal methods, proposed new evaluation metrics, and performed an
extensive and direct comparison of existing methods. Our primary goal
has been to enable practitioners to make more informed decisions when
considering use of detection proposals and selecting the optimal proposal
method for a given scenario. Additionally, our open source benchmark
will enable more complete and informative evaluations in future research
on detection proposals. We conclude by summarising our key findings
and suggesting avenues for future work.

\paragraph*{Repeatability}

We found that the \emph{repeatability} of virtually all proposal methods
is limited: imperceivably small changes to an image cause a noticeable
change in the set of produced proposals. Even changing a single image
pixel already exhibits measurable differences in repeatability. We
foresee room for improvement by using more robust superpixel (or boundary
estimation) methods. However, while better repeatability for object
detection would be desirable, it is not the most important property
of proposals. Image independent methods such as \texttt{Sli\-ding\-Win\-dow}
and \texttt{Cracking\-Bing} have perfect repeatability but are inadequate
for detection. Methods such as \texttt{Se\-lec\-tive\-Search} and
\texttt{Edge\-Boxes} seem to strike a better balance between recall
and repeatability. We suspect that high quality proposal methods that
are also more repeatable would yield improved detection accuracy,
however this has yet to be verified experimentally.

\paragraph*{Localisation Accuracy}

Our analysis showed that for object detection improving proposal\textbf{
}\emph{localisation accuracy} (improved IoU) is as important as improving
recall. Indeed, we demonstrated that the popular metric of recall
at IoU of 0.5 is not predictive of detection accuracy. As far as we
know, our experiments are the first to demonstrate this. Proposals
with high recall but at low overlap are not effective for detection.

\paragraph*{Average Recall}

To simultaneously measure both proposal recall and localisation accuracy,
we report \emph{average recall} (AR), which summarises the distribution
of recall across a range of overlap thresholds. For a fixed number
of proposals, AR correlates surprisingly well with detector performance
(for LM-LLDA, R-CNN, and Fast R-CNN). AR proves to be an excellent
predictor of detection performance both for comparing competing methods
as well as tuning a specific method's parameters. We encourage future
work to report average recall (as shown in figures \ref{fig:recall-vs-iuo-area-vs-number-of-proposals-pascal}/\ref{fig:recall-vs-iuo-area-vs-number-of-proposals-imagenet})
as the primary metric for evaluating proposals for object detection.
For detectors more robust to localisation errors (e.g. Fast R-CNN),
the IoU range of the AR metric can be modified to better predict detector
performance.

\paragraph*{Top Methods}

Amongst the evaluated methods,\texttt{ Se\-lec\-tive\-Search},
\texttt{Ri\-gor}, \texttt{MCG}, and \texttt{Edge\-Boxes} consistently
achieved top object detection performance when coupled with diverse
object detectors. If fast proposals are required, \texttt{Edge\-Boxes}
provides a good compromise between speed and quality. Surprisingly,
these top methods all achieve fairly similar detection performance
even though they employ very different mechanisms for generating proposals.
\texttt{Se\-lec\-tive\-Search} merges superpixels,\texttt{ Ri\-gor}
computes multiple graph cut segmentations, \texttt{MCG} generates
hierarchical segmentations, and \texttt{Edge\-Boxes} scores windows
based on edge content.

\paragraph*{Generalisation}

Critically, we measured no significant drop in recall when going from
the 20 PASCAL categories to the 200 ImageNet categories. Moreover,
while MS COCO is substantially harder and has very different statistics
(more and smaller objects), relative method ordering remains mostly
unchanged. These are encouraging result indicating that \emph{current
methods do indeed generalise to different unseen categories}, and
as such can be considered true ``objectness'' methods.

\paragraph*{Oracle Experiments}

The best Fast R-CNN results reported in this paper used the large
model L and \texttt{Edge\-Boxes\-AR} proposals, achieving mAP of
67.8 on PASCAL 2007 test. Using an oracle to rectify all localisation
and recall errors improved performance to 71.2 mAP, and adding an
oracle for perfect non-maximum suppression further improved mAP to
78.1 (see \S\ref{sub:oracle-experiments} for details). The remaining
gap of 21.9 mAP to reach perfect detection is caused by high scoring
detections on the background and object misclassifications. This best
case analysis for proposals that are perfectly localised shows that
further improvement can only be gained by removing false positives
in the proposal stage (producing fewer proposals while maintaining
high AR) or training a more discriminative classifier.

\paragraph*{Discussion}

Do object proposals improve detection quality or are they just a transition
technology until we have sufficient computing power? On the one hand,
simply increasing the number of proposals, or using additional random
proposals, may actually harm detection performance as shown in \cite{Girshick2015arXiv}.
On the other hand, there is no fundamental difference between the
pipeline of object proposals with a detector and a cascaded detector
with two stages. Conceptually, a sliding window detector with access
to the features of the proposal method may be able to perform at least
as well as the cascade and as such detection proposals independent
of the final classifier may eventually become unnecessary. Given enough
computing power and an adequate training procedure, one might expect
that a dense evaluation of CNNs could further improve performance
over R-CNNs.

While in this work we have focused on object detection, object proposals
have other uses. For example, they can be used to handle unknown categories
at test time, or to enable weakly supervised learning \cite{Vicente2011Cvpr,Guillaumin2014Ijcv,Tang2014Cvpr}. 

Finally, we observe that current proposal methods reach high recall
while using features that are not utilised by detectors such as LM-LLDA,
R-CNN, and Fast R-CNN (e.g. object boundaries and superpixels). Conversely,
with the exception of \texttt{Mul\-ti\-box} \cite{Erhan2014Cvpr},
none of the proposal methods use CNN features. We expect some cross-pollination
will occur in this space. Indeed, there has been some very recent
work in this space \cite{Weicheng2015arXiv,Ren2015arXiv,Pinheiro2015arXiv}
that shows promising results.

In the future, detection proposals will surely improve in repeatability,
recall, localisation accuracy, and speed. Top-down reasoning will
likely play a more central role as purely bottom-up processes have
difficulty generating perfect object proposals. We may also see a
tighter integration between proposals and the detector, and the segmentation
mask generated by many proposal methods may play a more important
role during detection. One thing is clear: progress has been rapid
in this young field and we expect proposal methods to evolve quickly
over the coming years.

\appendices{}

\section{Analysis of Metrics}

\label{sec:additional-metrics}Average recall (AR) between 0.5 and
1 can also be computed by averaging over the overlaps of each annotation
$\text{{gt}}_{i}$ with the closest matched proposal, that is integrating
over the $y$ axis of the plot instead of the $x$ axis. Let $o$
be the IoU overlap and recall($o$) the function shown for example
in figure \ref{fig:recall-vs-iuo-at-1000-windows}. Let IoU($\text{{gt}}_{i}$)
denote the IoU between the annotation $\text{{gt}}_{i}$ and the closest
detection proposal. We can then write:

\[
\text{{AR}}=2\intop_{0.5}^{1}\text{{recall}}(o)\diff o=\frac{2}{n}\sum_{i=1}^{n}\max\left(\text{{IoU}}(\text{{gt}}_{i})-0.5,0\right)
\]
which is the same as the \textit{average best overlap} (ABO) \cite{Carreira2012Pami}
or the average \textit{best spatial support} (BSS) \cite{Malisiewicz2007Bmvc}
truncated at 0.5 IoU.

The ABO and BSS are typically computed by assigning the closest proposal
to each annotation, i.e.~a proposal can match more than one annotation.
In contrast, for all our experiments we compute a bipartite matching
to assign proposals to annotations (using a greedy algorithm for efficiency
instead of the optimal Hungarian algorithm). 

The \textit{volume-under-surface} metric (VUS) \cite{Manen2013Iccv}
plots recall as a function of both overlap and proposal count and
computes the volume under that surface. Since in practice detectors
utilize a fixed number of proposals, the VUS of a proposal method
is only an indirect predictor of detection accuracy.

\bibliographystyle{IEEEtran}
\bibliography{2015_pami_detection_proposals}

\begin{thebibliography}{10}
\providecommand{\url}[1]{#1}
\csname url@samestyle\endcsname
\providecommand{\newblock}{\relax}
\providecommand{\bibinfo}[2]{#2}
\providecommand{\BIBentrySTDinterwordspacing}{\spaceskip=0pt\relax}
\providecommand{\BIBentryALTinterwordstretchfactor}{4}
\providecommand{\BIBentryALTinterwordspacing}{\spaceskip=\fontdimen2\font plus
\BIBentryALTinterwordstretchfactor\fontdimen3\font minus
  \fontdimen4\font\relax}
\providecommand{\BIBforeignlanguage}[2]{{%
\expandafter\ifx\csname l@#1\endcsname\relax
\typeout{** WARNING: IEEEtran.bst: No hyphenation pattern has been}%
\typeout{** loaded for the language `#1'. Using the pattern for}%
\typeout{** the default language instead.}%
\else
\language=\csname l@#1\endcsname
\fi
#2}}
\providecommand{\BIBdecl}{\relax}
\BIBdecl

\bibitem{Papageorgiou2000}
C.~Papageorgiou and T.~Poggio, ``A trainable system for object detection,''
  \emph{IJCV}, 2000.

\bibitem{Viola2004Ijvc}
P.~Viola and M.~Jones, ``Robust real-time face detection,'' in \emph{IJCV},
  2004.

\bibitem{Felzenszwalb2010Pami}
P.~Felzenszwalb, R.~Girshick, D.~McAllester, and D.~Ramanan, ``Object detection
  with discriminatively trained part-based models,'' \emph{PAMI}, 2010.

\bibitem{Everingham2014Ijcv}
M.~Everingham, S.~Eslami, L.~{Van Gool}, C.~Williams, J.~Winn, and
  A.~Zisserman, ``The pascal visual object classes challenge -- a
  retrospective,'' \emph{IJCV}, 2014.

\bibitem{Deng2009Cvpr}
J.~Deng, W.~Dong, R.~Socher, L.-J. Li, K.~Li, and L.~Fei-Fei, ``{ImageNet: A
  Large-Scale Hierarchical Image Database},'' in \emph{CVPR}, 2009.

\bibitem{mscoco2015}
T.-Y. Lin, M.~Maire, S.~Belongie, L.~Bourdev, R.~Girshick, J.~Hays, P.~Perona,
  D.~Ramanan, C.~L. Zitnick, and P.~Doll{\'a}r, ``Microsoft {COCO}: Common
  objects in context,'' \emph{arXiv:1405.0312}, 2015.

\bibitem{Wang2013Iccv}
X.~Wang, M.~Yang, S.~Zhu, and Y.~Lin, ``Regionlets for generic object
  detection,'' in \emph{ICCV}, 2013.

\bibitem{Girshick2014Cvpr}
R.~Girshick, J.~Donahue, T.~Darrell, and J.~Malik, ``Rich feature hierarchies
  for accurate object detection and semantic segmentation,'' in \emph{CVPR},
  2014.

\bibitem{Szegedy2014arXiv}
C.~Szegedy, S.~Reed, D.~Erhan, and D.~Anguelov, ``Scalable, high-quality object
  detection,'' \emph{arXiv:1412.1441}, 2014.

\bibitem{He2014Eccv}
K.~He, X.~Zhang, S.~Ren, and J.~Sun, ``Spatial pyramid pooling in deep
  convolutional networks for visual recognition,'' in \emph{ECCV}, 2014.

\bibitem{Cinbis2013Iccv}
R.~G. Cinbis, J.~Verbeek, and C.~Schmid, ``Segmentation driven object detection
  with fisher vectors,'' in \emph{{ICCV}}, 2013.

\bibitem{Alexe2010Cvpr}
B.~Alexe, T.~Deselaers, and V.~Ferrari, ``What is an object?'' in \emph{CVPR},
  2010.

\bibitem{Carreira2010Cvpr}
J.~Carreira and C.~Sminchisescu, ``Constrained parametric min-cuts for
  automatic object segmentation,'' in \emph{CVPR}, 2010.

\bibitem{Endres2010Eccv}
I.~Endres and D.~Hoiem, ``Category independent object proposals,'' in
  \emph{ECCV}, 2010.

\bibitem{Sande2011Iccv}
K.~{van de Sande}, J.~Uijlings, T.~Gevers, and A.~Smeulders, ``Segmentation as
  selective search for object recognition,'' in \emph{ICCV}, 2011.

\bibitem{Girshick2015arXiv}
R.~Girshick, ``{Fast R-CNN},'' \emph{arXiv:1504.08083}, 2015.

\bibitem{Hosang2014Bmvc}
J.~Hosang, R.~Benenson, and B.~Schiele, ``How good are detection proposals,
  really?'' in \emph{BMVC}, 2014.

\bibitem{Cheng2014Cvpr}
M.-M. Cheng, Z.~Zhang, W.-Y. Lin, and P.~H.~S. Torr, ``{BING}: Binarized normed
  gradients for objectness estimation at 300fps,'' in \emph{CVPR}, 2014.

\bibitem{Carreira2012Pami}
J.~Carreira and C.~Sminchisescu, ``Cpmc: Automatic object segmentation using
  constrained parametric min-cuts.'' \emph{PAMI}, 2012.

\bibitem{Zitnick2014Eccv}
C.~Zitnick and P.~Doll\'ar, ``Edge boxes: Locating object proposals from
  edges,'' in \emph{ECCV}, 2014.

\bibitem{Endres2014Pami}
I.~Endres and D.~Hoiem, ``Category-independent object proposals with diverse
  ranking,'' in \emph{PAMI}, 2014.

\bibitem{Kraehenbuehl2014Eccv}
P.~Kr\"ahenb\"uhl and V.~Koltun, ``Geodesic object proposals,'' in \emph{ECCV},
  2014.

\bibitem{Arbelaez2014Cvpr}
P.~Arbelaez, J.~Pont-Tuset, J.~Barron, F.~Marqu{\'e}s, and J.~Malik,
  ``Multiscale combinatorial grouping,'' in \emph{CVPR}, 2014.

\bibitem{Alexe2012Pami}
B.~Alexe, T.~Deselares, and V.~Ferrari, ``Measuring the objectness of image
  windows,'' \emph{PAMI}, 2012.

\bibitem{Rahtu2011Iccv}
E.~Rahtu, J.~Kannala, and M.~Blaschko, ``Learning a category independent object
  detection cascade,'' in \emph{ICCV}, 2011.

\bibitem{Manen2013Iccv}
S.~Man\'en, M.~Guillaumin, and L.~{Van Gool}, ``Prime object proposals with
  randomized prim's algorithm,'' in \emph{ICCV}, 2013.

\bibitem{Rantalankila2014Cvpr}
P.~Rantalankila, J.~Kannala, and E.~Rahtu, ``Generating object segmentation
  proposals using global and local search,'' in \emph{CVPR}, 2014.

\bibitem{Humayun2014Cvpr}
A.~Humayun, F.~Li, and J.~M. Rehg, ``Rigor: Recycling inference in graph cuts
  for generating object regions,'' in \emph{CVPR}, 2014.

\bibitem{Uijlings2013Ijcv}
J.~Uijlings, K.~{van de Sande}, T.~Gevers, and A.~Smeulders, ``Selective search
  for object recognition,'' \emph{IJCV}, 2013.

\bibitem{Tuytelaars2008FoundationsandTrends}
T.~Tuytelaars and K.~Mikolajczyk, ``Local invariant feature detectors: a
  survey,'' \emph{Foundations and Trends in Computer Graphics and Vision},
  2008.

\bibitem{Mikolajczyk2005Ijcv}
K.~Mikolajczyk, T.~Tuytelaars, C.~Schmid, A.~Zisserman, J.~Matas,
  F.~Schaffalitzky, T.~Kadir, and L.~V. Gool, ``A comparison of affine region
  detectors,'' \emph{IJCV}, 2005.

\bibitem{Tuytelaars2010Cvpr}
T.~Tuytelaars, ``Dense interest points,'' in \emph{CVPR}, 2010.

\bibitem{Fragkiadaki2015Cvpr}
K.~Fragkiadaki, P.~Arbelaez, P.~Felsen, and J.~Malik, ``Learning to segment
  moving objects in videos,'' in \emph{CVPR}, 2015.

\bibitem{Gu2009Cvpr}
C.~Gu, J.~Lim, P.~Arbelaez, and J.~Malik, ``Recognition using regions,'' in
  \emph{CVPR}, 2009.

\bibitem{Arbelaez2011Pami}
P.~Arbelaez, M.~Maire, C.~Fowlkes, and J.~Malik, ``Contour detection and
  hierarchical image segmentation,'' \emph{PAMI}, 2011.

\bibitem{Dollar2015Pami}
P.~Doll\'ar and C.~L. Zitnick, ``Fast edge detection using structured
  forests,'' \emph{PAMI}, 2015.

\bibitem{Felzenszwalb2004IJCV}
P.~F. Felzenszwalb and D.~P. Huttenlocher, ``Efficient graph-based image
  segmentation,'' \emph{IJCV}, 2004.

\bibitem{Chang2011Iccv}
K.-Y. Chang, T.-L. Liu, H.-T. Chen, and S.-H. Lai, ``Fusing generic objectness
  and visual saliency for salient object detection,'' in \emph{ICCV}, 2011.

\bibitem{Lim2013Cvpr}
J.~Lim, C.~L. Zitnick, and P.~Doll\'ar, ``Sketch tokens: A learned mid-level
  representation for contour and object detection,'' in \emph{CVPR}, 2013.

\bibitem{Blaschko2013Scia}
M.~Blaschko, J.~Kannala, and E.~Rahtu, ``{Non Maximal Suppression in Cascaded
  Ranking Models},'' in \emph{{Scandanavian Conference on Image Analysis}},
  2013.

\bibitem{Zhao2014Bmvc}
Q.~Zhao, Z.~Liu, and B.~Yin, ``Cracking {BING} and beyond,'' in \emph{BMVC},
  2014.

\bibitem{DollarICCV13edges}
P.~Doll\'ar and C.~L. Zitnick, ``Structured forests for fast edge detection,''
  in \emph{ICCV}, 2013.

\bibitem{Feng2011Iccv}
J.~Feng, Y.~Wei, L.~Tao, C.~Zhang, and J.~Sun, ``Salient object detection by
  composition,'' in \emph{ICCV}, 2011.

\bibitem{Zhang2011Cvpr}
Z.~Zhang, J.~Warrell, and P.~H.~S. Torr, ``Proposal generation for object
  detection using cascaded ranking svms,'' in \emph{CVPR}, 2011.

\bibitem{Bergh2013Iccv}
M.~{Van Den Bergh}, G.~Roig, X.~Boix, S.~Manen, and L.~{Van Gool}, ``Online
  video seeds for temporal window objectness,'' in \emph{ICCV}, 2013.

\bibitem{Bergh2014Ijcv}
M.~{Van den Bergh}, X.~Boix, G.~Roig, and L.~{Van Gool}, ``Seeds: Superpixels
  extracted via energy-driven sampling,'' \emph{IJCV}, 2014.

\bibitem{Kim2012Eccv}
J.~Kim and K.~Grauman, ``{Shape Sharing for Object Segmentation},'' in
  \emph{ECCV}, 2012.

\bibitem{Erhan2014Cvpr}
D.~Erhan, C.~Szegedy, A.~Toshev, and D.~Anguelov, ``Scalable object detection
  using deep neural networks,'' in \emph{CVPR}, 2014.

\bibitem{Bourdev2005Cvpr}
L.~Bourdev and J.~Brandt, ``Robust object detection via soft cascade,'' in
  \emph{CVPR}, 2005.

\bibitem{Harzallah2009Iccv}
H.~Harzallah, F.~Jurie, and C.~Schmid, ``Combining efficient object
  localization and image classification,'' in \emph{ICCV}, 2009.

\bibitem{Dollar2012Eccv}
P.~Doll\'ar, R.~Appel, and W.~Kienzle, ``Crosstalk cascades for frame-rate
  pedestrian detection,'' in \emph{ECCV}, 2012.

\bibitem{Torralba2007Pami}
A.~Torralba, K.~P. Murphy, and W.~T. Freeman, ``Sharing visual features for
  multiclass and multiview object detection,'' \emph{PAMI}, 2007.

\bibitem{Zehnder2008Bmvc}
P.~Zehnder, E.~Koller-Meier, and L.~{Van Gool}, ``An efficient shared
  multi-class detection cascade,'' in \emph{BMVC}, 2008.

\bibitem{Chavali2015arXiv}
N.~{Chavali}, H.~{Agrawal}, A.~{Mahendru}, and D.~{Batra}, ``Object-proposal
  evaluation protocol is 'gameable','' \emph{arXiv:1505.05836}, 2015.

\bibitem{Hoiem2012Eccv}
D.~Hoiem, Y.~Chodpathumwan, and Q.~Dai, ``{Diagnosing Error in Object
  Detectors},'' in \emph{{ECCV}}, 2012.

\bibitem{Girshick2013ICCV}
R.~Girshick and J.~Malik, ``Training deformable part models with decorrelated
  features,'' in \emph{ICCV}, 2013.

\bibitem{Vicente2011Cvpr}
S.~Vicente, C.~Rother, and V.~Kolmogorov, ``Object cosegmentation,'' in
  \emph{CVPR}, 2011.

\bibitem{Guillaumin2014Ijcv}
M.~Guillaumin, D.~Kuttel, and V.~Ferrari, ``Imagenet auto-annotation with
  segmentation propagation,'' \emph{IJCV}, 2014.

\bibitem{Tang2014Cvpr}
K.~Tang, A.~Joulin, L.-J. Li, and L.~Fei-Fei, ``Co-localization in real-world
  images,'' in \emph{CVPR}, 2014.

\bibitem{Weicheng2015arXiv}
W.~Kuo, B.~Hariharan, and J.~Malik, ``Deepbox: Learning objectness with
  convolutional networks,'' \emph{arXiv:1505.02146}, 2015.

\bibitem{Ren2015arXiv}
S.~{Ren}, K.~{He}, R.~{Girshick}, and J.~{Sun}, ``Faster {R-CNN}: Towards
  real-time object detection with region proposal networks,''
  \emph{arXiv:1506.01497}, 2015.

\bibitem{Pinheiro2015arXiv}
P.~O. Pinheiro, R.~Collobert, and P.~Doll\'ar, ``Learning to segment object
  candidates,'' \emph{arXiv:1506.06204}, 2015.

\bibitem{Malisiewicz2007Bmvc}
T.~Malisiewicz and A.~A. Efros, ``Improving spatial support for objects via
  multiple segmentations,'' in \emph{BMVC}, 2007.

\end{thebibliography}
\end{document}